%% file: main.tex
\documentclass[10pt,twocolumn,letterpaper]{article}

\usepackage[]{cvpr}      
\usepackage[colorinlistoftodos,prependcaption,textsize=tiny]{todonotes}
\usepackage{bbm}
\usepackage{soul}
\usepackage[accsupp]{axessibility}


\definecolor{cvprblue}{rgb}{0.0,1.0,0.0}
\usepackage[pagebackref,breaklinks,colorlinks,citecolor=cvprblue]{hyperref}
\usepackage{graphicx}
\usepackage{multirow}
\usepackage{booktabs}
\usepackage[export]{adjustbox}
\usepackage{tabularx}
\usepackage{multicol}
\def\paperID{7921} 
\def\confName{CVPR}
\def\confYear{2024}

\newcommand{\glidar}{\texttt{\textbf{GLiDR}}}

\title{GLiDR: Topologically Regularized Graph Generative Network for Sparse
LiDAR Point Clouds}

\author{Prashant Kumar$^{1}$ \quad Kshitij Madhav Bhat$^{2}$ \quad Vedang Bhupesh Shenvi Nadkarni$^{3}$ \quad Prem Kalra$^{1}$ \vspace{0.3em} \\
{\normalsize $^1$IIT Delhi} \quad
{\normalsize $^2$IIT Indore} \quad {\normalsize $^3$BITS Pilani} \\
{\tt\small prashantk.nan@gmail.com} \quad
{\tt\small kshitijmbhat@gmail.com} \quad {\tt\small vedang.nadkarni@gmail.com} \quad
{\tt\small pkalra@cse.iitd.ac.in}
}

\begin{document}
\twocolumn[{%
\renewcommand\twocolumn[1][]{#1}%
\maketitle
\begin{center}
\setlength{\belowcaptionskip}{0pt}
    \centering
    \captionsetup{type=figure}
    \includegraphics[width=\linewidth]{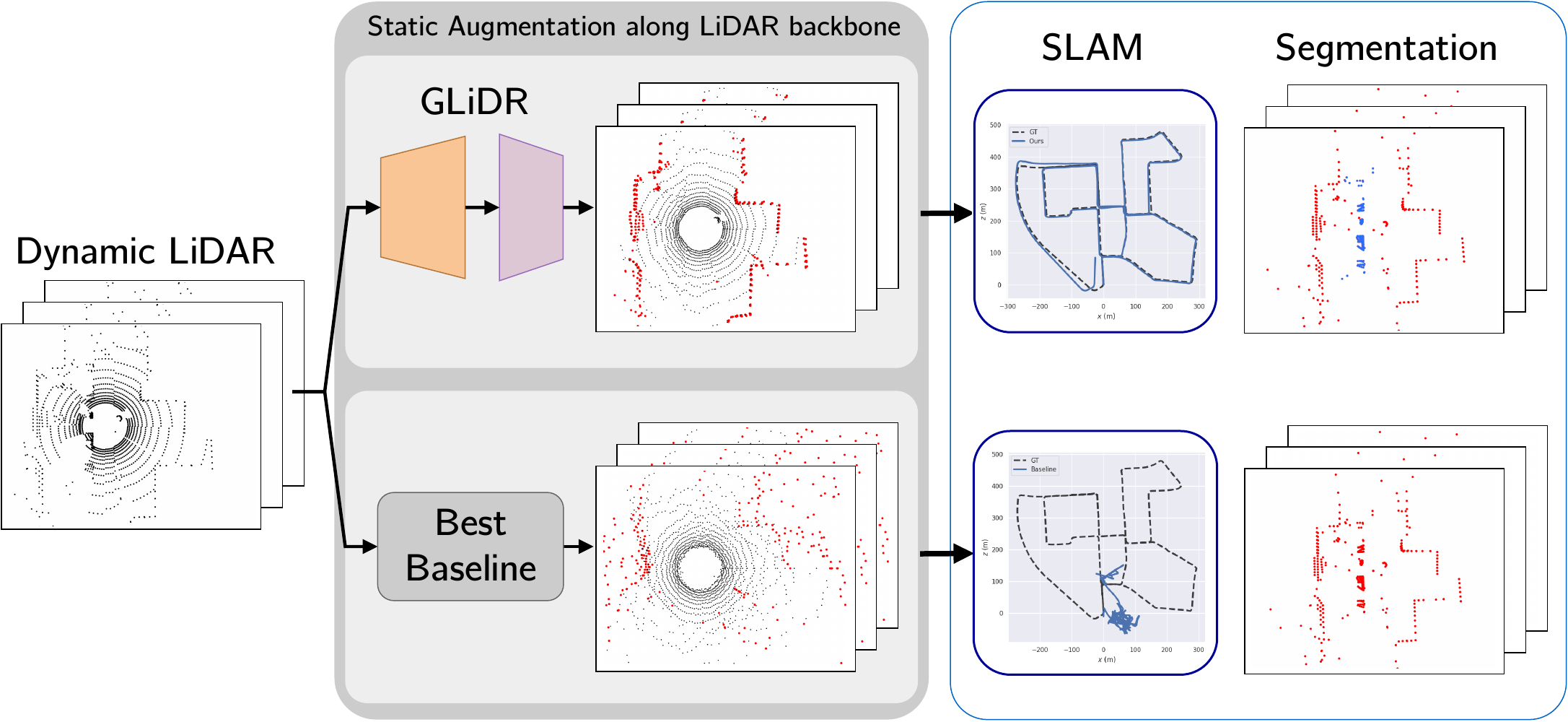}
    \captionof{figure}{\glidar{} outperforms the best baselines on SLAM results with sparse LiDAR scans. Binary segmentation masks of dynamic and static points (blue denotes dynamic and red denotes static points) obtained using \glidar{} are also better.}
    \label{fig:lidar_recon}
\end{center}%
}]


\input{sec/00abstract}    
\input{sec/01introduction}

\input{sec/02related-work}

\input{sec/03Background}

\input{sec/04Methodology}
\input{sec/05Experiments}

\input{sec/06ablation}
\input{sec/07conclusion}

{
    \small
    \bibliographystyle{ieeenat_fullname}
    \bibliography{main}
}

\end{document}


\maketitle

\section{Persistent Homology}
\begin{figure}
    \centering
    \includegraphics[width=\linewidth]{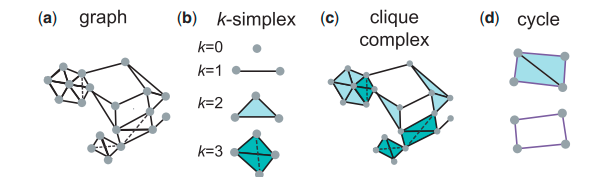}
    \includegraphics[width=\linewidth]{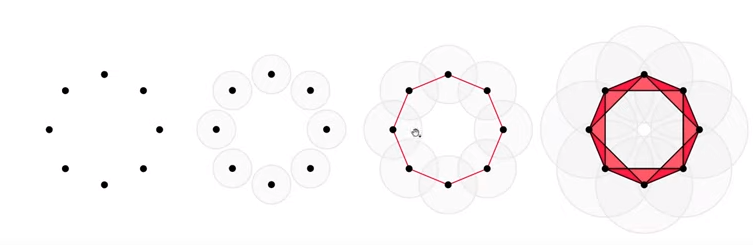}
    \includegraphics[width=\linewidth]{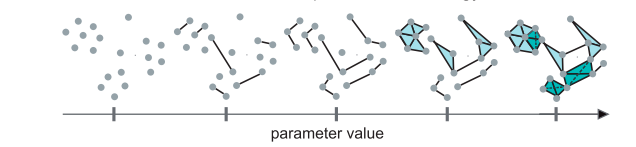}
    \caption{\textbf{Row 1} denotes \textit{k}-simplexes and simplicial complexes. These are generalizations of graphs to higher dimensions. \textbf{Row 2}: $\mathcal{PH}$ discovers the global shape of a dataset. \textbf{Row 3} denotes the progression of filtration on a point cloud. As the filtration progresses, new simplicies are added to the simplicial complex. Image taken from Lia \etal \cite{papadopoulos2018network}.}
    \label{fig:topology-simplex}
\end{figure}

We provide a detailed description of Persistent Homology ($\mathcal{PH}$) and its application on point clouds and images. 
A topological space can be encoded as cell complexes - a collection of $k$-dim simplicies $(k=(0,1,2...)$ (Figure \ref{fig:topology-simplex} - row 1). \textit{Homology}, an algebraic invariant of a topological space, uses local computation to capture global shape information ($k$-dim holes) of a topological space (Figure \ref{fig:topology-simplex}- row 2 and 3). These holes, generalized to various dimensions, form the basis of homology \cite{edelsbrunner2008computational}.

\begin{table*}
\setlength{\tabcolsep}{1pt}
\begin{tabular}{c c c c} 
\includegraphics[width=0.25\linewidth]{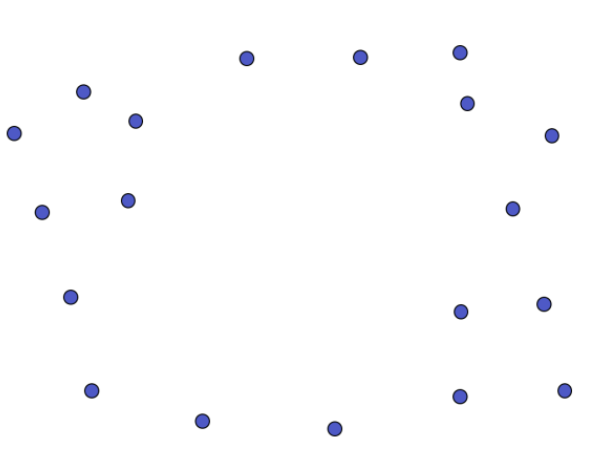} & \includegraphics[width=0.25\linewidth]{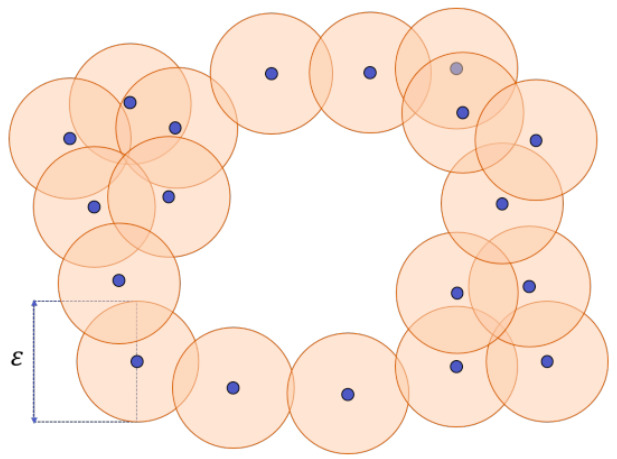} & \includegraphics[width=0.25\linewidth]{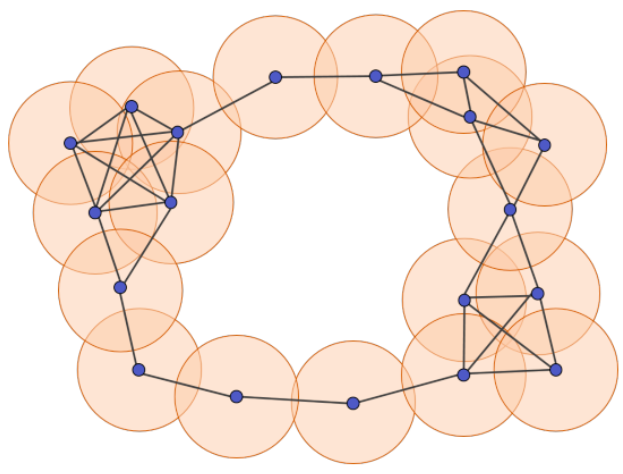} &
\includegraphics[width=0.25\linewidth]{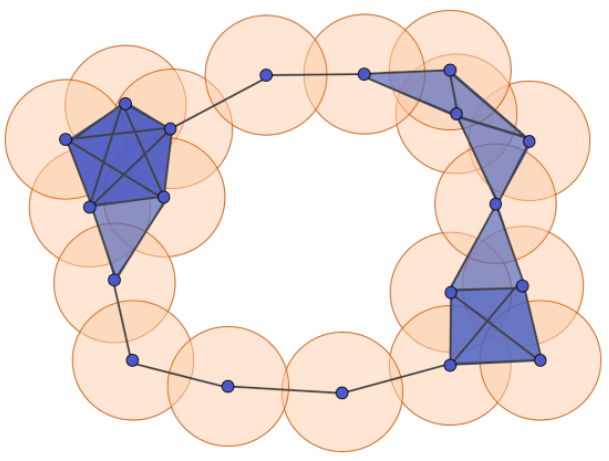}\\
a & b & c & d
\end{tabular}
\caption{Left to Right. Progression of filtration on a point cloud over different spatial resolutions. Image taken from Wright \etal\cite{perhomo}.}
\label{tab:filtration}
\end{table*}

Persistent Homology ($\mathcal{PH}$) is an algebraic method to discover topological features of datasets. It converts a dataset (e.g., point cloud) to a simplicial complex and studies the change of homology across an increasing sequence of simplicial complexes $ \phi \subseteq \mathcal{C}_1 \subseteq \mathcal{C}_2 \subseteq \mathcal{C}_3....\mathcal{C}_i....\mathcal{C}_n = \mathcal{C}$, known as filtration \cite{edelsbrunner2008computational}. Topological features are computed at different spatial resolutions across the subsequence. Examining the persistence of the features over a range of scales reveals insights about the underlying patterns in datasets. For point clouds, the filtration is defined on the edges of the complex. We define a sub-level filtration over $\mathcal{C}$. Every simplicial complex in the subsequence can be mapped to a number using a filtration function $f((v_0,v_1...v_n)) = \max\limits_{i<j; i,j\in 0,1,2,3...n }f(v_i,v_j))$. This filtration, known as flag filtration, is based on the pairwise distance between points and is monotonic (every subsequent simplicial complex has a value higher than the previous).

For example - given a set of points in a 3D space, the filtration can be generated by an increasing $\alpha$-neighbourhood ball for each point (See Table \ref{tab:filtration}). For the given $\alpha$, two balls intersect when two points are no further apart than distance $2\alpha$. At this moment, the two points are connected by an edge. 
For a given $\alpha$, we connect all points that are no further than $2\alpha$. Assuming $k$ edges are introduced at a given $\alpha$,  an ordering is given to these edges (i.e., the edges appear in that sequence). Further, the introduction of an edge (1-D simplex) can lead to the creation of a higher-order simplex (triangle, tetrahedron, and so on). These are also a part of the filtration - lower dimension simplicies are added to the filtration before the higher dimensions. The progress of filtration leads to the construction or destruction of homology ($k$\textit{-dim} holes - connected components, cycles, voids, and so on) based on the following principles-

$\bullet$  The appearance of an edge leads to the creation of a $k$\textit{-dim} hole that is not part of an existing in the previous sequence. This leads to the birth of a feature.

$\bullet$  The appearance of an edge leads to the completion of a $k$\textit{-dim} hole that was discovered previously (i.e., the $k$\textit{-dim} hole is completely filled with equal and lesser dimension simplicies). This leads to the death of the feature that was born at the birth of the hole. For example - the occurrence of four points connected to form a rectangle creates a \textit{1-d} hole (cycle). When the diagonal edge is introduced at a later time in the filtration (for a higher value of $\alpha$), it leads to the rectangle filled with two triangles (2-simplex). This leads to the death of the  \textit{1-d} hole feature. 

Each 1-dim hole appears at a particular value of $\alpha$ and disappears at another value of $\alpha$. The addition of an edge either creates or destroys a homology. For our case, given a set of points in 2D space (a point cloud or an image with pixel values), 
we do not know what $\alpha's$ to use to obtain the most important features. Therefore, we consider all values and observe the change in homology for different values of  $\alpha$. This leads to a nested sequence of increasing subsets of simplicial complexes referred to as a filtration. Each hole appears at a particular value of $\alpha$ (b) and disappears at another value of $\alpha$ (d). The persistence of the hole is represented as a birth-death pair $(b,d)$. Visualizing all such pairs for all holes in the form of bars leads to a \textit{barcode}. Short bars might represent noise, while long bars represent important features.

Unlike point clouds, images do not have a concept of pairwise distance. Instead, the intensity value of the pixels of an image help to create a filtration. For images, the final simplicial complex, $\mathcal{C}$ can be taken to be the triangulation of the image grid where the intersection refer to the pixels of the image. Like point clouds, sub-level set filtrations are used here. The filtration function for the sublevel filtrations is $f((v_0,v_1...v_n)) = \max\limits_{i=0,1,2,3...n }f(v_i))$,  which translates to the maximum intensity value of a pixel in a simplicial complex. The only difference is - instead of using pairwise distance for generating sub-level set filtrations, we use the pixel intensity values. $\alpha$ is initialized to the minimum intensity value. As it increases, the sub-level set increases, and more pixels with intensity $\leq$ $\alpha$ are included in the filtration. New simplicies are added to the existing complex, and the filtration progresses.

\section{Paired Scan Generation}

\begin{figure*}[htbp]
\begin{subfigure}{\textwidth}
  \centering
 \includegraphics[width=1\linewidth]{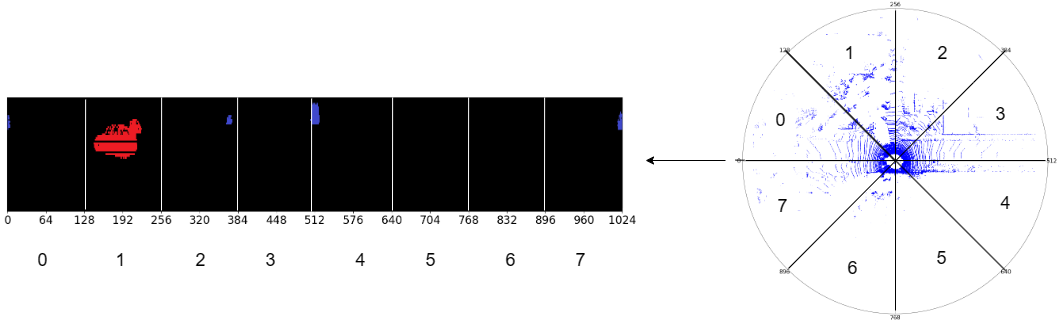} 
  \caption{A LiDAR scan is divided into eight sectors that translate to corresponding regions in the range image segmentation mask.}
   \label{fig:sector_division}
  \includegraphics[width=\linewidth]{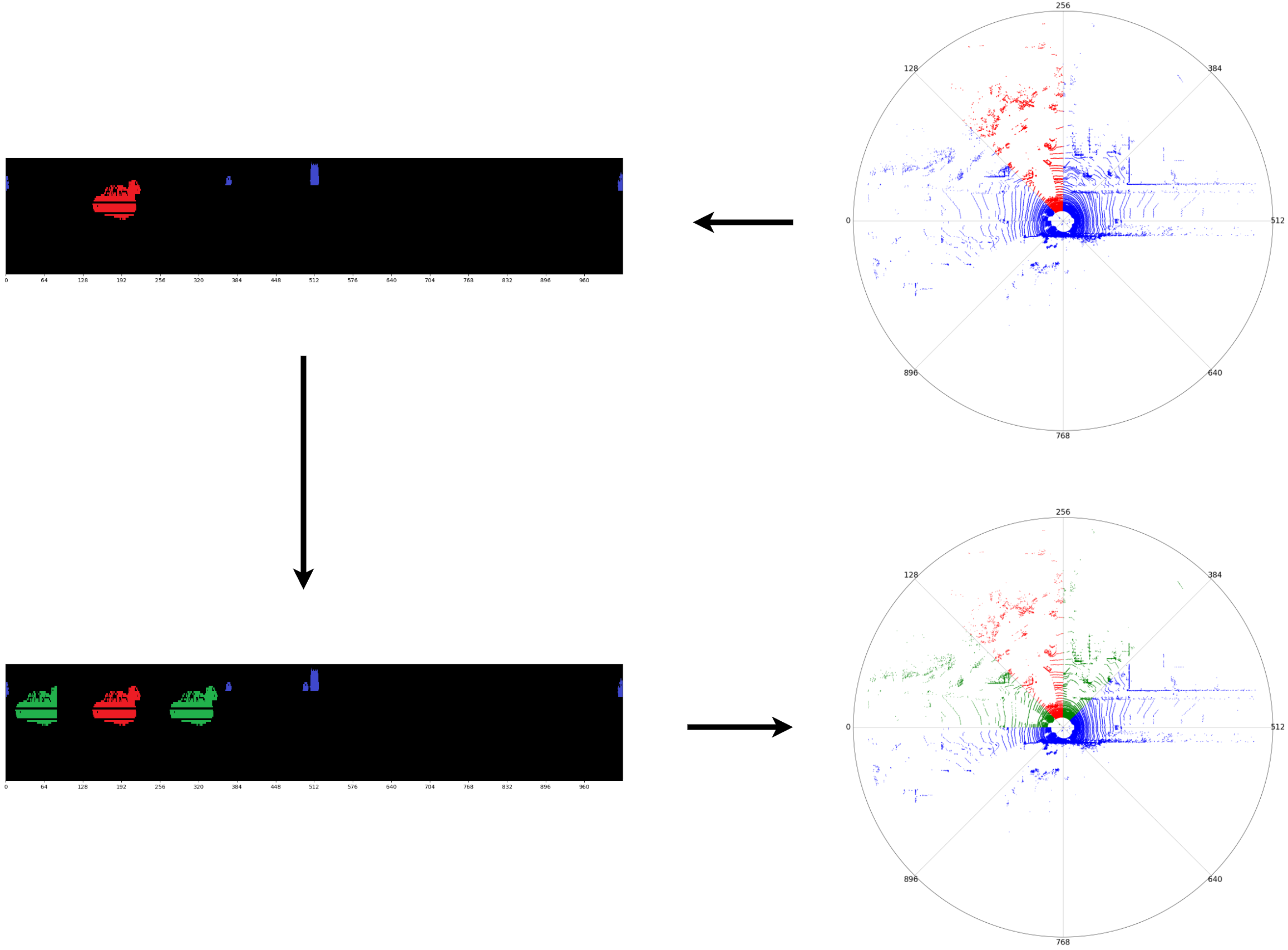}  
  \caption{View in the clockwise direction from top right. We identify the source sector (highlighted in red in the LiDAR scan) and extract the segmentation mask (also in red) in the range image mask. The source sector mask is inserted in target sectors (0 and 3 - indicated by green) in the new mask. Overlaying the new mask over the LiDAR scan augments the LiDAR with dynamic objects (in the sectors indicated by the green.)}
  \label{fig:pipeline}
\end{subfigure}
\caption{The corresponding pair pipeline results in the creation of a static-dynamic pair for KITTI. Static scan - original scan with dynamic object removed using segmentation mask. Dynamic scan - the output of the pipeline with the augmented new dynamic objects.}
\label{fig:pair_generation}
\end{figure*}

We have described the method for generating correspondence pairs for datasets already existing in the literature. Assume we have a sequence of scans $K$ = \{$k_i$: 1,2,3...\}. We generate pairs $K_D$, $K_S$ — $K_D$ is the set of dynamic, and $K_S$ denotes the static scans by utilizing semantic segmentation information. 

We divide the LiDAR scan and the range image into eight sectors. We identify a source and a target sector. Dynamic objects from the source sector are introduced into the target sector. We may have multiple target sectors depending on the availability of probable regions where new objects can be augmented. We depict the division of the LiDAR into eight sectors in Figure \ref{fig:sector_division}. We further demonstrate the augmentation process for a correspondence pair generation in Figure \ref{fig:pipeline}. 

We do not claim that the augmented generated dynamic LiDAR would be totally realistic and follow all rules of the road. For instance, in a certain case with two vehicles: one being the already present and the other being the augmented one in the opposite lane of the AV, both may appear to overlay over each other. However, the data is sufficient to train \glidar{} accurately. These issues are minor, and a large percentage of LiDAR frames are augmented with realistic dynamism at realistic locations. 
This is also demonstrated by the effect of augmentation of static structures on SLAM performance as shown in Table \ref{tab:slam_with_seg_rpe}, \ref{tab:slam_without_seg_rpe} and Figure \ref{fig:slam_trajectory_supp}.

\section{Segmentation mask generated by \glidar{}}

We demonstrate the segmentation masks for dynamic and static objects generated by \glidar{}. We provide a video in the Supplementary material that demonstrates the accuracy of the binary segmentation mask generated by \glidar{}. We remove the ground points from the LiDAR before generating the mask. Ground points are not useful for navigation and are also removed by SLAM algorithms before navigation. We classify all the above-ground points into static and dynamic objects. Dynamic objects - whether moving or stationary are marked with blue, while static objects are marked with red.
\section{Baselines}
\textbf{CP3} \citet{xu2023cp3} present an architecture for shape completion which is inspired by the prompting methods in NLP. The approach involves a Pretrain-Prompt-Predict paradigm, treating point cloud generation and refinement as the prompt and predict steps. They use IOI pretraining to achieve higher robustness in point cloud generation by learning a pretext task in a self-supervised manner. The authors also introduce a novel Semantic Conditional Refinement (SCR) at the prediction step to discriminatively modulate refinement with semantic guidance.

\textbf{Coase Net} \citet{wang2022coarse} is a two-stage approach for dynamic to static image translation applicable for 2D images. The authors formulate the problem as an image inpainting problem and aim to mitigate the loss of detail in the original static reconstructions. In the initial stage, a basic encoder-decoder network is used to generate a coarse representation of the static image, with which dynamic objects and their shadows are identified. In the second stage, the missing static pixels in the estimated dynamic regions are recovered based on their coarse predictions, which are improved by a mutual texture-structure attention module, allowing dynamic regions to incorporate textures and structures from distant areas with similar contexts.

\textbf{Topological Autoencoders} \citet{pmlr-v119-moor20a} introduces an approach of constraining autoencoders to preserve topological structures from the input space in the latent space of the autoencoder. The authors employ persistent homology (with the Vietoris-Rips complex) to formulate a differentiable loss function considering the topological signatures of both input and latent space. This approximation of the persistent homology loss calculations is combined with backpropagation on the level of mini-batches. 

\textbf{DSLR} \citet{kumar2021dynamic} tackles the challenge of translating dynamic LiDAR data into static representations using range images. They employ an adversarial training approach involving an autoencoder consisting of a pre-trained generator and a discriminator network to produce static reconstructions of dynamic scans. This approach is further extended to work in a real-world setting using Unsupervised Domain Adaptation. Furthermore, the authors also introduce a variation that enhances the quality of reconstruction by integrating segmentation information into the process.

\textbf{MOVES} \citet{kumar2023movese} employs a GAN-based adversarial model that segments out movable and moving objects in LiDAR scans. The network consists of a generator and a discriminator, that is coupled with a contrastive loss on a LiDAR scan triplet using hard-negative mining. The authors also introduce an approach that integrates Unsupervised Domain Adaptation into the network for datasets lacking static-dynamic correspondence. This method integrates the domain discrepancy loss between paired and unpaired latent space domains.

\section{LiDAR Reconstruction Evaluation Metrics}
We evaluate the difference between model reconstructed static and ground truth static LiDAR scan using the following metrics.
\begin{itemize}
    \item \textbf{CD:} Chamfer Distance tries to capture the average mismatch between points in two given point clouds $S_i, \overline{S}_i \in \mathbb{R}^3$ and is given by:
    \[ d_{CD}(S_i, \overline{S}_i) = \sum_{x \in S_i}\min_{y \in \overline{S}_i}||x-y||_2^2 + \sum_{y \in \overline{S}_i}\min_{x \in S_i}||x-y||_2^2\]
    \item \textbf{JSD:} Jenson Shannon Divergence is a measure of the distance between two empirical distributions $P$ and $Q$, defined as follows.
    \[JSD(P||Q) = \frac{1}{2}(D(P||M)+D(Q||M))\]
    where $M = \frac{1}{2}(P+Q)$ and $D(\cdot||\cdot)$ represents the Kullback-Leibler-divergence between the two marginal distributions 
    \item \textbf{MMD:} Suppose $x\in S_i$ and $y\in \overline{S}_i$ and $\phi $ is a function used to map the data to a Reproducing Kernel Hilbert Space (RKHS). The Minimal Matching Distance is approximated as follows.
    \[d_{MMD}(S_i, \overline{S}_i) = \left|\left|\frac{1}{|S_i|} \sum_{x\in S_i}\phi(x) - \frac{1}{|\overline{S}_i|}\sum_{y\in\overline{S}_i}\phi(y)\right|\right|\]
    \item \textbf{RMSE:} The Root Mean Squared Error between $S_i$ and $\overline{S}_i$ is given by
    \[RMSE(S_i, \overline{S}_i) = \left(MSE(S_i, \overline{S}_i)\right)^{\frac{1}{2}}\] where \(MSE(S_i, \overline{S}_i) = \sum\limits_{x \in S_i}\frac{(x-y)^2}{|S_i|}\) and $y\in \overline{S}_i$
    \item \textbf{EMD:} If $S_i, \overline{S}_i \in \mathbb{R}^3$ have the same size, i.e., $s=|S_i|=|\overline{S}_i|$ the Earth Mover Distance between the two point clouds is defined as follows.
    \[d_{EMD}(S_i,\overline{S}_i) = \min_{\phi:S_i \rightarrow \overline{S}_i} \sum_{x \in S_i} ||x-\phi(x)||_2\]
    where $\phi:S_i\rightarrow\overline{S}_i$ is a bijection. 
    
\end{itemize}

\section{Navigation Results using SLAM}




\subsection{Experimental Setup}
We use Google Cartographer \cite{hess2016real}, a widely known LiDAR-based SLAM algorithm, to test the LiDAR sequences for navigation. We use an Intel Core i5 processor with 16 GB RAM running ROS Noetic Distribution. 

We also evaluate our work against a recent, popular LiDAR-based SLAM algorithm - KISS-ICP \cite{vizzo2023kiss}. We show the results in segmentation-devoid baseline (Column 4) and the segmentation assisted  baseline (Column 5) in Table \ref{tab:metrics}. We provide the translation error(ATE) on a set of seven sparse version of seven KITTI sequences in Table \ref{tab:metrics}. These sequences have significant variation across the scenes they navigate and are longer.

\subsection{Datasets} 
\textbf{(a)} KITTI Odometry dataset \cite{geiger2012we} is a 64-beam LiDAR dataset. It has 11 sequences with ground truth poses. For SLAM, we test \glidar{} on all the ten sequences.\\
\textbf{(b)} ARD-16 is a 16-beam industrial dataset collected using an UGV. We follow the SLAM protocol mentioned at \citet{dslr-git} and test on the available single SLAM sequence.\\
\textbf{(c)} CARLA-64 is an extensive simulated 64-beam urban dataset with correspondence information. Is has four available SLAM sequnences with ground truth poses. We use these sequences for navigation.

\subsubsection{Baselines}
We use the augmented points generated by \glidar{} along the static LiDAR backbone to assist navigation. We study the performance of \glidar{} against segmentation-assisted and devoid baselines in sparse and dense settings. 

\glidar{} always works in label-devoid settings. To show its benefits over segmentation-assisted navigation, we use segmentation labels to remove dynamic objects from KITTI sequences before using them for SLAM. We call this baseline - KITTI-Seg. We compare the SLAM performance of KITTI-Seg with \glidar{} in Table \ref{tab:slam_with_seg_rpe}. For segmentation-free settings, we use the best baseline from Table 1 in the main paper for SLAM and compare the results against \glidar{} in Table \ref{tab:slam_without_seg_rpe}. We also evaluate \glidar{} against the original dynamic LiDAR scans on SLAM  in Table \ref{tab:slam_without_seg_rpe} and Section \ref{sec:comp_seg_dev_bas}.

\subsection{SLAM Evaluation Metrics}
We evaluate the difference between pose estimates generated using model-reconstructed static and ground truth static LiDAR scans using the following metrics.
\begin{itemize}
    \item \textbf{ATE:} Absolute Trajectory Error is used for measuring the overall global consistency between two trajectories by measuring the difference between the translation components of the two trajectories. The two trajectories specified in arbitrary coordinate frames are aligned in closed form.
    Subsequently, the absolute trajectory error matrix at time $i$ is defined as $\mathbf{E}_i := \mathbf{Q}_i^{-1}\mathbf{SP}_i,$ where $\mathbf{S}$ is the rigid-body transformation using the least squares solution that maps $\mathbf{P}$ (the estimated trajectory) onto $\mathbf{Q}$ (the ground truth trajectory). 
    The ATE is defined as the root mean square error from error matrices over all indices of time.
    \[ATE_{RMSE} = \left(\frac{1}{n} \sum_{i=1}^n ||trans(\mathbf{E}_i)||^2\right)^{\frac{1}{2}}\]
    where $trans(E_i)$ refers to the translational components of the absolute trajectory error matrix $E_i$
    
    \item \textbf{RPE:} Relative Pose Error, similar to the ATE, measures the pose error between estimated and ground truth trajectories. It quantifies the local accuracy of the trajectory over a fixed time interval $\delta$. The relative pose error at time step $i$ is defined as follows.
    \[\mathbf{E}_i = (\mathbf{P}_i^{-1}\mathbf{P}_{i+\delta})^{-1}(\mathbf{Q}_i^{-1}\mathbf{Q}_{i+\delta})\]
    The RMSE over all time indices is computed for the translational and rotational components as follows.
    \[RPE_{trans} = \left(\frac{1}{m} \sum_{i=1}^m ||trans(\mathbf{E}_i)||^2\right)^{\frac{1}{2}}\]
    \[RPE_{rot} = \left(\frac{1}{m} \sum_{i=1}^m ||rot(\mathbf{E}_i)||^2\right)^{\frac{1}{2}}\]
    For further information on these metrics, please refer to Sturm et al., \cite{sturm2012benchmark}

\end{itemize}

\subsubsection{Comparison with Segmentation-based Baselines}

\begin{table*}[htbp]
\setlength{\tabcolsep}{3.5pt}
    \scriptsize{
\begin{tabular}{c c  c c c c c c c c c c c c} 
\toprule
Dataset &\multirow{4}{*}{Sequence} & \multicolumn{6}{c}{\textbf{Sparse LIDAR}} & \multicolumn{6}{c}{\textbf{Dense LiDAR}}  \\ 
\cmidrule(rl){3-8} \cmidrule(l){9-14}

&&\multicolumn{2}{c}{\textbf{Original Dynamic}}& \multicolumn{2}{c}{\textbf{Segmented Out}}& \multicolumn{2}{c}{\textbf{Ours}} &\multicolumn{2}{c}{\textbf{Original Dynamic}}& \multicolumn{2}{c}{\textbf{Segmented Out}}&\multicolumn{2}{c}{\textbf{Ours}}                \\
\cmidrule(rl){3-4} \cmidrule(l){5-6} \cmidrule(l){7-8} \cmidrule(l){9-10} \cmidrule(l){11-12} \cmidrule(l){13-14}
&& RPE Trans&RPE Rot & RPE Trans&RPE Rot & RPE Trans&RPE Rot & RPE Trans&RPE Rot & RPE Trans&RPE Rot &RPE Trans&RPE Rot        \\
\hline
        \multirow{11}{*}{\sc KITTI}&0 & 1.10&\textbf{0.060} & \textbf{1.096}&\textbf{0.060} & 1.1&\textbf{0.060} & \textbf{1.10}&\textbf{0.060} & \textbf{1.100}&\textbf{0.060} & \textbf{1.10}&\textbf{0.060} \\ 
        &1 & 1.73&0.106 & 2.047&\textbf{0.039} &\textbf{ 1.56}&\textbf{0.039} & 1.595&\textbf{0.039} & 1.542&\textbf{0.039} & \textbf{1.54}&\textbf{0.039} \\ 
        &2 & \textbf{1.48}&\textbf{0.488} & 1.501&0.049 & 1.482&0.049 & 1.549&\textbf{0.049} & \textbf{1.547}&\textbf{0.049} & 1.55&\textbf{0.049 }\\ 
        &4 & 114.587&0.008 & \textbf{1.924}&0.007 & 2.06&\textbf{0.005} & \textbf{2.04}&\textbf{0.005} & 2.05&\textbf{0.005} & \textbf{2.04}&\textbf{0.00}5 \\ 
        &5 & \textbf{1.210}&\textbf{0.043} & 1.214&0.0433 &\textbf{ 1.21}&\textbf{0.043 }& \textbf{1.21}&\textbf{0.043} & 1.215&\textbf{0.043} & \textbf{1.21}&\textbf{0.043} \\ 
        &6 & 1.66&\textbf{0.052} &\textbf{ 1.643}&\textbf{0.052} & 1.66&\textbf{0.052 }& \textbf{1.655}&\textbf{0.051} & 1.670&0.052 & 1.66&\textbf{0.051} \\ 
        &7 &\textbf{ 1.04}&\textbf{0.057} & \textbf{1.040}&\textbf{0.057} & \textbf{1.04}&\textbf{0.05}7 & \textbf{1.040}&\textbf{0.0567} & \textbf{1.040}&0.057 & \textbf{1.04}&0.057 \\ 
        &8 & 53.339&\textbf{0.109} & \textbf{25.274}&0.134 & 30.85& 0.146 & \textbf{46.170}&\textbf{0.107} & 55.603&0.129 & 60.09&0.112 \\ 
        &9 & \textbf{1.768}&\textbf{0.046} & \textbf{1.768}&\textbf{0.046 }& 1.77&\textbf{0.04}6 & \textbf{1.773}&\textbf{0.046} & \textbf{1.773}&\textbf{0.046} & 1.78&0.048 \\ 
        &10 & 1.366&\textbf{0.450} & 1.363&\textbf{0.045} &\textbf{ 1.36}&\textbf{0.045} & \textbf{1.355}&\textbf{0.04488} & 1.363&0.045 & 1.36&0.045 \\ \hline
        
        \multirow{4}{*}{\sc CARLA-64}&
        0 & \textbf{0.099}&0.451 & - & - & \textbf{0.099}&\textbf{0.446} & \textbf{0.103}&\textbf{0.414} & - & - & 0.083&0.438 \\ 
        
        &1 & 0.068&\textbf{0.408} & - & - & \textbf{0.057}&0.410 & \textbf{0.051}& 0.451 & - & -  & 0.056&\textbf{0.406} \\ 
        &2 & 0.100&\textbf{0.547} & - & - & \textbf{0.099}&0.55 & 0.078&\textbf{0.527 }&  - & - & \textbf{0.066}&0.557 \\ 
        
        &3 & \textbf{0.079}&\textbf{0.204} & - & - & 0.117&0.395 & \textbf{0.137}&\textbf{0.375 }& - & - & 0.212&0.380 \\ \hline
 
        \sc ARD-16&0 & \textbf{0.166}&5.07 & - & - &0.171 & \textbf{4.955}& 0.178 & \textbf{4.878} &  - & -  & \textbf{0.171} & 4.934 \\ \hline
\end{tabular}
}
\caption{\glidar{} comparison against segmentation-assisted baseline on KITTI-Seg and against original dynamic LiDAR sequence based on the RPE metric.}
\label{tab:slam_with_seg_rpe}
\end{table*}

\begin{table}[htbp]
\centering
\footnotesize

\begin{tabular}{lccccc}
\toprule
\textbf{Seq} & \textbf{Length} & \glidar{} & \textbf{Best Baseline} & \textbf{KITTI-Seg} & \textbf{Original} \\
\midrule
0 & 4390& \textbf{7.53} & 190.67 & 7.98 & 7.7 \\
2 & 4485& \textbf{12.17} & 292.65 & 13.57 & 12.42 \\
5 & 2750& \textbf{4.27} & 152.03 & 4.41 & 4.41 \\
6 & 1091& \textbf{1.65} & 126.23 & 1.67 & 1.67 \\
7 & 1111&0.98 & 85.83 & 0.9 & \textbf{0.7} \\
8 & 5149&\textbf{4.19} & 188.5 & 4.21 & 4.31 \\
9 & 1599&\textbf{4.45} & 212.24 & 4.82 & 4.48 \\
\bottomrule
\end{tabular}
\caption{ATE numbers with KISS-ICP SLAM algorithm}
\label{tab:metrics}
\end{table}

\label{sec:comp_seg_baseline}
 KITTI has segmentation information available. We compare \glidar{} against the segmentation-based KITTI baseline - KITTI-Seg. It uses segmentation information to remove dynamic points from LiDAR sequences before navigation. We also compare \glidar{} against the original dynamic LiDAR sequences of all three datasets in Table \ref{tab:slam_with_seg_rpe} and Figure  \ref{fig:slam_trajectory_supp}.
\glidar{} performs better than both methods by a fair margin without the assistance of segmentation labels for almost every sequence of KITTI and CARLA-64. Our trajectory estimates are on par with or better than KITTI-Seg as shown in Figure \ref{fig:slam_trajectory_supp}. The results demonstrate \glidar{}'s capability to reinforce accurate points along existing static structures as well as newer points along occluded static structures by simply following the $0$-dim $\mathcal{PH}$ based static backbone. It strengthens our premise that \textbf{(a)} augmenting static structures - both visible and occluded with newer points is superior to existing pre-processing approaches for SLAM in sparse LiDAR settings and
\textbf{(b)} $0$-dim $\mathcal{PH}$ based constraint and the LiDAR graph representation are highly effective at preserving global shape of LiDAR topology while generating newer static points with high precision.

We notice that for certain sequences, KITTI-Seg has higher navigation errors compared to the original dynamic sequences. Our investigation yields the following reasons. These sequences consist of certain dynamic objects that are stationary and help navigation. KITTI-Seg removes all dynamic objects, which may also include objects that are movable but stationary in the sequence. In such cases, it is devoid of several stationary points that are available in the original dynamic sequence. In these scenarios, KITTI-Seg performs inferior to the original dynamic scans.

We notice that for sequences 6,7,9 and 10 in the dense settings, \glidar{} performs equivalent to but not better than the two settings. We observe that in dense settings, augmenting existing static structures that are already dense has no major impact on navigation. However, augmenting occluded static structures along the static backbone brings newer points that are not available in the original scan and can help navigation. We observe that sequences 6, 7, 9, and 10 have very few dynamic occlusions (and hence fewer occluded static structures - 8, 15, 13, 10, respectively), resulting in negligible effect on navigation performance.

A natural question arises: What levels of sparsity can \glidar{} handle? We perform several experiments similar to Table \ref{tab:slam_with_seg_rpe} for KITTI with sparser LiDAR scans - 8-beam  and 4-beam scans. The results of the experiments are available in Table \ref{tab:sparser-navig} and Section \ref{sec:sparse_exp}. Our investigations suggest that high sparsity levels in the above cases destroy the LiDAR topology to the extent that $0$-dim $\mathcal{PH}$ can no longer recover an accurate backbone of the LiDAR scene. Due to this, \glidar{} navigation results perform poorly against the original dynamic and the  KITTI-Seg baseline. \glidar{} does not perform well with such sparsity.

\subsubsection{Comparison with Segmentation Devoid Baseline}

\begin{table*}[htbp]
\setlength{\tabcolsep}{10pt}
\scriptsize{
\begin{tabular}{c c c c c c c c c c} 
\toprule
\multirow{3}{*}{Dataset}& \multirow{3}{*}{Sequence} & \multicolumn{4}{c}{\textbf{Sparse LIDAR}} & \multicolumn{4}{c}{\textbf{Dense LiDAR}}  \\ 
\cmidrule(rl){3-6} \cmidrule(l){7-10}
& & \multicolumn{2}{c}{\textbf{Best Baseline}}& \multicolumn{2}{c}{\textbf{Ours}}& \multicolumn{2}{c}{\textbf{Best Baseline}} &\multicolumn{2}{c}{\textbf{Ours}}               \\
\cmidrule(rl){3-4} \cmidrule(l){5-6} \cmidrule(l){7-8} \cmidrule(l){9-10}

& & RPE Trans&RPE Rot & RPE Trans&RPE Rot & RPE Trans&RPE Rot & RPE Trans&RPE Rot \\
\hline
        \multirow{11}{*}{\sc KITTI} & 0 & 1.17 & \textbf{0.059} & \textbf{1.1} & 0.060 & 1.10 & 0.060 & \textbf{1.10} & \textbf{0.060} \\ 
        &1 & 2.07 & \textbf{0.037} & \textbf{1.56} & 0.039 & 2.40 & 0.039 & \textbf{1.54} & \textbf{0.039} \\ 
        &2 & \textbf{1.38} & \textbf{0.048} & 1.48 & 0.049 & \textbf{1.54} & 0.049 & 1.55 & \textbf{0.049} \\ 
        &4 & \textbf{1.68} & \textbf{0.006} & 2.06 & 0.005 & 2.04 & 0.0052 & \textbf{2.04} & \textbf{0.005} \\ 
        &5 & \textbf{1.10} & 0.043 & 1.21 & \textbf{0.043} & 1.21 & 0.043 & \textbf{1.21} & \textbf{0.043} \\ 
        &6 & \textbf{1.42} & \textbf{0.051} & 1.66 & 0.0516 & \textbf{1.64} & 0.051 & 1.66 & \textbf{0.051} \\ 
        &7 & \textbf{0.93} & \textbf{0.056} & 1.04 & 0.057 & 1.04 & 0.057 & \textbf{1.04} & \textbf{0.057} \\ 
        &8 & \textbf{20.85} & \textbf{0.129} & 30.85 & 0.146 & \textbf{33.075} & \textbf{0.071} & 60.00 & 0.112 \\ 
        &9 & \textbf{1.24} & \textbf{0.045} & 1.77 & 0.046 & \textbf{1.77} & \textbf{0.046} & 1.78 & 0.048 \\ 
        &10 & \textbf{1.01} & \textbf{0.044} & 1.36 & 0.045 & \textbf{1.31} & \textbf{0.044} & 1.36 & 0.045 \\ \hline
        \multirow{4}{*}{\sc CARLA-64}&0 & \textbf{0.099} & \textbf{0.439} & \textbf{0.099} & 0.446 & 0.090 & \textbf{0.414} & \textbf{0.083} & 0.438 \\ 
        &1 & 0.156 & 0.411 & \textbf{0.057} & \textbf{0.41} & 0.065 & \textbf{0.394} & \textbf{0.056} & 0.406 \\ 
        &2 & 0.125 & \textbf{0.546} & \textbf{0.099} & 0.55 & 0.141 & \textbf{0.553} & \textbf{0.066} & 0.557 \\ 
        &3 & \textbf{0.106} & 0.397 & 0.117 & \textbf{0.395} & \textbf{0.110} & 0.399 & 0.120 & \textbf{0.393} \\ \hline
        \sc ARD-16&0 & \textbf{0.170} & \textbf{4.947} & 0.171 & 4.955 & 0.182 & \textbf{4.85} & \textbf{0.171} & 4.934 \\ \bottomrule
\end{tabular}
}
\caption{Navigation performance comparison of \glidar{} against the best baseline in segmentation-devoid settings using RPE metric in sparse and dense settings. For RPE, lower is better.}
\label{tab:slam_without_seg_rpe}
\end{table*}

\label{sec:comp_seg_dev_bas}
We compare the navigation results of \glidar{} against the best baseline for static point augmentation in Table \ref{tab:slam_without_seg_rpe}, Figure \ref{fig:slam_trajectory_supp}. \glidar{} performs extremely well and consistently outperforms the baseline by a large margin for KITTI and CARLA-64 datasets in the sparse settings. In Figure \ref{fig:slam_trajectory_supp}, we observe that for the sparse case, the best baseline (blue)  misses the ground-truth trajectory (dotted line) by a heavy margin. \glidar{} is able to navigate accurately and performs better for most sequences. \glidar{} generates consistent points along the sparse structures by following the static topology-based backbone unlike the best baseline, which in the sparse case fails to generate consistent points along static structures. The $0$-dim $\mathcal{PH}$ prior ensures that \glidar{} reinforces static structures only along the single connected component outlined by the static backbone. This allows accurate scan matching using the predicted static points, leading to better navigation performance. It also performs well against the baseline in dense settings. 

We observe that for the ARD-16 dataset (which only works in segmentation devoid settings), \glidar{} performs comparable but not better than the original dynamic scan for navigation in Table \ref{tab:slam_without_seg_rpe}. The reasons lie in the semantics and the structure of the ARD-16 dataset. The improvement in navigation for ARD-16 is marginal in both settings. ARD-16 is collected in a closed industrial environment, unlike KITTI and CARLA-64. Every LiDAR scan in the sequence observes a large part of the navigation environment. Unlike urban settings, the collected LiDAR scans overlap significantly, even at turns. Consecutive scans have sufficient static points for accurate scan matching despite sparsity, even in the absence of augmented static points. The static points introduced by \glidar{} do not provide any significant improvement to navigation performance. However, the newly introduced static points along occluded static structures help to generate accurate binary segmentation masks (Section 6.1.1 in the main paper), which are vital for safe navigation.

\begin{table*}
\setlength{\tabcolsep}{9.5pt}
\scriptsize{
\begin{tabular}{c c c c c c c c} 
\toprule
 &
  \multicolumn{3}{c}{{\textbf{8-beam}}} &
  \multicolumn{3}{c}{{\textbf{4-beam}}} &
  \multicolumn{1}{c}{{\textbf{}}} \\
  \cmidrule(rl){2-4} \cmidrule(rl){5-7}
 \multicolumn{1}{c}{{\textbf{Sequence}}} &
  \textbf{Original Dynamic} &
  \textbf{Segmented Out} &
  \textbf{Ours} &
  \textbf{Original Dynamic} &
  \textbf{Segmented Out} &
  \textbf{Ours} &
   \\
   \cmidrule(rl){2-2}\cmidrule(rl){3-3} \cmidrule(rl){4-4} \cmidrule(rl){5-5} \cmidrule(rl){6-6} \cmidrule(rl){7-7}
\multicolumn{1}{l}{} &
  \multicolumn{3}{c}{\textbf{ATE/RPE Trans/RPE Rot}} &
  \multicolumn{3}{c}{\textbf{ATE/RPE Trans/RPE Rot}}
   \\
   \hline
0 &
  11.08/\textbf{1.10}/\textbf{0.06} &
  \textbf{10.98}/\textbf{1.10}/\textbf{0.06} &
  100.61/\textbf{1.10}/\textbf{0.06} &
  23.46/1.09/\textbf{0.06} &
  \textbf{21.63}/1.09/\textbf{0.06} &
  183.36/\textbf{1.01}/\textbf{0.06} &
   \\
1 &
  440.63/2.30/\textbf{0.04} &
  \textbf{290.35}/\textbf{2.00}/\textbf{0.04} &
  732.18/2.82/\textbf{0.04} &
  \textbf{715.32}/3.07/0.04 &
  718.71/3.18/0.04 &
  734.16/\textbf{2.89}/\textbf{0.03} &
   \\
2 &
  113.12/1.75/\textbf{0.05} &
  \textbf{29.84}/1.52/\textbf{0.05} &
  223.09/\textbf{1.40}/\textbf{0.05} &
  273.32/1.67/\textbf{0.05} &
  \textbf{252.56}/1.65/\textbf{0.05} &
  303.64/\textbf{1.34}/\textbf{0.05} &
   \\
4 &
  117.61/2.26/\textbf{0.01} &
  \textbf{117.38}/2.29/\textbf{0.01} &
  119.44/\textbf{2.03}/\textbf{0.01} &
  \textbf{117.41}/2.20/0.01 &
  {118.31/\textbf{2.15}/0.01} &
  118.62/2.42/\textbf{0.003} &
   \\
5 &
  \textbf{3.15}/1.21/\textbf{0.04} &
  4.69/1.21/\textbf{0.04} &
  150.11/\textbf{0.97}/\textbf{0.04} &
  12.16/1.21/\textbf{0.04} &
  \textbf{12.12}/1.20/\textbf{0.04} &
  162.72/\textbf{1.02}/\textbf{0.04} &
   \\
6 &
  22.07/1.66/\textbf{0.05} &
  \textbf{13.25}/1.66/\textbf{0.05} &
  132.87/\textbf{1.38}/\textbf{0.05} &
  80.59/1.66/\textbf{0.05} &
  \textbf{61.30}/1.70/\textbf{0.05} &
  134.75/\textbf{1.54}/\textbf{0.05} &
   \\
7 &
  2.80/1.04/\textbf{0.06} &
  \textbf{2.21}/ 1.04/ \textbf{0.06} &
  4.95/\textbf{1.03}/\textbf{0.06} &
  2.04/1.03/\textbf{0.06} &
  \textbf{1.59}/1.03/\textbf{0.06} &
  81.92/\textbf{0.87}/\textbf{0.06} &
   \\
8 &
\textbf{106.79}/44.14/\textbf{0.09} &
  157.50/43.88/\textbf{0.09} &
  202.36/\textbf{13.22}/0.15 &
  195.29/39.43/0.14 &
\textbf{164.25}/40.34/0.18 &
200.01/\textbf{21.36}/\textbf{0.012} &
   \\
9 &
  13.19/1.76/\textbf{0.05} &
  \textbf{12.51}/1.75/\textbf{0.05} &
  15.22/\textbf{1.67}/\textbf{0.05} &
  \textbf{20.50}/\textbf{1.70}/\textbf{0.05} &
  32.83/\textbf{1.70}/\textbf{0.05} &
  50.0/1.74/\textbf{0.05} \\
  \bottomrule
\end{tabular}
}
\caption{GLiDR comparison against a segmentation-assisted baseline - KITTI-Seg and against original dynamic LiDAR sequence for 8 beam and 4 beam LiDAR scans. Lower is better. \glidar{} does not perform better for these sparsity levels.}
\label{tab:sparser-navig}
\end{table*}


\begin{figure*}
\setlength{\tabcolsep}{1pt}
\begin{tabular}{c c c} 
\includegraphics[width=0.34\linewidth]{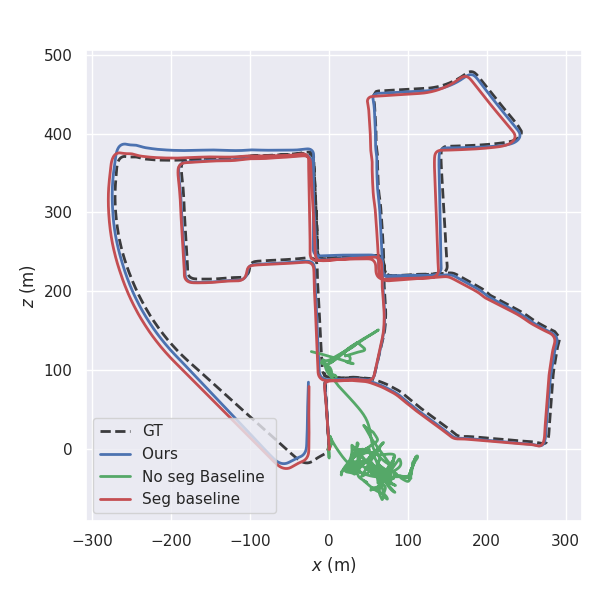} & \includegraphics[width=0.34\linewidth]{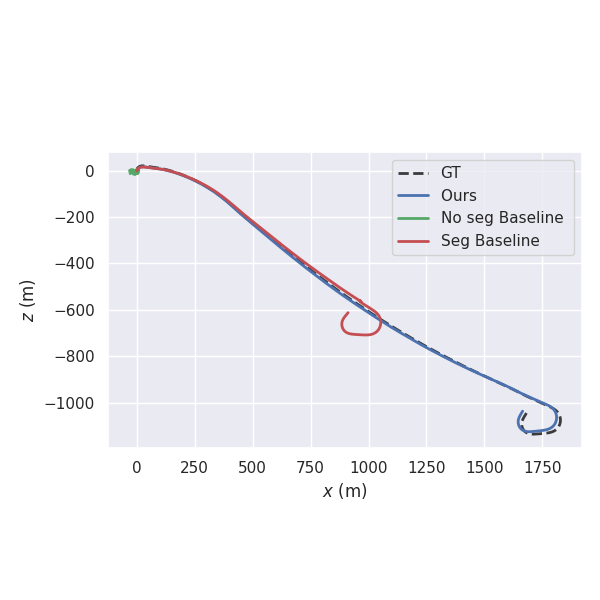} & \includegraphics[width=0.34\linewidth]{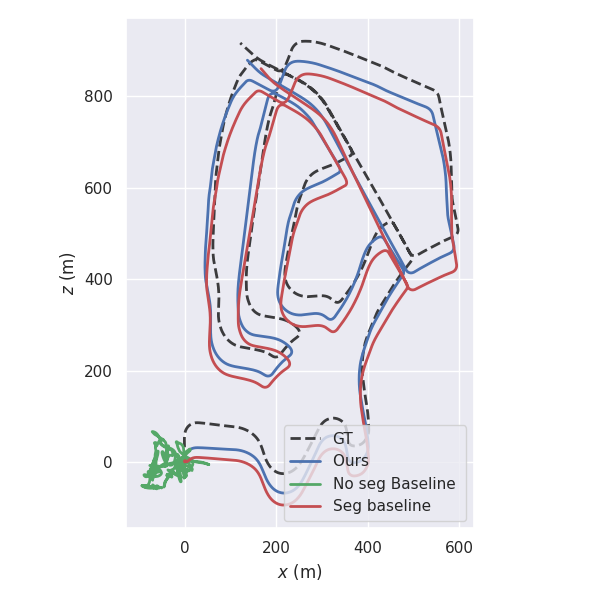}  \\
Sequence 0 & Sequence 1 & Sequence 2 \\
\includegraphics[width=0.34\linewidth]{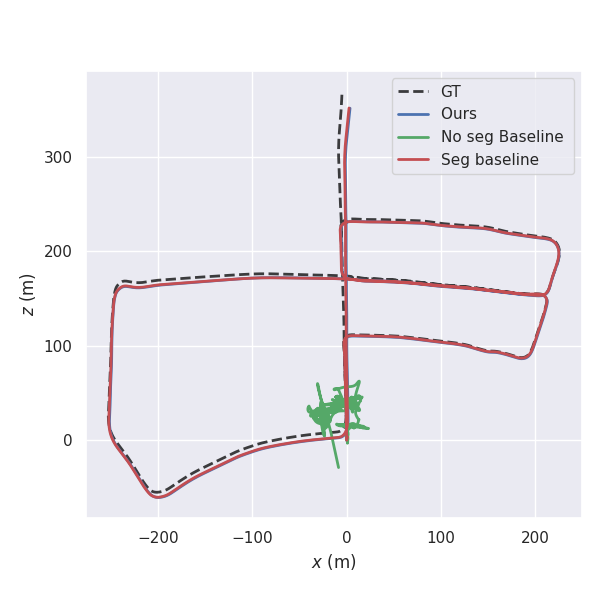} & \includegraphics[width=0.34\linewidth]{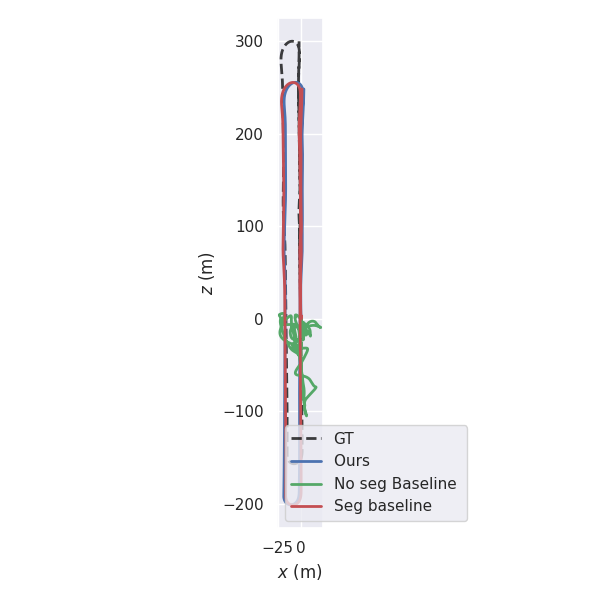} & \includegraphics[width=0.34\linewidth]{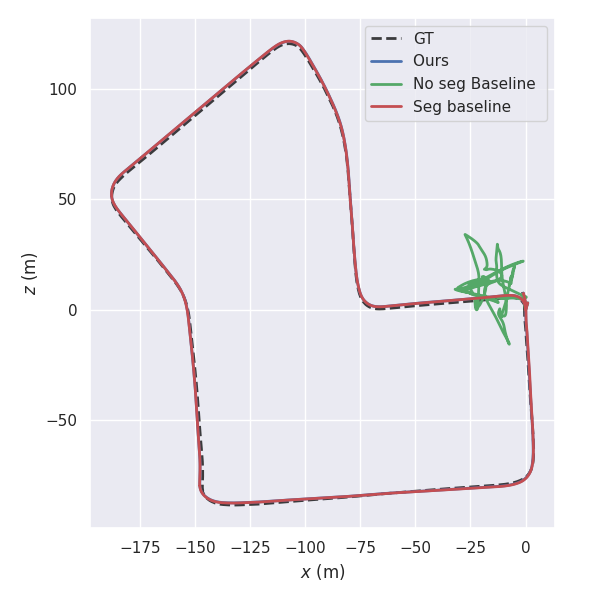}  \\
Sequence 5 & Sequence 6 & Sequence 7 \\
\end{tabular}
\caption{SLAM Trajectory comparison of \glidar{} for sparse LiDAR scans against segmentation assisted and segmentation devoid baseline.}
\label{fig:slam_trajectory_supp}
\end{figure*}

\subsubsection{Experiments on more sparse datasets}
\label{sec:sparse_exp}
What level of sparsity can be handled by \glidar{}? To answer this question, we perform experiments using more sparse LiDAR scans -  8 and 4-beam LiDAR dataset and test them on navigation. We perform experiments with the same baselines as Table 3 in the main paper - Original Dynamic, KITTI-Seg. Original Dynamic is the LiDAR range image without any pre-processing. KITTI-Seg uses segmentation labels to remove dynamic objects from the LiDAR range image before using them for navigation. The experiments are performed using the KITTI dataset. The navigation results are available in Table \ref{tab:sparser-navig}. 

We observe that, unlike 64 and 16 beam cases, for the more sparse cases (8 and 4-beam), \glidar{} does not perform well on the ATE metric for most of the sequences. ATE is a strong measure of the global consistency of the predicted trajectory with the groundtruth. We conclude that with sparsity beyond 16 beam LiDAR scans, \glidar{} does not outperform the existing baselines. Our investigation suggests that for 8 and 4-beam LiDAR datasets, the 0-dim $\mathcal{PH}$ based backbone is not able to calculate an accurate global backbone and misses many details that were captured in the 16-beam datasets. We demonstrate these findings in Table \ref{tab:backbone_1} and \ref{tab:backbone_2}. In the figures, column 1 denotes the original LiDAR, while 
Column 2, 3, and 4 denotes the 0-dim $\mathcal{PH}$ backbone for 16, 8, and 4 beam LiDAR sparse LiDAR scans, respectively. We observe that while the 16-beam LiDAR backbone captures the global outline of the static structures accurately, the 8 and 4-beam LiDAR miss a lot of details and are unable to capture the global static LiDAR structure accurately. This results in insufficient and sub-optimal augmentation of static points, which translates into inferior navigation performance for 8 and 4 LiDAR beam based scans.

\begin{table*}
\centering
\setlength{\tabcolsep}{7pt}
\begin{tabular}{c c c c}
 \frame{\includegraphics[width=0.20\linewidth, trim=5cm 7cm 3cm 9cm,clip]{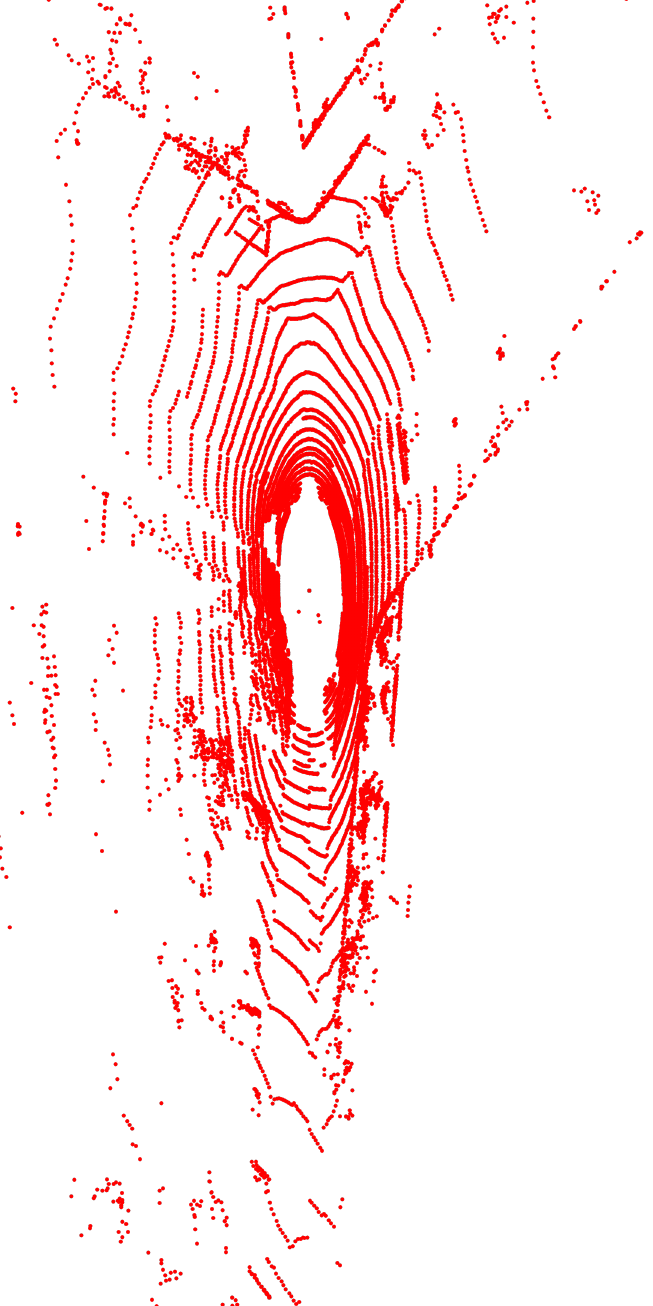}}
 &\frame{\includegraphics[width=0.20\linewidth, trim=5cm 7cm 3cm 9cm,clip]{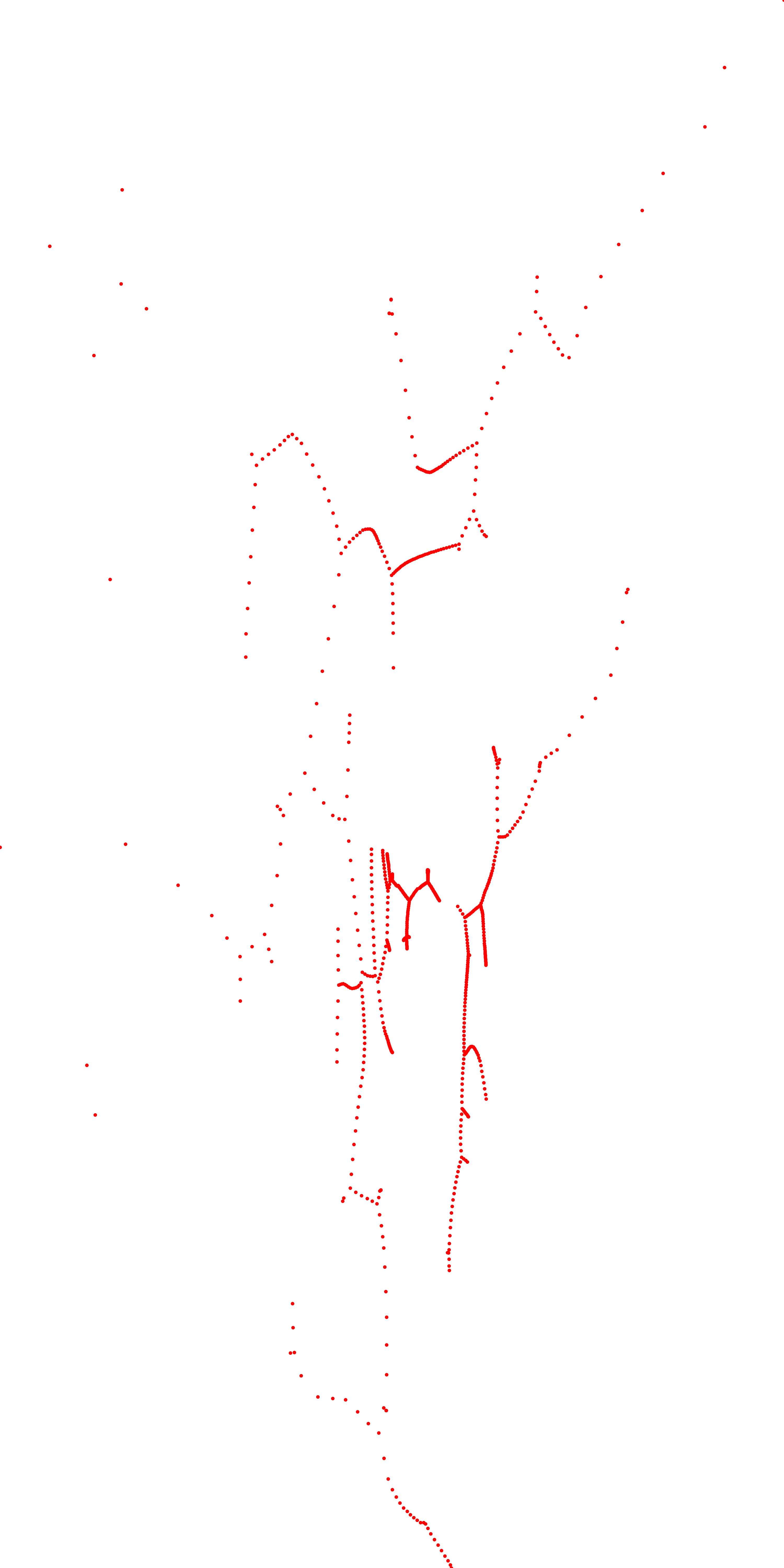}}
 & \frame{\includegraphics[width=0.20\linewidth, trim=5cm 7cm 3cm 9cm,clip]{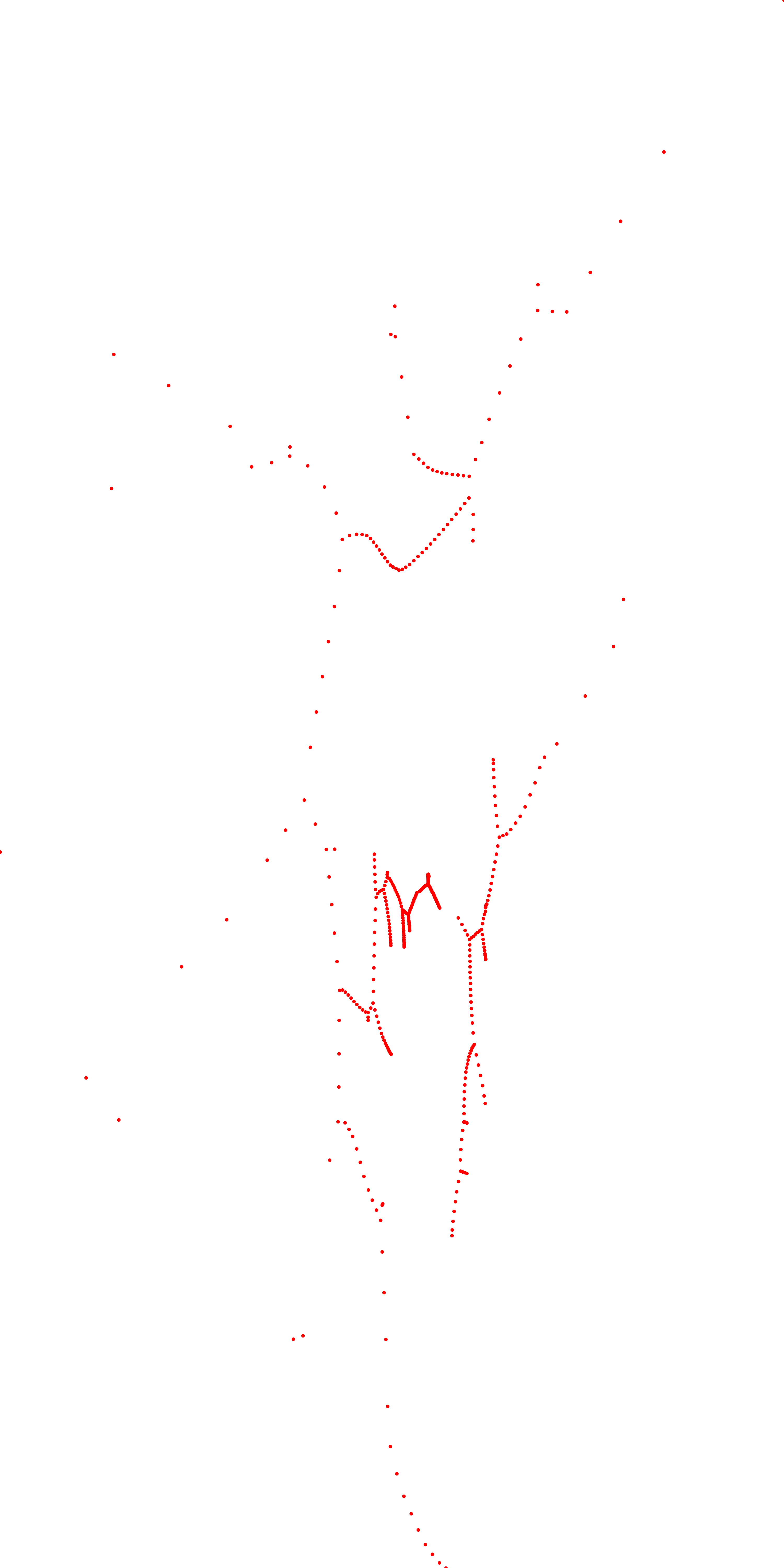}}
 & \frame{\includegraphics[width=0.20\linewidth, trim=5cm 7cm 3cm  9cm,clip]{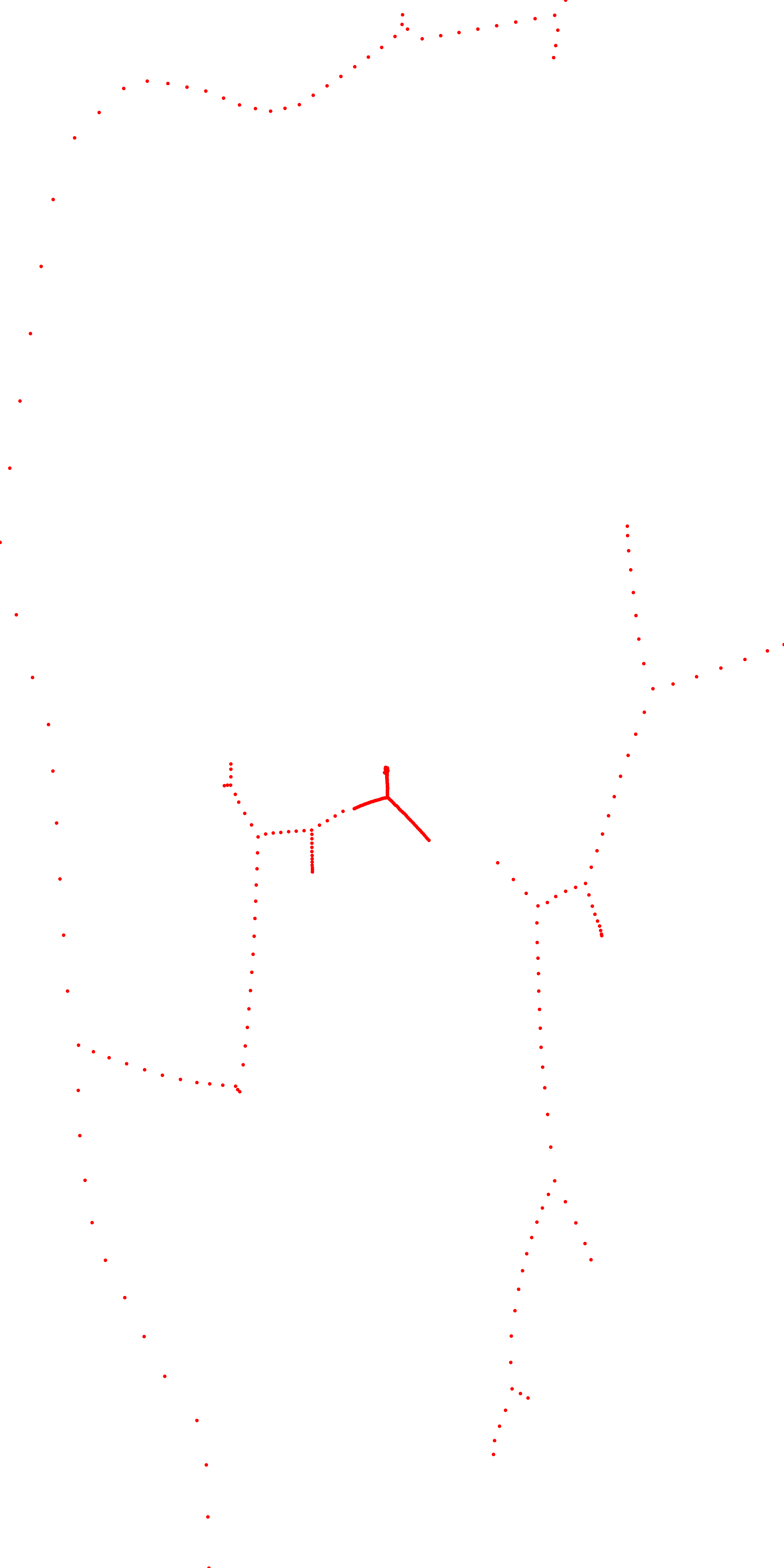}} \\ 

\frame{\includegraphics[width=0.20\linewidth, trim=5cm 7cm 3cm  9cm,clip]{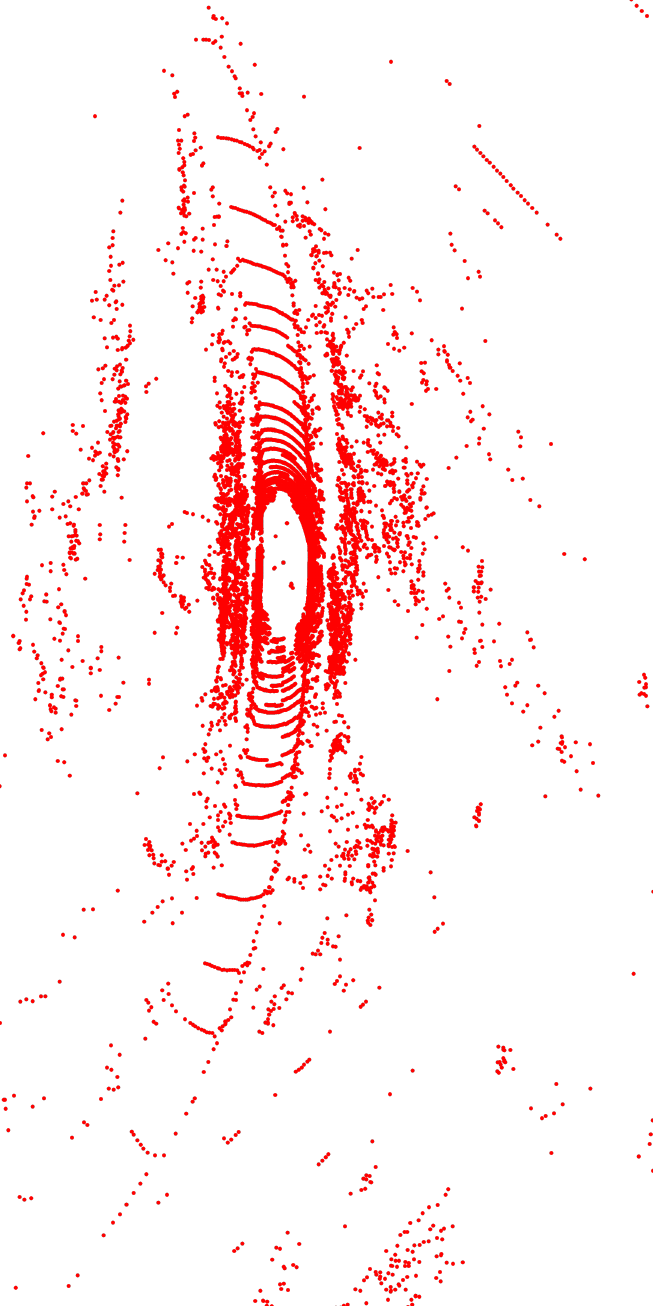}}& \frame{\includegraphics[width=0.20\linewidth, trim=5cm 7cm 3cm  9cm,clip]{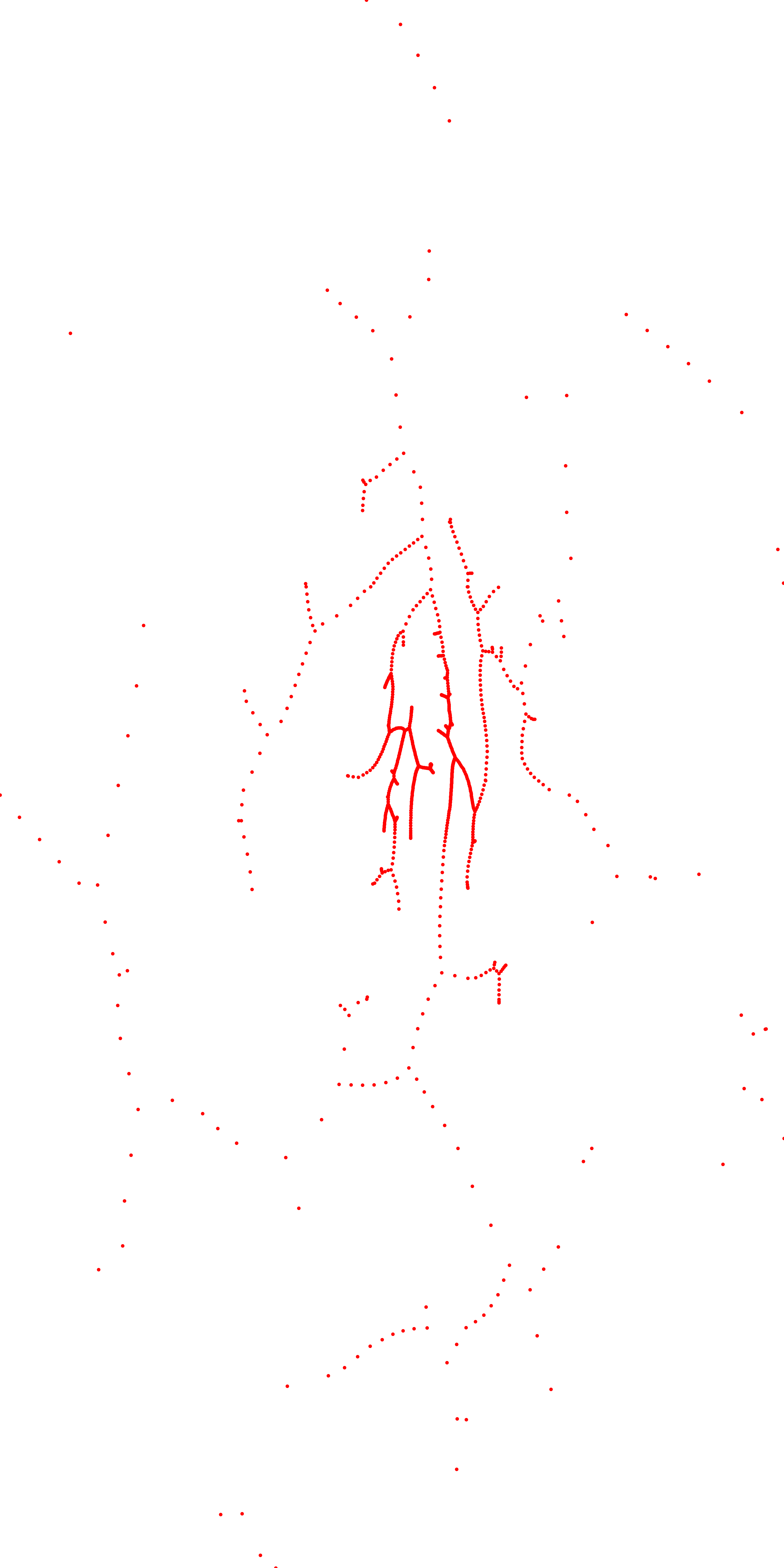}}& \frame{\includegraphics[width=0.20\linewidth, trim=5cm 7cm 3cm  9cm,clip]{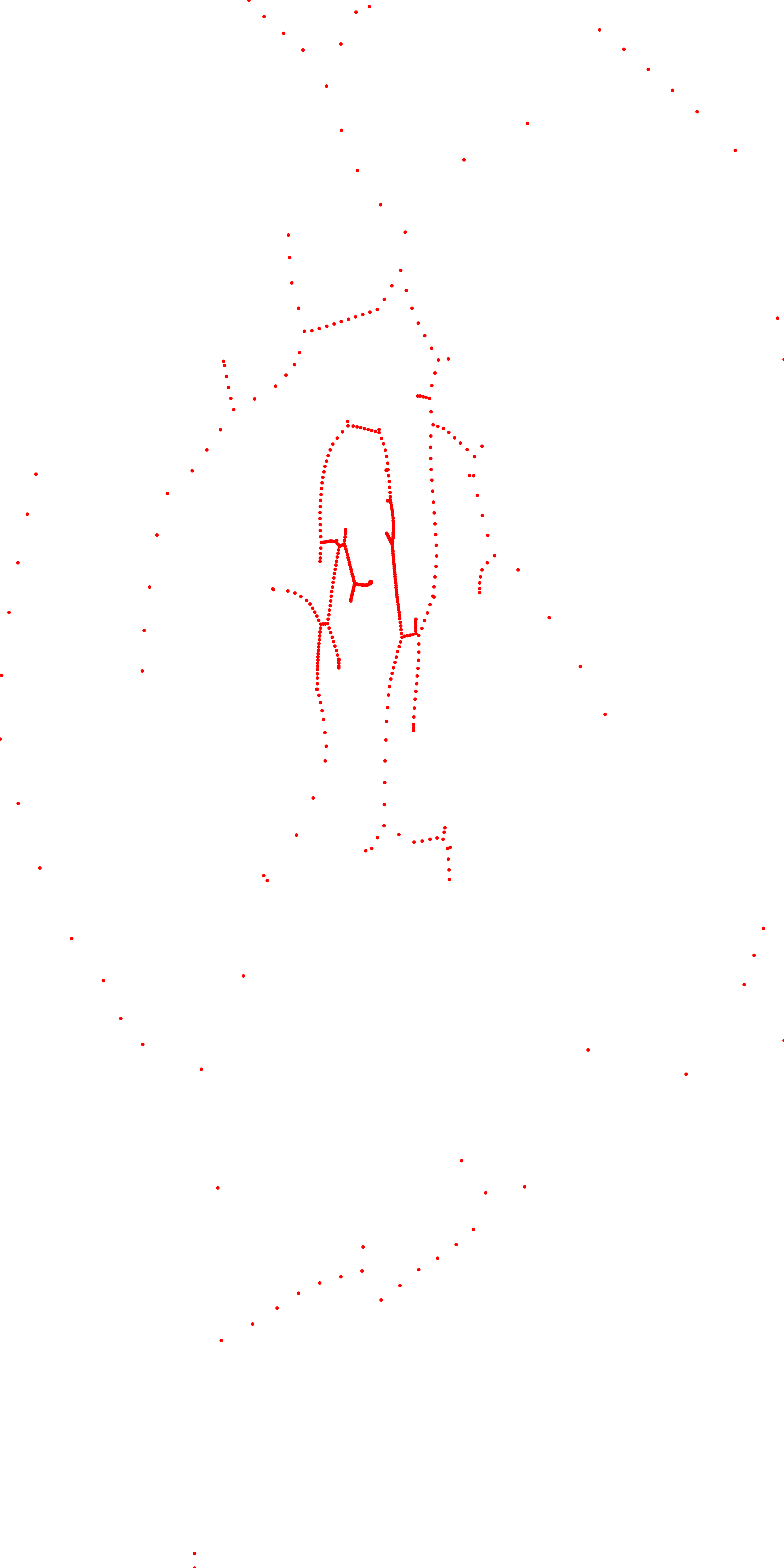}}& \frame{\includegraphics[width=0.20\linewidth, trim=5cm 7cm 3cm  9cm,clip]{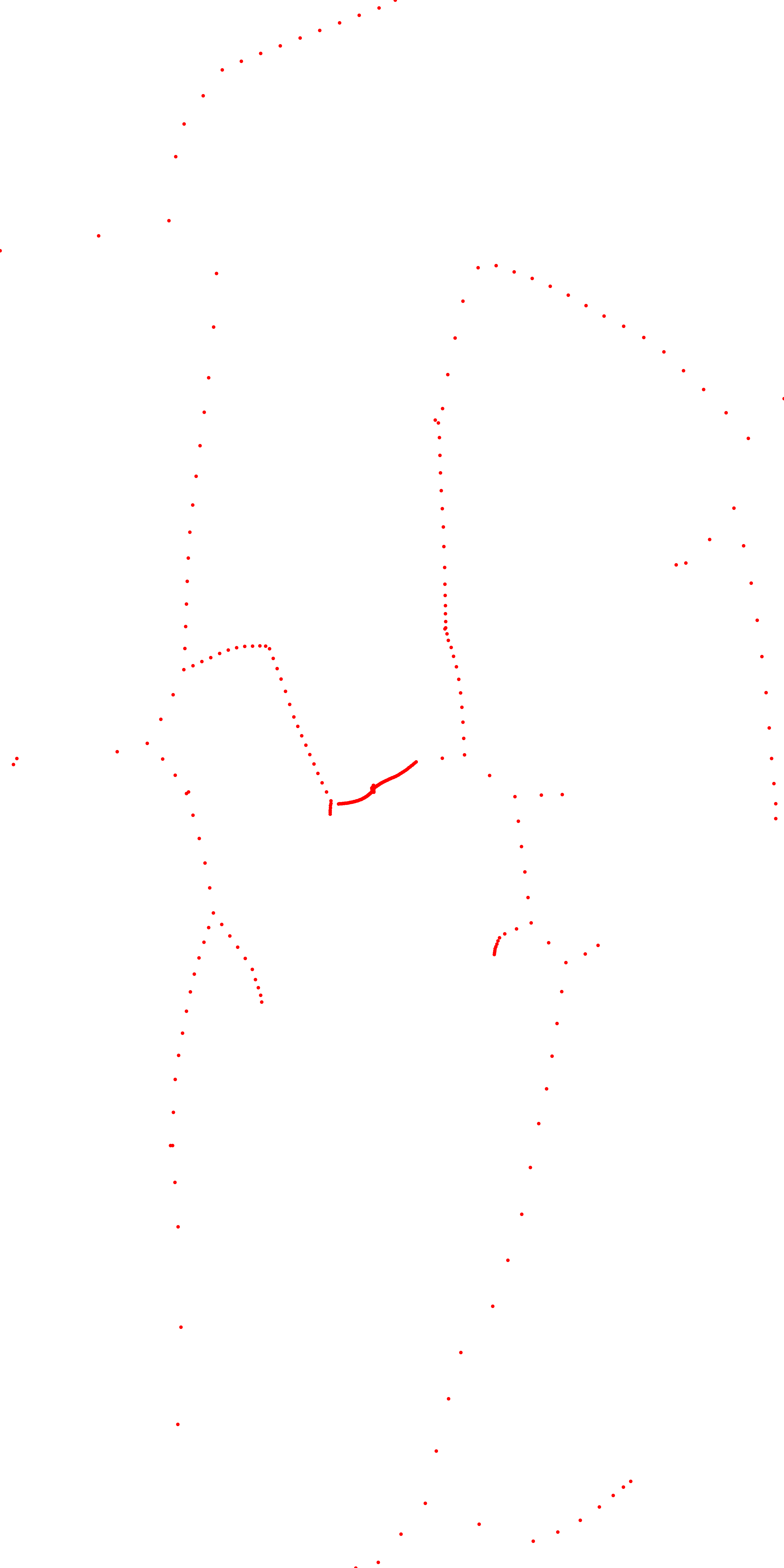}} \\ 

\frame{\includegraphics[width=0.20\linewidth, trim=5cm 7cm 3cm  9cm,clip]{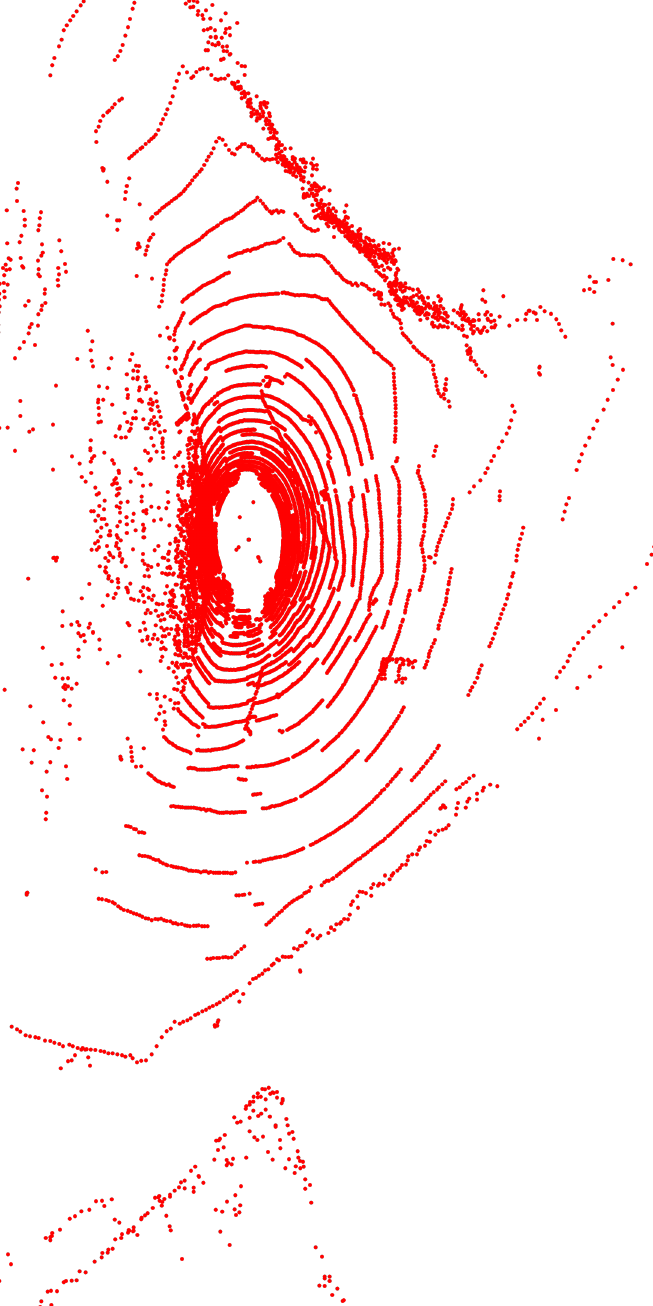}}
& \frame{\includegraphics[width=0.20\linewidth, trim=5cm 7cm 3cm  9cm,clip]{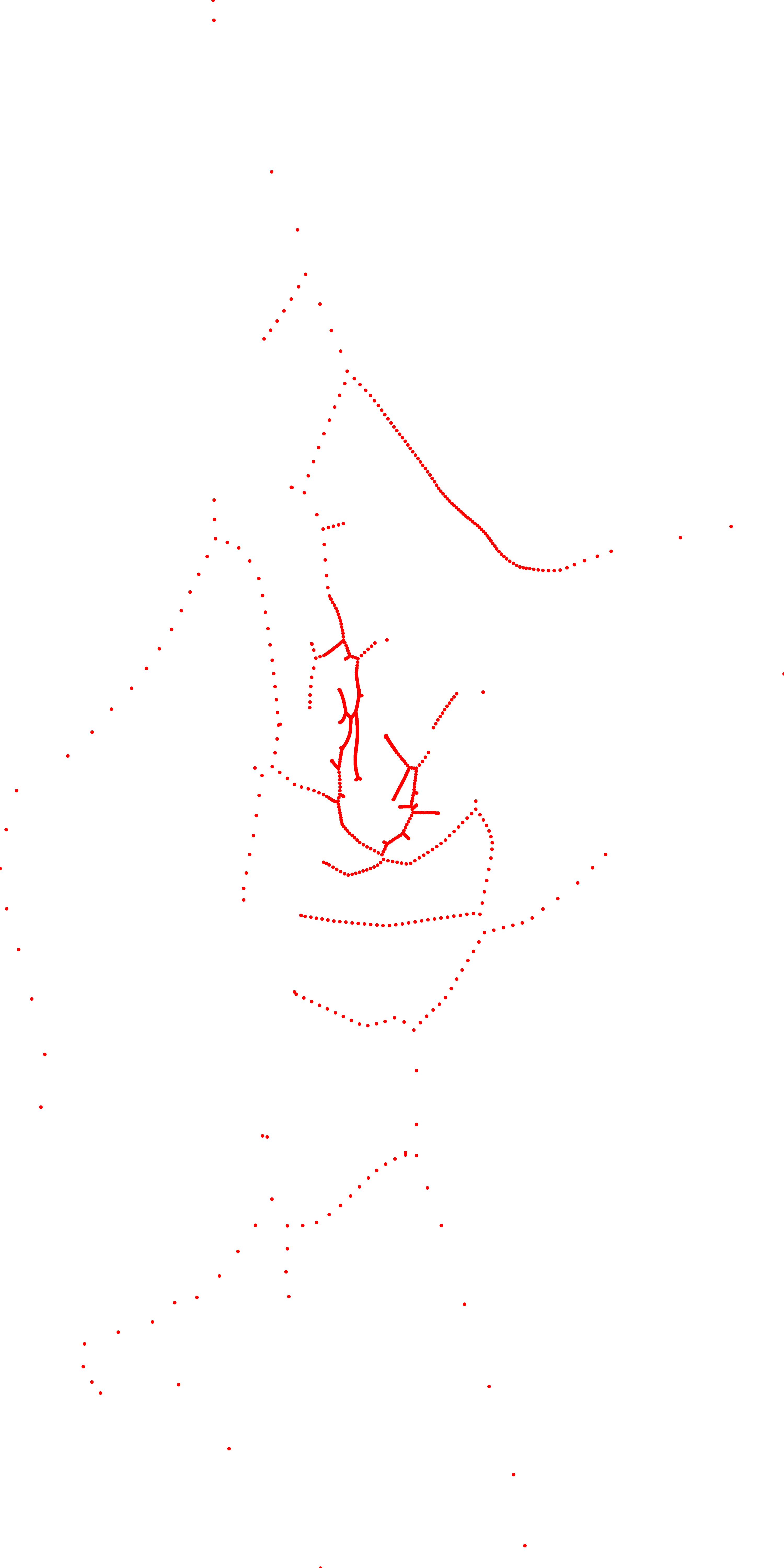}}& \frame{\includegraphics[width=0.20\linewidth, trim=5cm 7cm 3cm  9cm,clip]{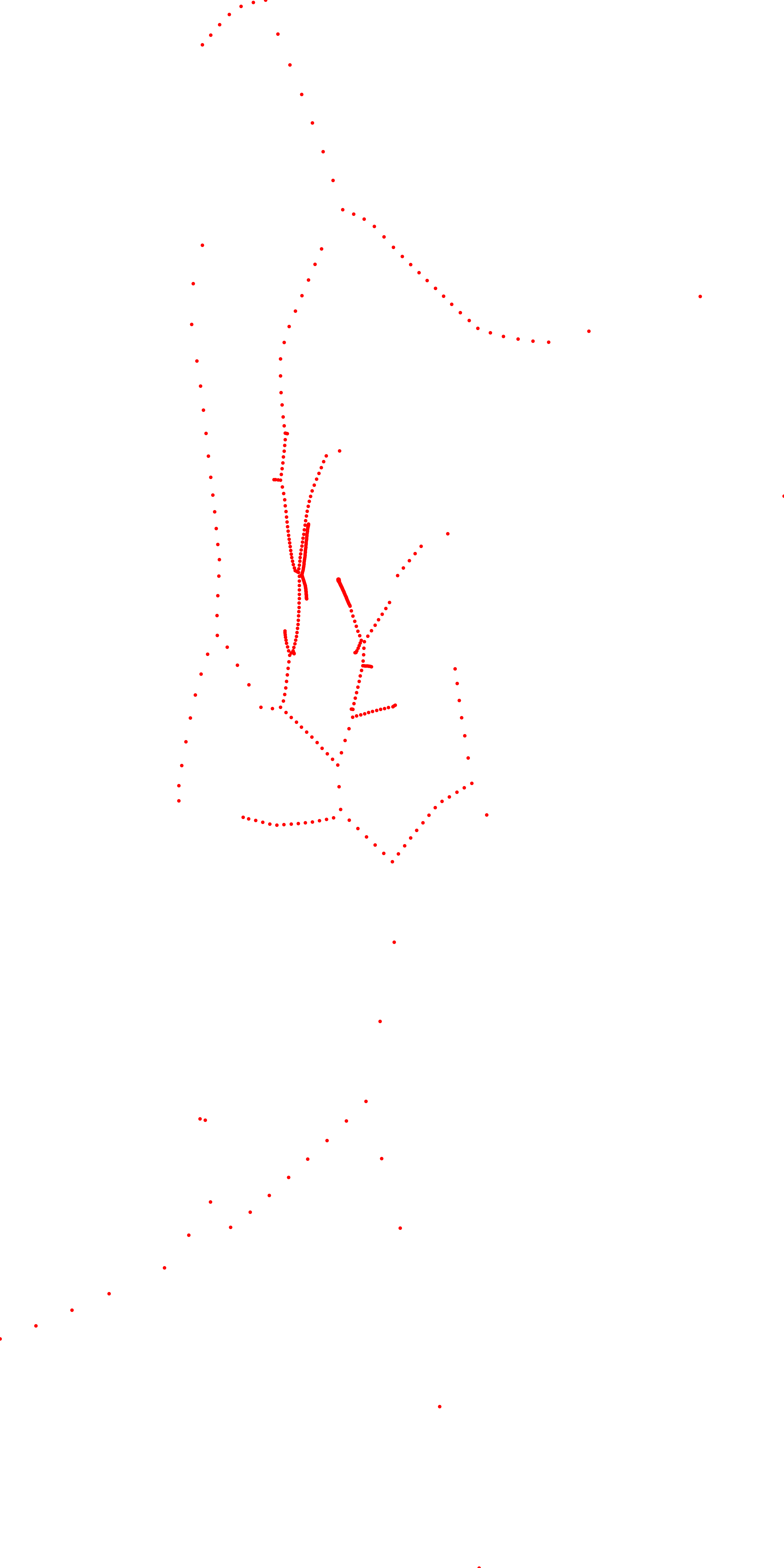}}& \frame{\includegraphics[width=0.20\linewidth, trim=5cm 7cm 3cm 9cm,clip]{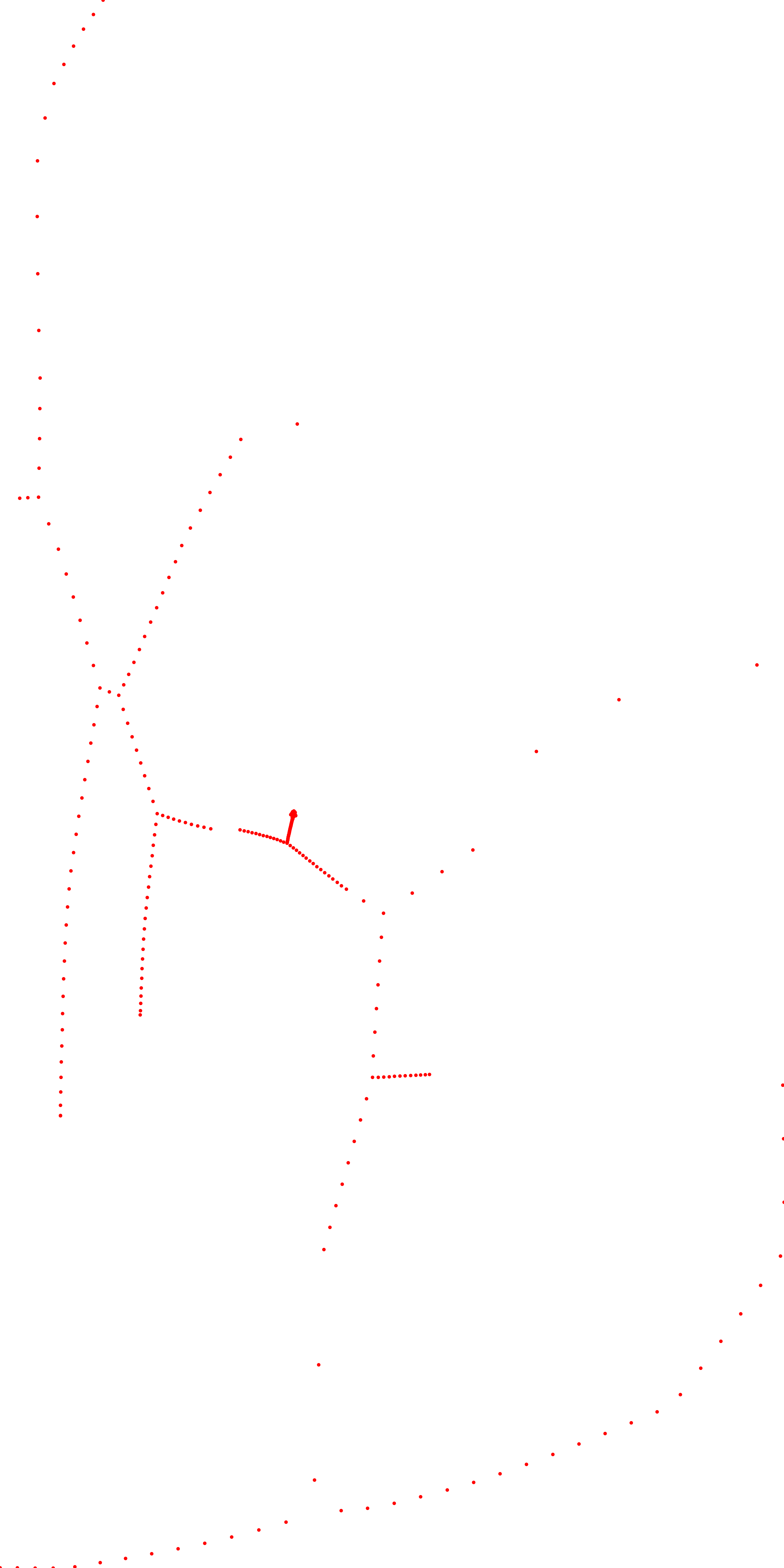}} \\ 
 
\end{tabular}
\caption{0-dim $\mathcal{PH}$ backbone results for several LiDAR scans. View from left to right. \textbf{Left}: Original 16-beam LiDAR scans \textbf{Second from left}: Backbone for 16-beam LiDAR scan. \textbf{Third from left}: Backbone for 8-beam LiDAR scan. \textbf{Right}: Backbone for 4-beam LiDAR scan. While the backbone for a 16-beam LiDAR scan is precise and accurate, the backbone for an 8 and 4-beam LIDAR scan misses several details of the original LiDAR. Thus, \glidar{} is unable to handle sparsity levels of 8 and 4-beam LiDAR scans.}
\label{tab:backbone_1}
\end{table*}

\begin{table*}
\centering
\setlength{\tabcolsep}{7pt}
\begin{tabular}{c c c c}

\frame{\includegraphics[width=0.20\linewidth, trim=5cm 7cm 3cm  9cm,clip]{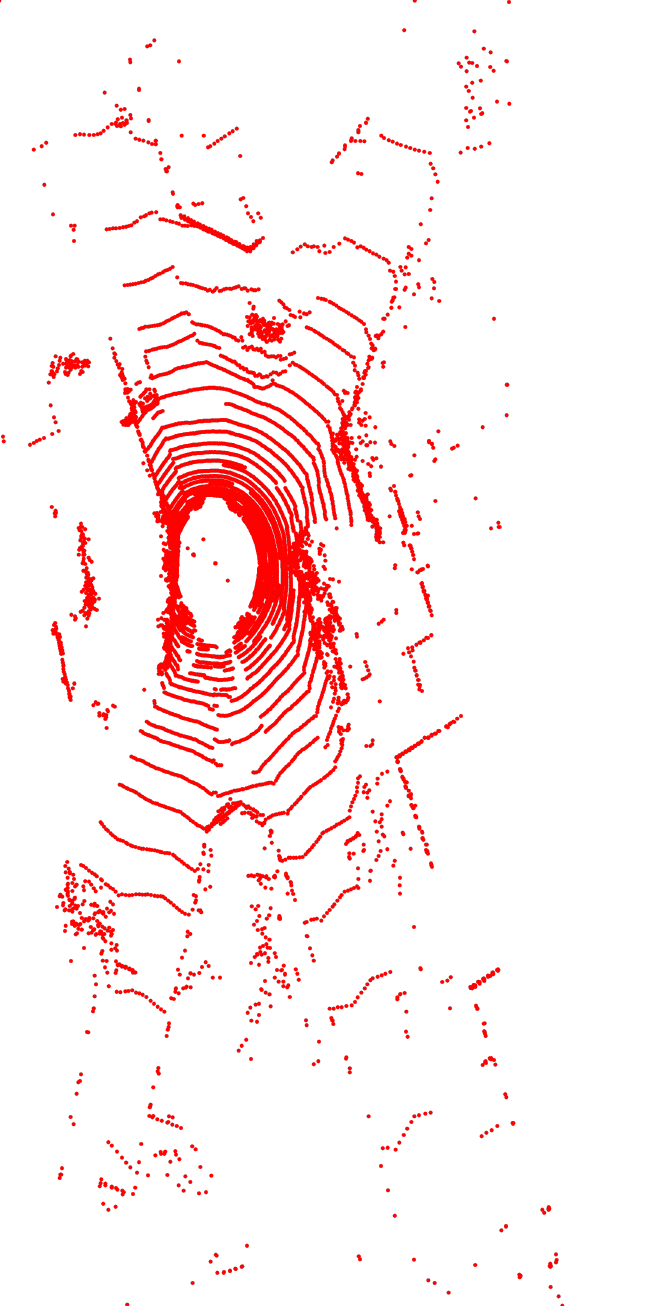}}& \frame{\includegraphics[width=0.20\linewidth, trim=5cm 7cm 3cm  9cm,clip]{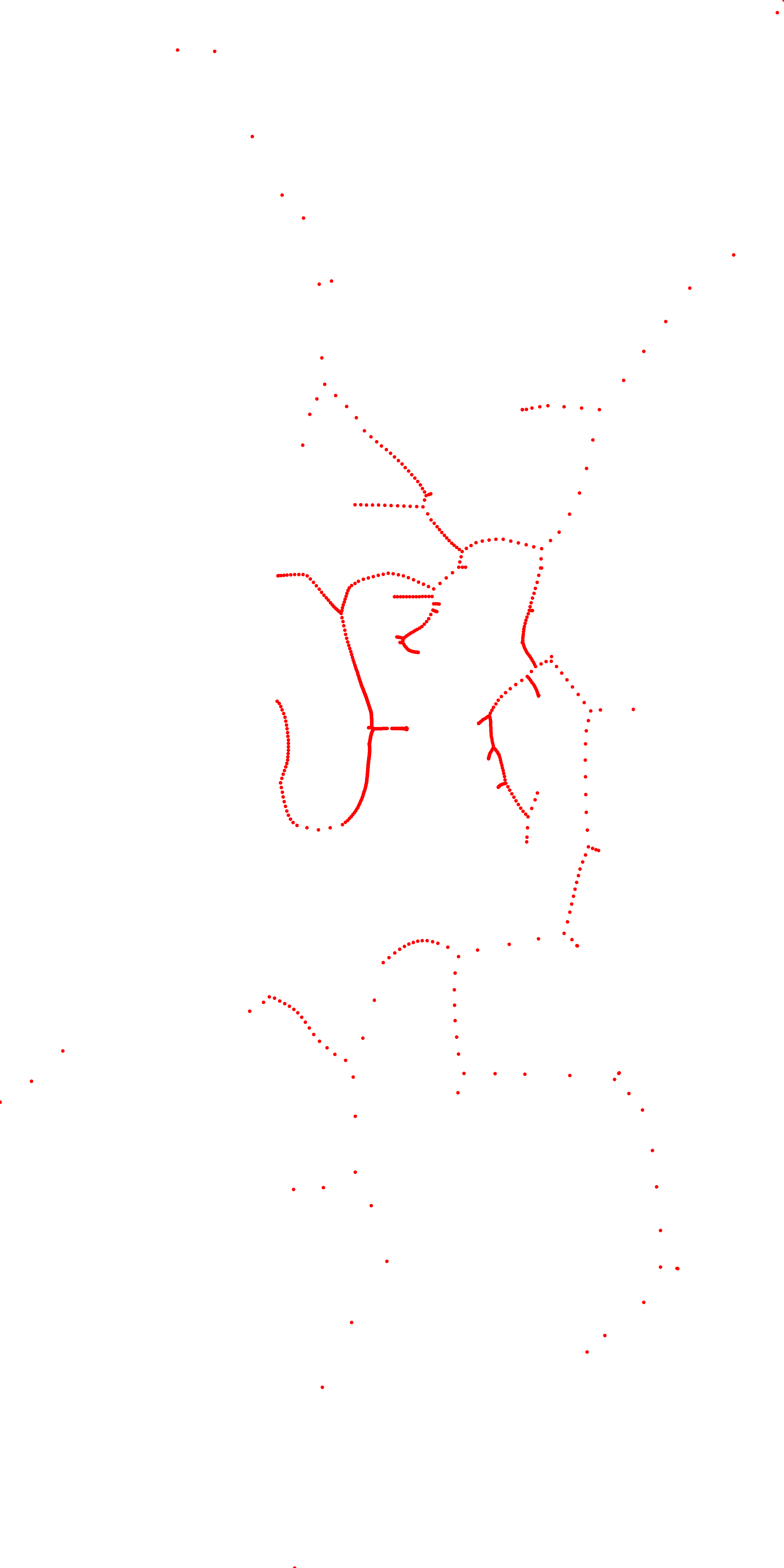}}& \frame{\includegraphics[width=0.20\linewidth, trim=5cm 5cm 3cm 11cm,clip]{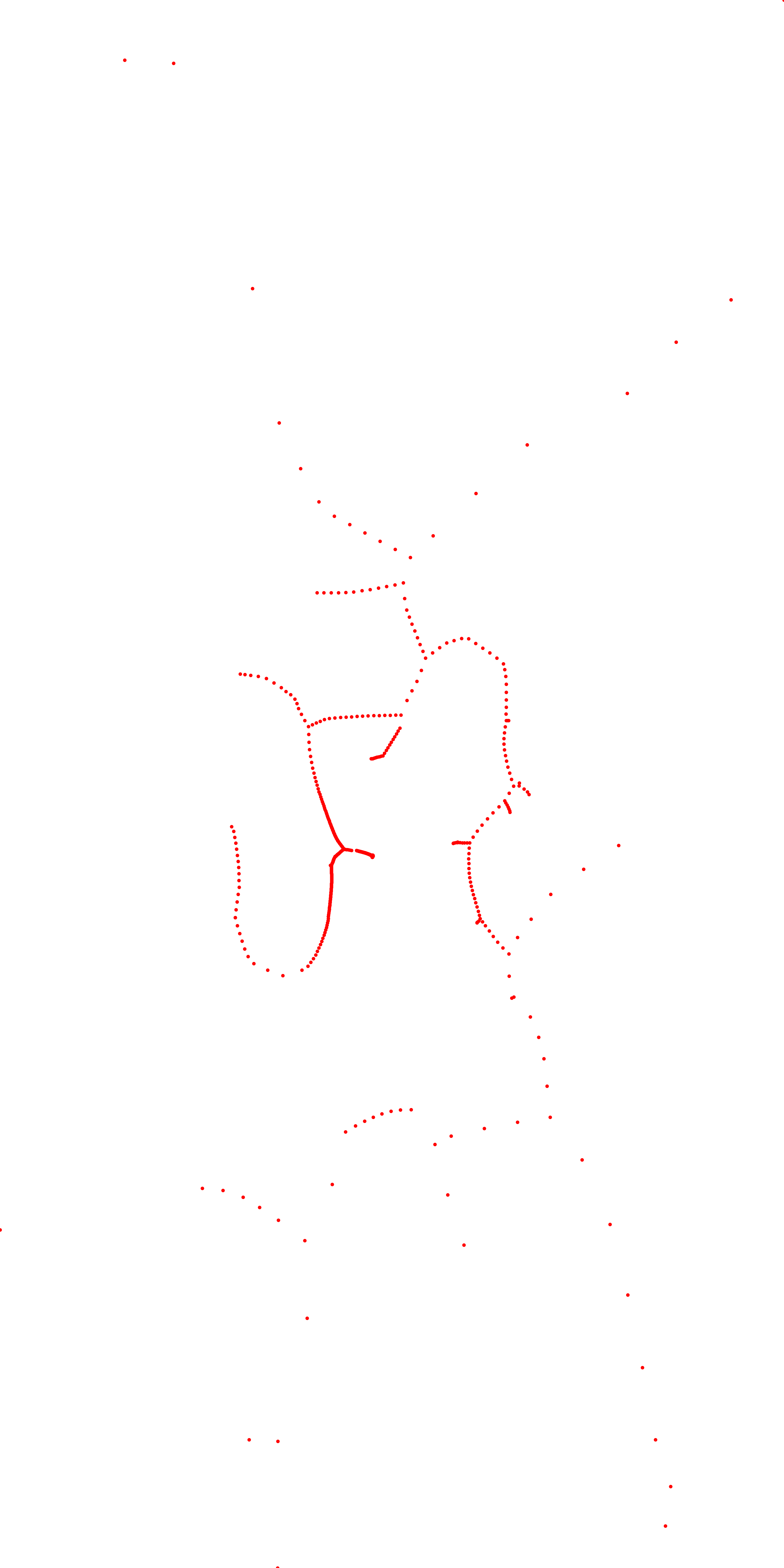}}& \frame{\includegraphics[width=0.20\linewidth, trim=5cm 7cm 3cm  9cm,clip]{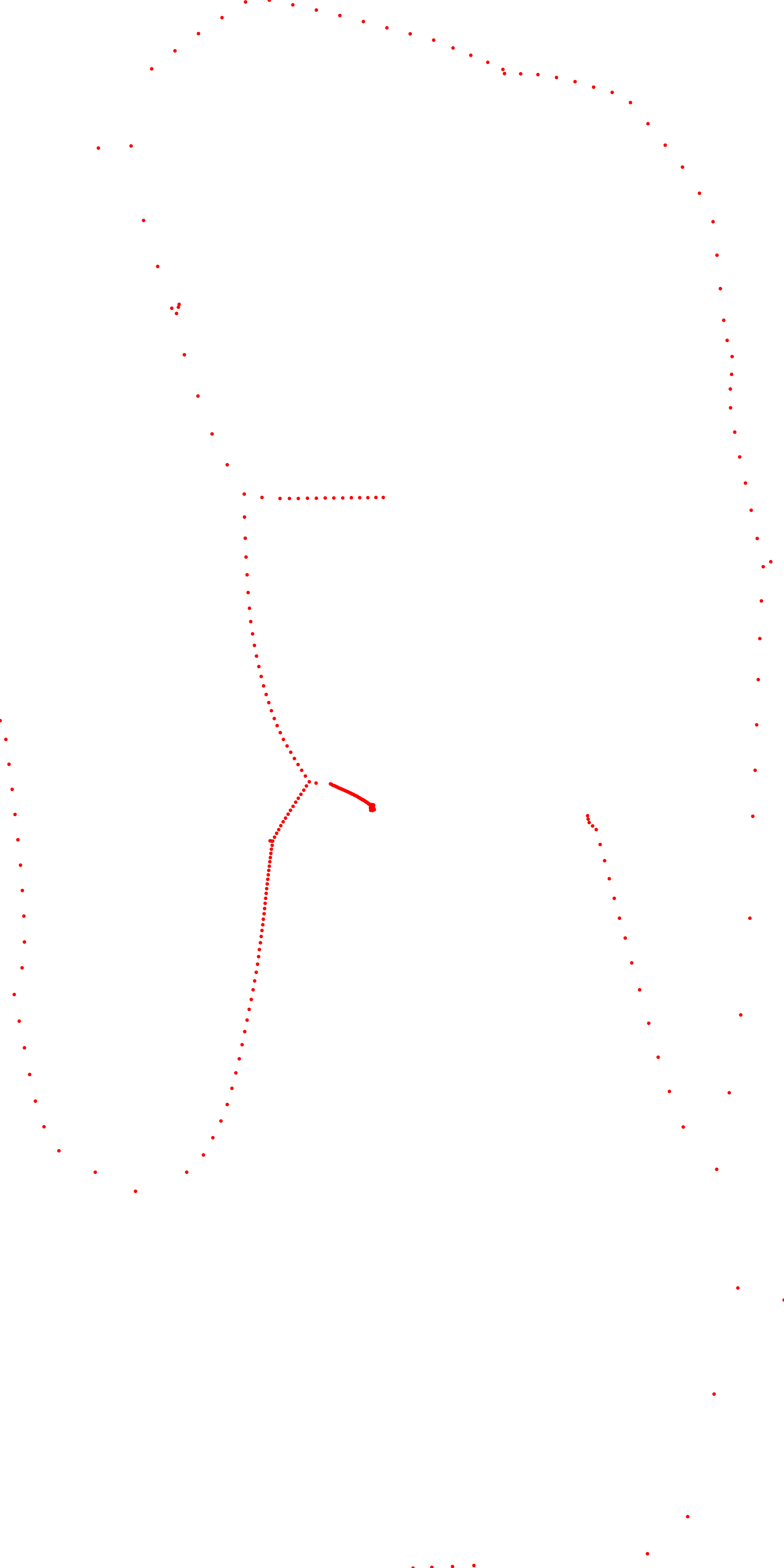}} 
 \\ 

\frame{\includegraphics[width=0.20\linewidth, trim=5cm 7cm 3cm  9cm,clip]{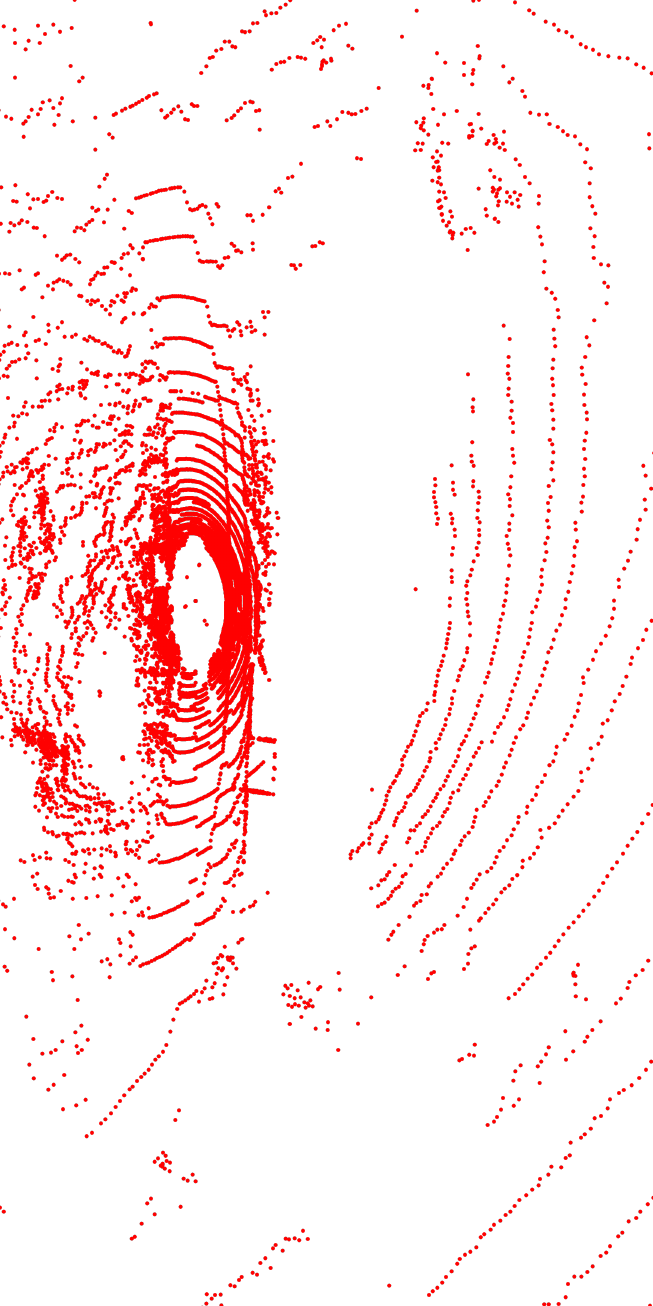}}& \frame{\includegraphics[width=0.20\linewidth, trim=5cm 7cm 3cm  9cm,clip]{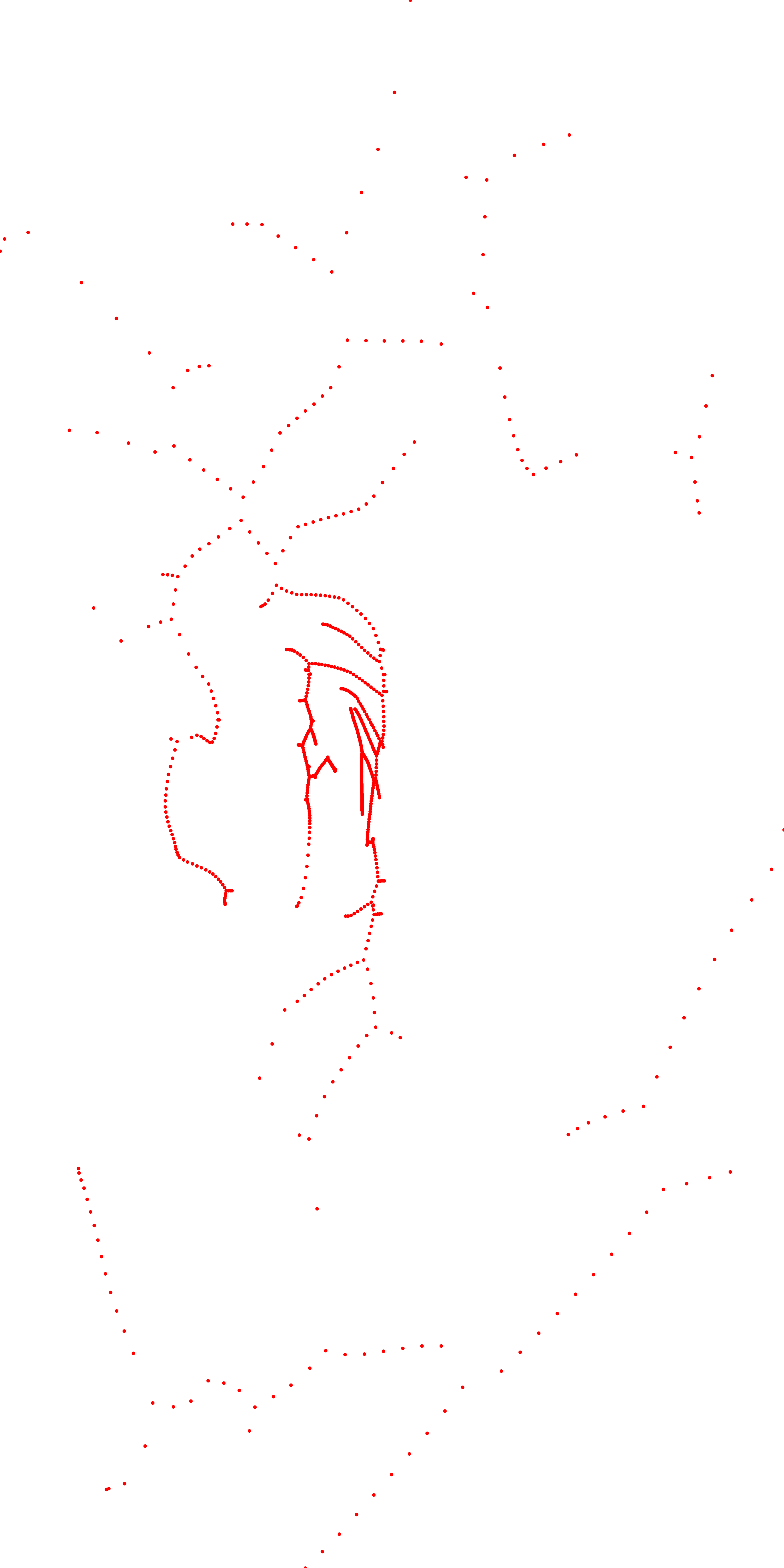}}& \frame{\includegraphics[width=0.20\linewidth, trim=5cm 7cm 3cm  9cm,clip]{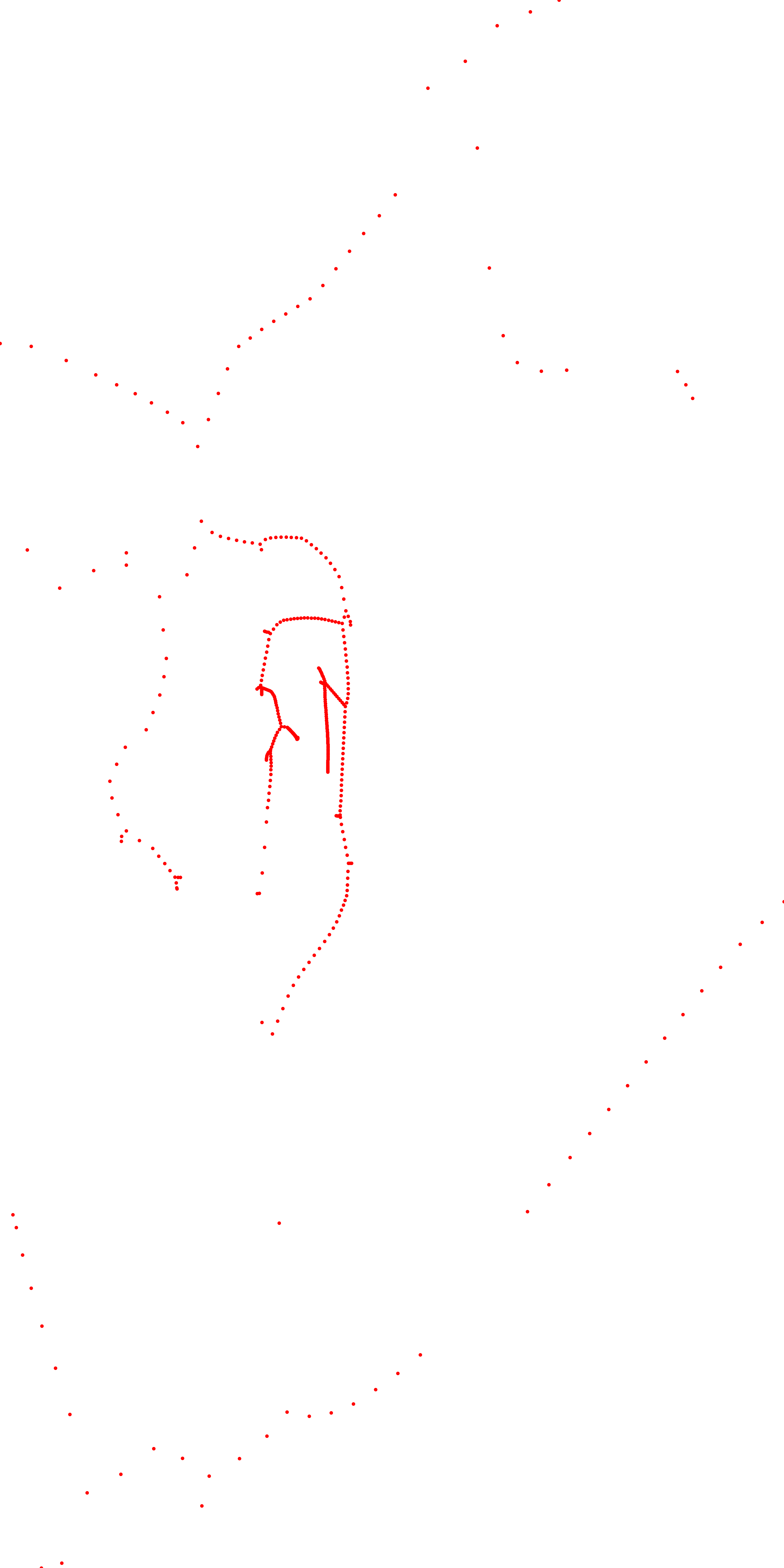}}& \frame{\includegraphics[width=0.20\linewidth, trim=5cm 7cm 3cm  9cm,clip]{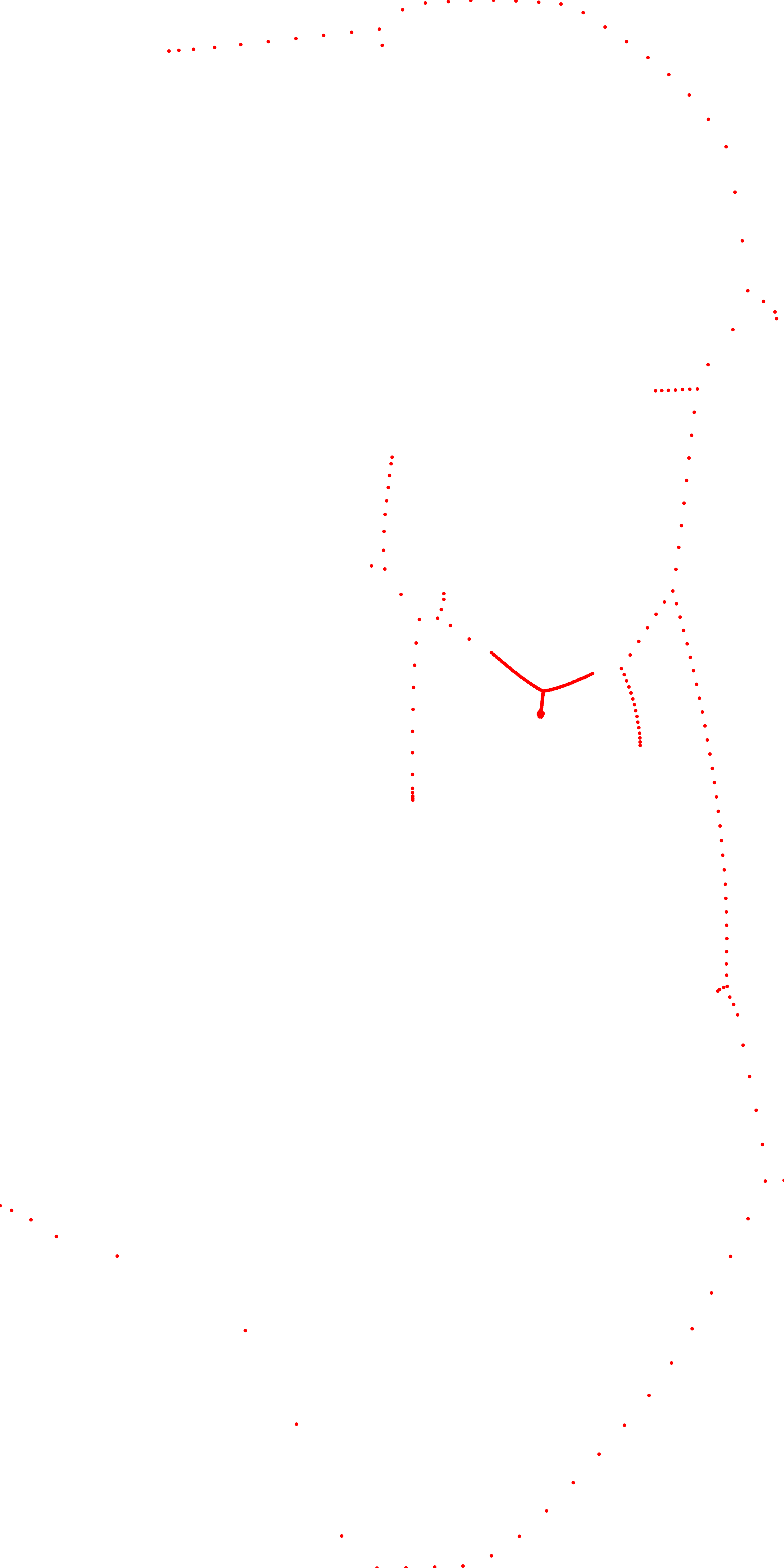}} \\ 

\frame{\includegraphics[width=0.20\linewidth, trim=5cm 7cm 3cm  9cm,clip]{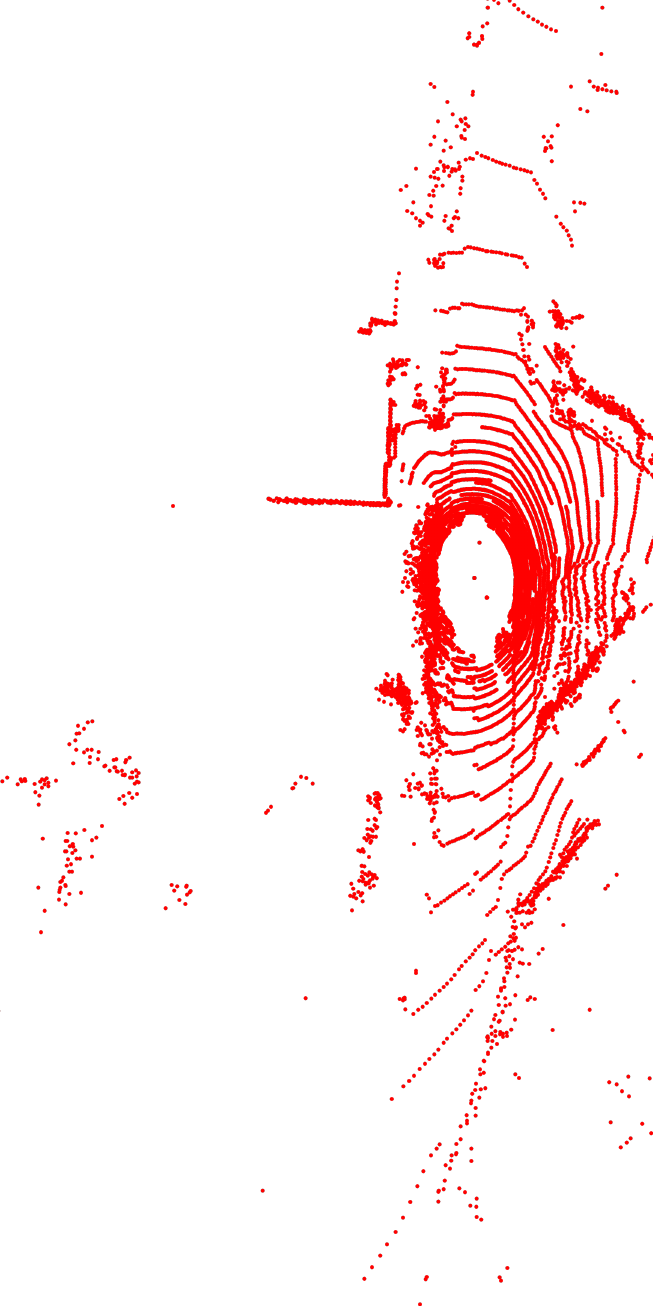}}& \frame{\includegraphics[width=0.20\linewidth, trim=5cm 5cm 3cm  11cm,clip]{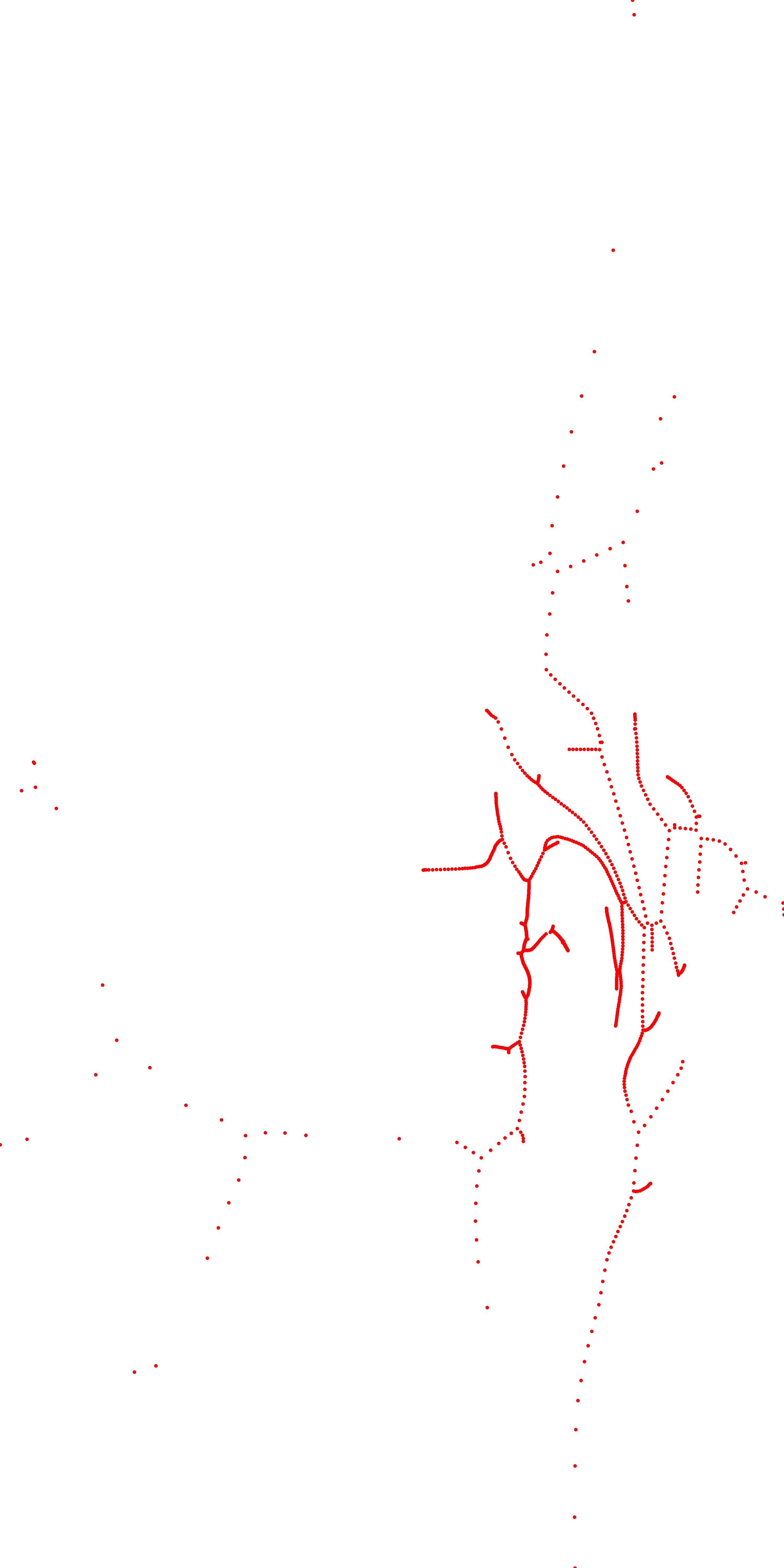}}& \frame{\includegraphics[width=0.20\linewidth, trim=5cm 7cm 3cm  9cm,clip]{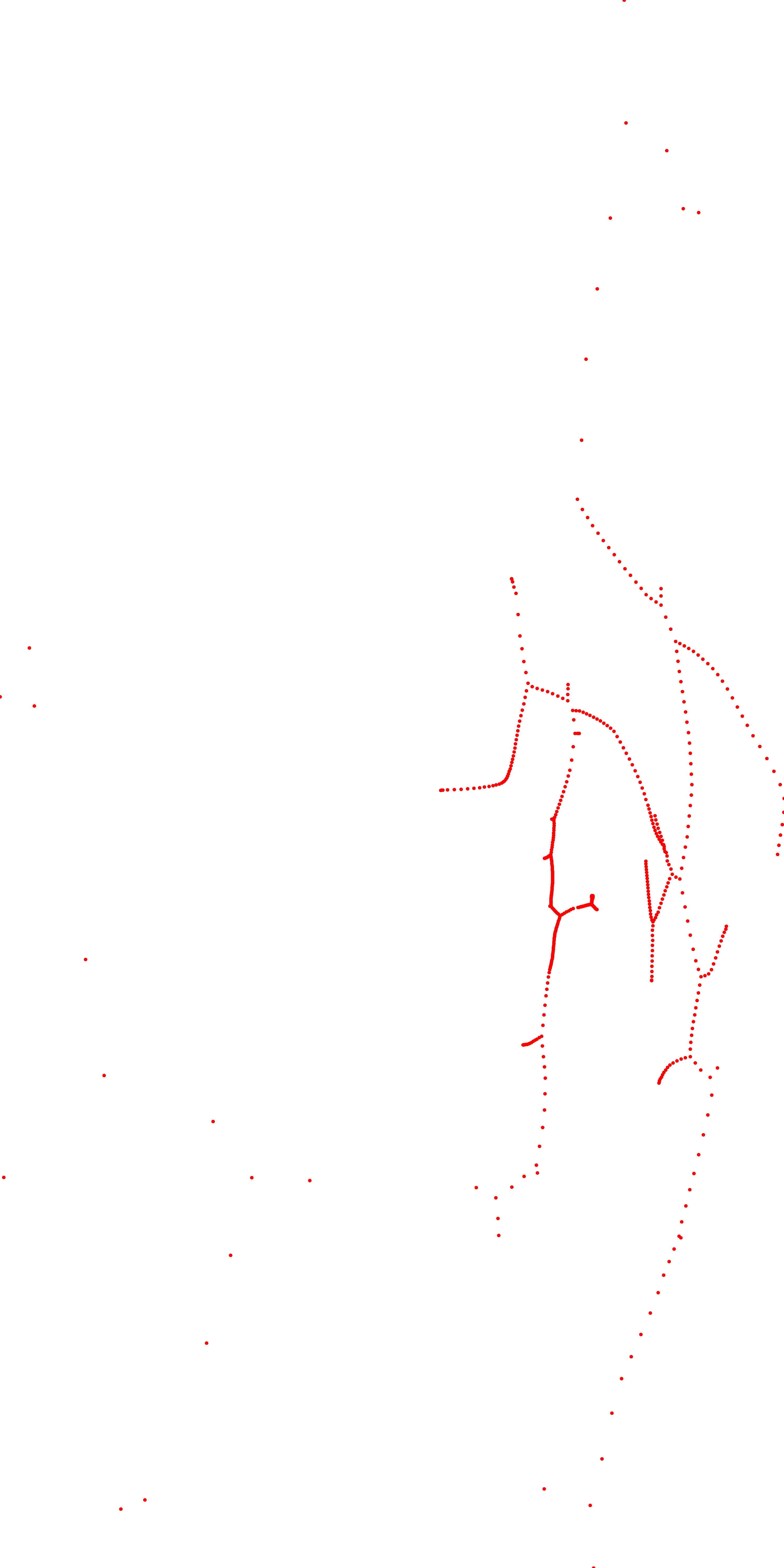}}& \frame{\includegraphics[width=0.20\linewidth, trim=5cm 7cm 3cm  9cm,clip]{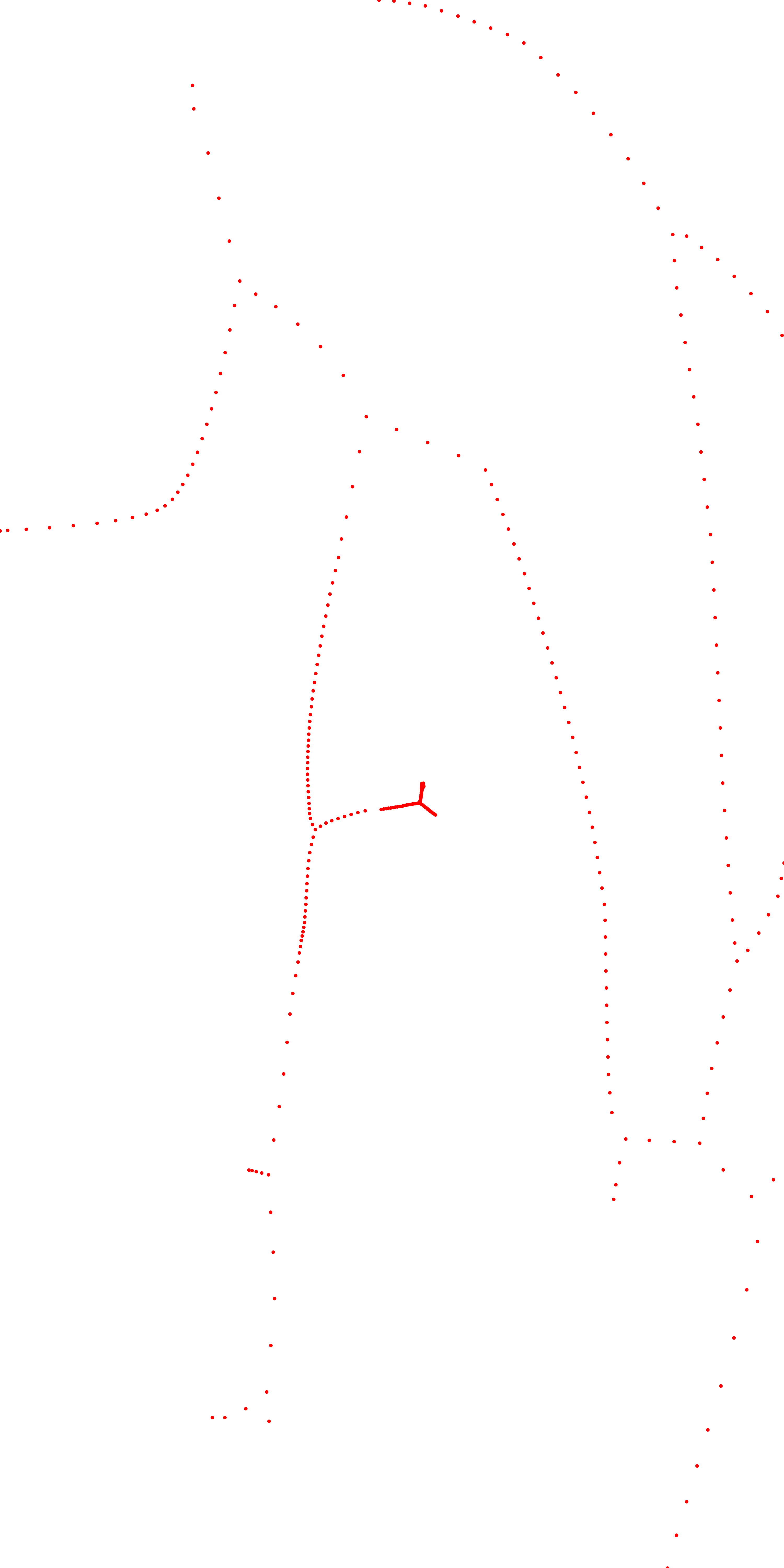}} \\ 
 
\end{tabular}
\caption{0-dim $\mathcal{PH}$ backbone results for several LiDAR scans. View from left to right. \textbf{Left}: Original 16-beam LiDAR scans \textbf{Second from left}: Backbone for 16-beam LiDAR scan. \textbf{Third from left}: Backbone for 8-beam LiDAR scan. \textbf{Right}: Backbone for 4-beam LiDAR scan. While the backbone for a 16-beam LiDAR scan is precise and accurate, the backbone for 8 and 4-beam LIDAR scan misses several details of the original LiDAR. Thus, \glidar{} is unable to handle sparsity levels of 8 and 4-beam LiDAR scans.}
\label{tab:backbone_2}
\end{table*}





{
    \small
    \bibliographystyle{ieeenat_fullname}
    \bibliography{supplementary}
}


%% file: sec/00abstract.tex
\begin{abstract}
Sparse LiDAR point clouds cause severe loss of detail of static structures and reduce the density of static points available for navigation. Reduced density can be detrimental to navigation under several scenarios. We observe that despite high sparsity, in most cases, the global topology of LiDAR outlining the static structures can be inferred. We utilize this property to obtain a backbone skeleton of a LiDAR scan in the form of a single connected component that is a proxy to its global topology. We utilize the backbone to augment new points along static structures to overcome sparsity. Newly introduced points could correspond to existing static structures or to static points that were earlier obstructed by dynamic objects. To the best of our knowledge, we are the first to use such a strategy for sparse LiDAR point clouds. Existing solutions close to our approach fail to identify and preserve the global static LiDAR topology and generate sub-optimal points. We propose \glidar{}, a Graph Generative network that is topologically regularized using 0-dimensional Persistent Homology ($\mathcal{PH}$) constraints. This enables \glidar{} to introduce newer static points along a topologically consistent global static LiDAR backbone. \glidar{} generates precise static points using 32$\times$ sparser dynamic scans and performs better than the baselines across three datasets.  \glidar{} generates a valuable byproduct - an accurate binary segmentation mask of static and dynamic objects that are helpful for navigation planning and safety in constrained environments. The newly introduced static points allow \glidar{} to outperform LiDAR-based navigation using SLAM in several settings. Source code is available at \href{https://kshitijbhat.github.io/glidr}{https://kshitijbhat.github.io/glidr}.

\end{abstract}

%% file: sec/01introduction.tex


\section{Introduction}

Sparse LiDAR scans have fewer laser beams. Fewer points fall on object structures, leading to sparse point density. The sparsity increases further for distant objects and in adverse weather conditions. Sparse scans lead to fewer static points available for SLAM. Additionally, low-cost and limited compute autonomous LiDAR systems drop LiDAR points (reducing points by several factors) to work in real-time. Under moderate to high \textit{dynamic} environments, the loss of static points to dynamic obstructions increases sparsity further. Increasing the density of points along the LiDAR scan cross-section introduces a significant portion of redundant points that are not useful for SLAM (e.g., ground points).
Further, compute-constrained systems deployed with sparse LiDAR cannot handle dense point clouds. Several works solve the scarcity of static points by completing dynamic LiDAR point clouds to obtain a precise view of obstructed static regions \cite{kumar2023movese,kumar2021dynamic}. They fail to identify and preserve the global shape and LiDAR topology, which is diluted due to sparsity. Sparsity also hinders the learning of accurate representation of the local geometry. 

Current generative models represent LiDAR data as range images and are locally robust and receptive to a local area of the input. LiDAR scans have a global topology that requires explicit focus and needs to be preserved. Apart from local structures, they have global long-range structures that require global receptivity. These structures are diluted in the presence of sparsity, leading to detail and precision loss. Existing solutions find it harder to capture the global topology in sparse scenarios, leading to a loss of detail and precision. We solve these challenges using two tools — a Graph Generative network and Persistent Homology ($\mathcal{PH}$).

Popularized by \cite{wang2019dynamic}, several works explore graph representation to capture global dependencies between LiDAR points. Graph-based models are adept at learning global topology and long-range dependencies via message passing and ensuring global receptivity. They have been shown to perform well on several tasks, including 3D object detection \cite{shi2020point,wang2021object}, point cloud up-sampling \cite{qian2021pu,liu2023point}, registration \cite{cao2023dynamic}, and segmentation \cite{zhang2023improving}. Inspired by these works, we represent LiDAR point clouds in the form of a graph to learn rich global and local shapes accurately.

As sparsity levels increase, the density of points falling on objects reduces sharply, but the global topological structure is preserved.
We exploit this property to generate explicit attention on the global topology to preserve global structures while augmenting new points. Persistent Homology($\mathcal{PH}$) has been shown to learn the global shape and topology of complex datasets. It captures topological invariants (connected components, circles, and holes) that capture the global structure and embeds geometric information in the learned LiDAR representations \cite{pun2018persistent}.
It ensures that new points are augmented along the existing and occluded static LiDAR structures by following the backbone generated using 0-dim $\mathcal{PH}$.
Several works have used $\mathcal{PH}$ to integrate topological priors as regularization constraints \cite{chen2019topological,gabrielsson2020topology}. $\mathcal{PH}$ based techniques have been used to preserve topological properties of the input in generative models \cite{pmlr-v119-moor20a} and graph-based learning tasks  \cite{Horn2021TopologicalGN}.

We list the contributions of our work here:

$\bullet$ We present \glidar{} that visualizes a sparse LiDAR scan as an undirected graph. It learns global node representations via recursive updates from \textit{k}-nearest points. Despite sparsity, the updates allow the diffusion of global topology information from distant nodes into a given node. Multiple nodes belonging to the same LiDAR structure, though separated by large distances, learn coherent representations after successive updates.

$\bullet$ To ensure that \glidar{} augments new points in line with the global topology of a static scan even in the presence of sparsity, we enforce topological priors on the latent representations and the augmented output using 0-dimensional (dim) $\mathcal{PH}$. We generate 0-dim Persistence Diagram using sub-level set filtrations. This generates a single connected component that acts as a backbone (proxy to the global topology of static LiDAR scan). It guides the model to generate static points along the LiDAR backbone.

$\bullet$ \glidar{} demonstrates superior static points augmentation on 32$\times$ sparse and dense scans with five baselines using three datasets. It also demonstrates better SLAM performance in sparse and dense settings. Current standards in SLAM use labelled information to remove dynamic points before navigation. We perform better than them, irrespective of the availability of labelled information.

$\bullet$ We qualitatively demonstrate \glidar{}'s capacity to generate accurate segmentation masks of dynamic and static points for sparse LiDAR scans while the best baseline fails to distinguish between static and dynamic points.


%% file: sec/02related-work.tex
\section{Related Work}

\subsection{Static Point Augmentation and LiDAR Generative Modelling}
Point augmentation has been experimented with across several modalities. \citet{bescos2019empty} use a segmentation network to augment static points on images from urban driving datasets with a conditional-GAN. \citet{xu2023cp3} use a pretrain-prompt-predict paradigm inspired from NLP and use self-supervised pretraining and semantic-guided predicting for point cloud completion. \citet{zhang2021unsupervised} utilize a model pre-trained on complete shapes to generate complete shapes for incomplete point clouds. Standard shape-based point cloud methods do not work well when extended to LIDAR scans \cite{caccia2018deep,kumar2021dynamic}. LiDAR-based point augmentation has been used to generate novel static points to assist navigation \cite{kumar2021dynamic,kumar2023movese}.

\label{related_generative} 
LiDAR generative modelling was introduced by \citet{caccia2018deep}. They used range image-based polar coordinate systems and structure-aware loss functions. \citet{zyrianov2022learning} formulate generation as a stochastic denoising process and leverage the score-matching energy-based model. Their sampling rate of 20s/scan makes them unusable for real-time navigation. \citet{zhang2023nerf} introduce NeRF-LiDAR that learns a NeRF representation for real-world LiDAR scenes. They require multi-view images along with segmentation labels for training. \citet{xiong2023learning} present UltraLiDAR that learns discrete representations of LiDAR scenes by aligning sparse and dense point cloud representations for LiDAR generation and densification. It relies on manual intervention to identify explicit codes for objects, which can be perturbed for LiDAR completion. This is unrealistic for end-to-end real-time navigation. 
\subsection{Topological Machine Learning}
Topological Machine Learning is an emerging field at the intersection of Topological Data Analysis and Machine Learning. It captures and quantifies the global shape and evolution of topological features - connected components, holes, voids, etc. in a dataset. An important computation tool in applied topology is Persistent Homology ($\mathcal{PH}$). \citet{giansiracusa2017persistent,
pun2018persistent,hofer2017deep} use $\mathcal{PH}$ to generate topological features for learning-based tasks. $\mathcal{PH}$ has been used for geometric problems - shape matching \cite{carlsson2004persistence}, surface reconstruction \cite{bruel2020topology}, and pose-matching \cite{dey2010persistent}. 
\citet{gabrielsson2020topology} introduce a general purpose topology layer to calculate $\mathcal{PH}$ of datasets. Several works along the same line \cite{clough2019explicit,liu2016applying,chen2019topological,dey2010persistent,gebhart2019adversarial}  use the differentiability of $\mathcal{PH}$ for multiple deep learning applications. $\mathcal{PH}$ has been integrated with deep learning across various domains - medical imaging \cite{singh2023topological}, molecular biology \cite{cang2018representability}, and social networks \cite{nguyen2020bot}.

%% file: sec/03Background.tex
\section{Background }
\subsection{Preliminaries}
Given $k+1$ affinely independent points, a $k$-simplex (or $k$-cell) is the convex hull of the $k$ vertices. It is the simplest polytope in a given dimension — a line segment in one dimension, a triangle in two dimensions, a tetrahedron in three dimensions, and so on. A simplicial complex is a collection of simplices such that intersection of any two simplex is also a simplex. In topology, homology is used to count the number of $k$-dimensional holes in a simplicial complex — number of connected components in dimension $0$, loops in dimension $1$, voids in dimension $2$, and so on.

\subsection{Persistent Homology}
A topological space can be encoded as a cell complex — a collection of $k$-dimensional simplices $(k=0,1,2...)$. \textit{Homology}, an algebraic invariant of a topological space, uses local computation to capture global shape information ($k$-dim holes) of a topological space. These holes, generalized to various dimensions form the basis of homology \cite{edelsbrunner2008computational}.

Persistent Homology ($\mathcal{PH}$) is an algebraic method to discover topological features of datasets. It converts a dataset (e.g. point cloud) to a simplicial complex and studies the change of homology across an increasing sequence of simplicial complexes $ \phi \subseteq \mathcal{C}_1 \subseteq \mathcal{C}_2 \subseteq \mathcal{C}_3....\mathcal{C}_i....\mathcal{C}_n = \mathcal{C}$, known as filtration \cite{edelsbrunner2008computational}. Topological features are computed at different spatial resolutions across the subsequence. Examining the persistence of features over a range of scales reveals insights about the underlying patterns in datasets. For point clouds, filtration is defined on the edges of the complex. We define a sub-level filtration over $\mathcal{C}$.  Every simplicial complex in the subsequence can be mapped to a number using a filtration function $f((v_0,v_1...v_n)) = \max\limits_{i<j; i,j\in 0,1,2,3...n }f\left((v_i,v_j)\right)$. This is known as a flag filtration and is based on pairwise distances between points. It is monotonic — every subsequent simplicial complex has a value higher than the previous one. For a detailed treatment of $\mathcal{PH}$, please refer to the Supplementary.


%% file: sec/04Methodology.tex
\section{Problem Formulation}

 Given a sparse scan $dy_i$ from a LiDAR sequence, we formulate \glidar{} to augment $dy_i$ with newer points along an existing static LiDAR topology-based backbone outlining the static structures. It augments existing sparse static structures as well as occluded static structures with newer points along the static topology-based backbone. To obtain the backbone, we need a  static LiDAR scan ($st_i$) that corresponds to the original LiDAR $dy_i$, minus the dynamic objects and corresponding obstructions. It serves two purposes - (a) It enables the calculation of the static LiDAR backbone using $0$-dim $\mathcal{PH}$, which serves as a prior over $dy_i$ while augmenting new points. (b) It enables the introduction of points that were obstructed by dynamic objects along the static backbone.

\subsection{LiDAR-based Graph Representation}
\begin{figure}[htbp]
  \centering
  \includegraphics[width=\linewidth]{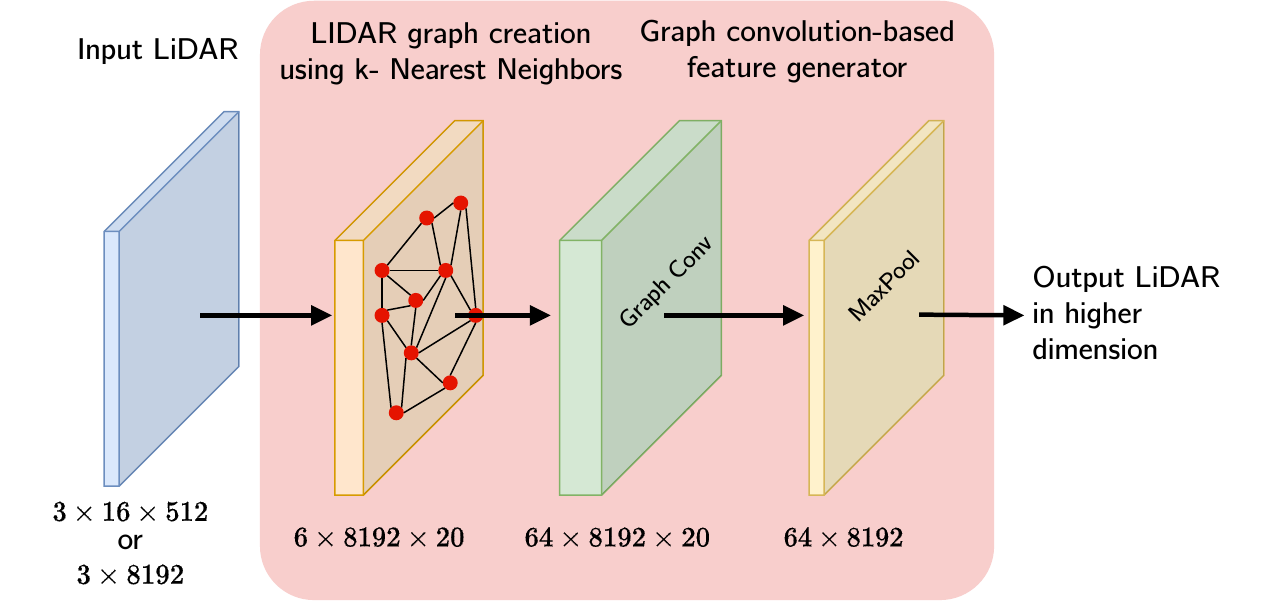}

  \caption{A sample LiDAR graph layer. The input LiDAR is transformed into a graph using \textit{k-nearest} neighbors (\textit{k}=20). Graph convolution extracts neighborhood features followed by max-pool transform the LiDAR embedding to a higher dimensional space.}
  \label{fig:lidar_graph_layer}
\end{figure}



\label{sec:lidar_graph_model}
We represent a sparse scan as an undirected graph to capture global dependencies and ensure global receptivity. A single LiDAR graph layer operates as follows:

$\bullet$ The scan points represent the nodes of the graph. Each node is associated with its unique embedding that represents its features. At the start, each node is initialized with initial features - the $x, y, z$ coordinates. Edges are drawn from each node to its $k$-nearest neighbors. A node can have several neighbors outside its sparse immediate neighborhood.

$\bullet$ Each node aggregates information from its immediate and distant neighbors using message passing, followed by a max-pooling operation to update its features. This transforms the sparse node features to higher dimensions. The current graph is then destroyed. Updated node embedding of the sparse LiDAR points represents the LiDAR graph in a higher-dimensional embedding space. This constitutes a single LiDAR graph layer (Figure \ref{fig:lidar_graph_layer}). 

For the next layers, the above steps are \textit{repeated} with the new node embeddings. It projects the nodes to even higher embedding space. Similar LiDAR nodes that were not neighbors in the original graph get a chance to become neighbors in the updated embedding space. When applied recursively, it allows points belonging to long-range structures that are distant in the original LiDAR scan to come closer or become neighbors in the new embedding space, thereby aggregating global features. Embeddings of each node in successive layers encode more global information. We stack four LiDAR graph layers that constitute the encoder of our network. At the final layer, every node is in a $512$-dimensional embedding space.
The decoder uses four $1$-D convolution layers. It projects the node features from the $512$-dimensional embedding space down to the original three dimensions ($x,y,z$) input space.

\begin{figure}
  \centering
  \includegraphics[width=1\linewidth]{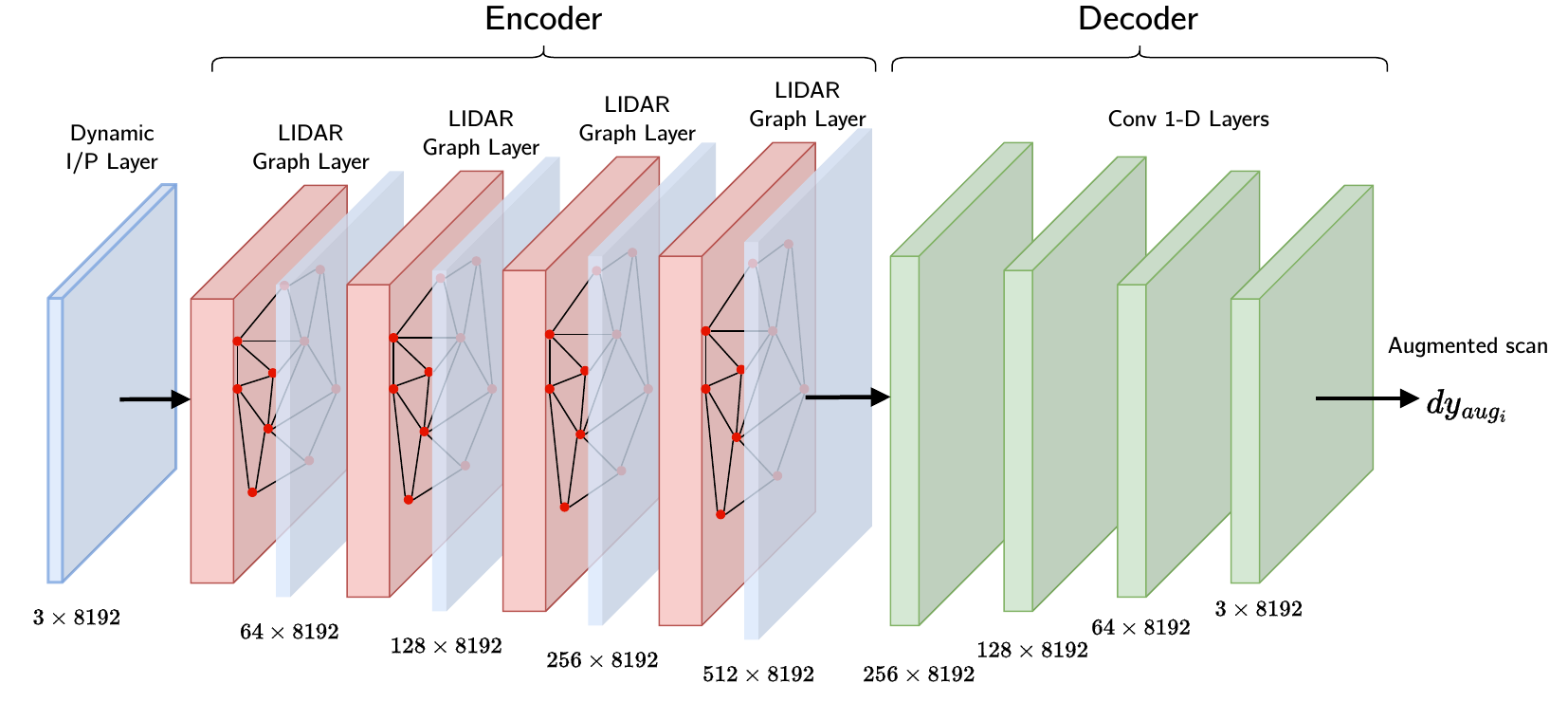}

  \caption{LiDAR Graph Generative Model using stacked LiDAR Graph layers}
  \label{fig:gnn}
\end{figure}

We use range-image representation for LiDAR by converting LiDAR point clouds to range image. It allows the flexibility to represent LiDAR as a point cloud and an image.  Loss in representation conversion only occurs while preprocessing the raw LiDAR point cloud into range image using spherical projection, before feeding to \glidar{}. This is standard across LIDAR AVs' \cite{chai2021point} and works well on navigation tasks in practice. After the range image is fed to \glidar{}, conversion of the range image to point cloud for graph convolution and vice-versa for loss computation, is a simple reshape operation that preserves one-to-one mapping. The output of \glidar{} — the augmented scan is compared against the ground-truth static scan. Range images allows us to use simple absolute error (AE) based augmentation loss instead of the costly Chamfer's Distance (CD) and Earth Mover's Distance (EMD) loss function. CD assumes the point cloud to have a uniform density, which is not the case with LiDAR. LiDAR point clouds are voluminous ($\approx$ 1,10,000 points per scan for a 64-beam LiDAR scan), unlike standard shape-based point clouds ($\approx$ 2600 points per scan). CD and EMD are memory and time-intensive for LiDAR scans, even on a powerful GPU.


\begin{figure*}
  \centering
  \includegraphics[width=1\linewidth]{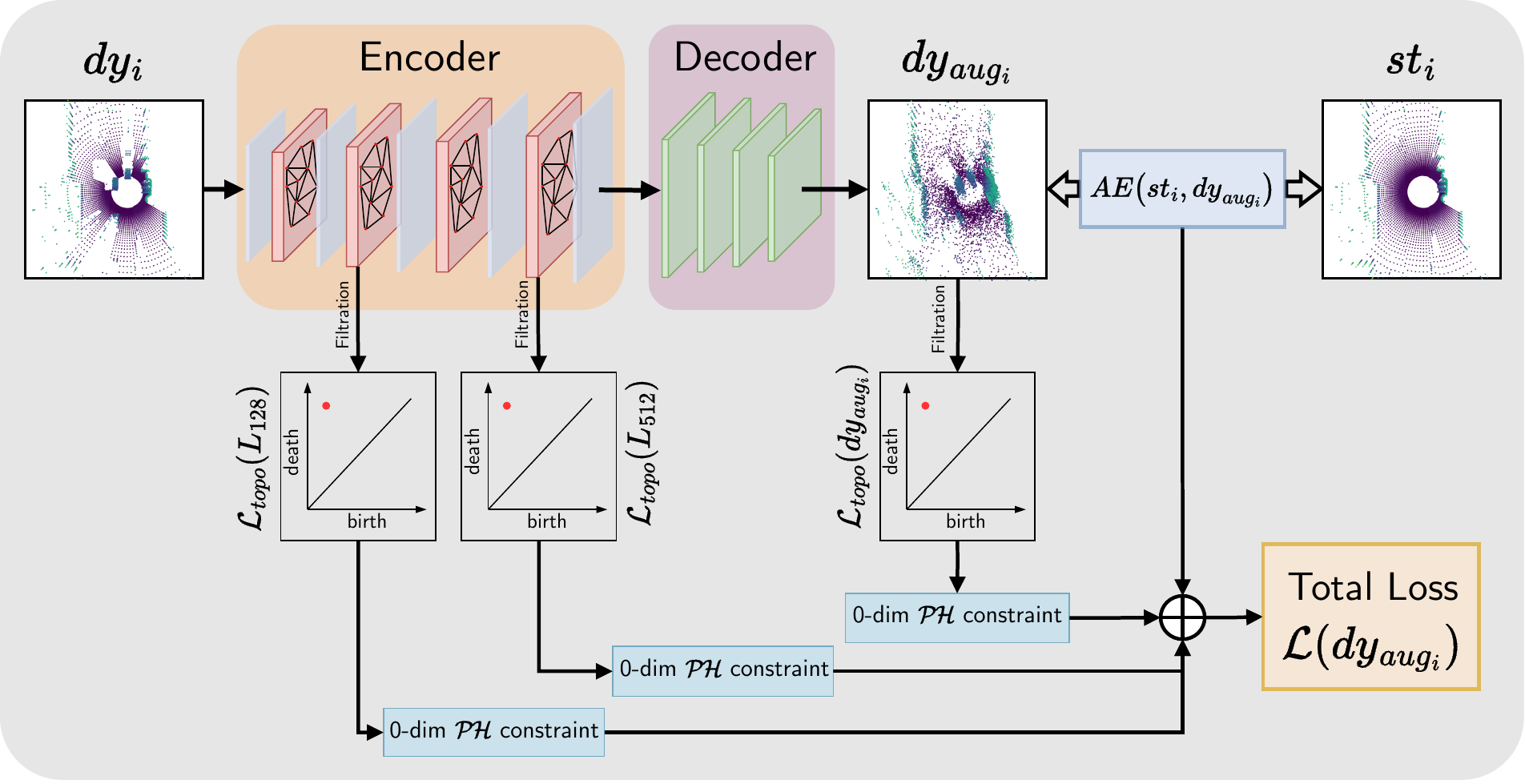}

  \caption{GLiDR network architecture with the generative LiDAR graph layers (upper branch). The lower branches constitute the 0-dim $\mathcal{PH}$ constraints and generate the topological loss.}
  \label{fig:glidar}
\end{figure*}

\subsection{Persistent Homology Constraints}
 Persistent Homology ($\mathcal{PH}$) integrates global topology into the learning problem. This is achieved either by integrating $\mathcal{PH}$ based fixed predefined features extracted from the input, or by using certain topological loss functions. The latter compute $\mathcal{PH}$ on the output or the latent representations and optimizes it to satisfy certain constraints and achieve a certain structure on the output and the latent representations.

\glidar{} follows the second strategy. The fundamental idea is to ensure shape consistency of the augmented output along a global backbone by maintaining long structures accurately, apart from local consistency. We achieve this by constraining \glidar{} to generate the static augmentation along a global prior that is estimated with the help of $\mathcal{PH}$.

Given a LiDAR point cloud, $\mathcal{L}$ (also a simplicial complex with all 0-simplices), we create a sub-level filtration on it. We define a real-valued filtration function over the simplicial complex and sub-complexes ($f^{-1}(-\infty, \alpha]$). $\alpha$ has an initial value of 0 (corresponds to the initial simplicial complex — the original LiDAR point cloud). Let the simplicial complex being created be denoted by $\mathcal{C}$. As $\alpha$ increases, new $k$-dim simplices ($k$-cells) are added to $\mathcal{C}$. These additions create or destroy homology ($k$-dim holes — connected components, cycles, voids, and so on). As the filtration progresses with increasing $\alpha$, newer $k$-dim simplices are added to the $\mathcal{C}$. Full information of the appearance and disappearance of homology of a certain dimension $k$ in a filtration is captured in a $k$-dimensional persistence diagram, $PD_k = \{(b_i, d_i)\}$. It consists of a set of (\textit{birth, death}) pairs representing the appearance and disappearance of homology. The difference (\textit{$b_i - d_i$}) denotes the persistence of a homology that appears during the filtration.

Our focus is on $PD_0$ and the largest possible $0$-dim hole — the \textit{single largest connected component} that can be generated using filtration on the augmented LiDAR. 
Calculation $\mathcal{PH}$ constraints while training is time-consuming as we increase the dimension of the Homology Group, even with very good compute. We show that sticking to \textit{0-dim}$  
 \mathcal{PH}$ features works very well for capturing long range dependencies and global topology details, especially when the size of each sample is large as in our case. 

For $0$-dim persistence, birth ($b_i$) of a homology (connected component) refers to the occurrence of the component in the $PD_0$, while death ($d_i$) refers to the case when the component merges into another component. The connected component serves as a prior — a backbone along which the graph autoencoder can generate the augmented points. Given $dy_i$ as input, our aim is to generate an augmented scan — $dy_{aug_i}$ such that filtration over it generates a single-connected component ($0$-dim homology) by the end of the filtration. The creation of the component is facilitated by a LiDAR topology loss, along with a static LiDAR augmentation loss. The augmentation loss diffuses information about the static LiDAR into the filtration. It ensures that the connected component created over $dy_{aug_i}$ at the end of the filtration for $PD_0$ follows the static LiDAR shape accurately.




\begin{figure}[h]
\setlength{\tabcolsep}{1.5pt}
\begin{tabular}{c c c} 
\frame{\includegraphics[width=0.32\linewidth, trim=2cm 9cm 5cm  9cm,clip]{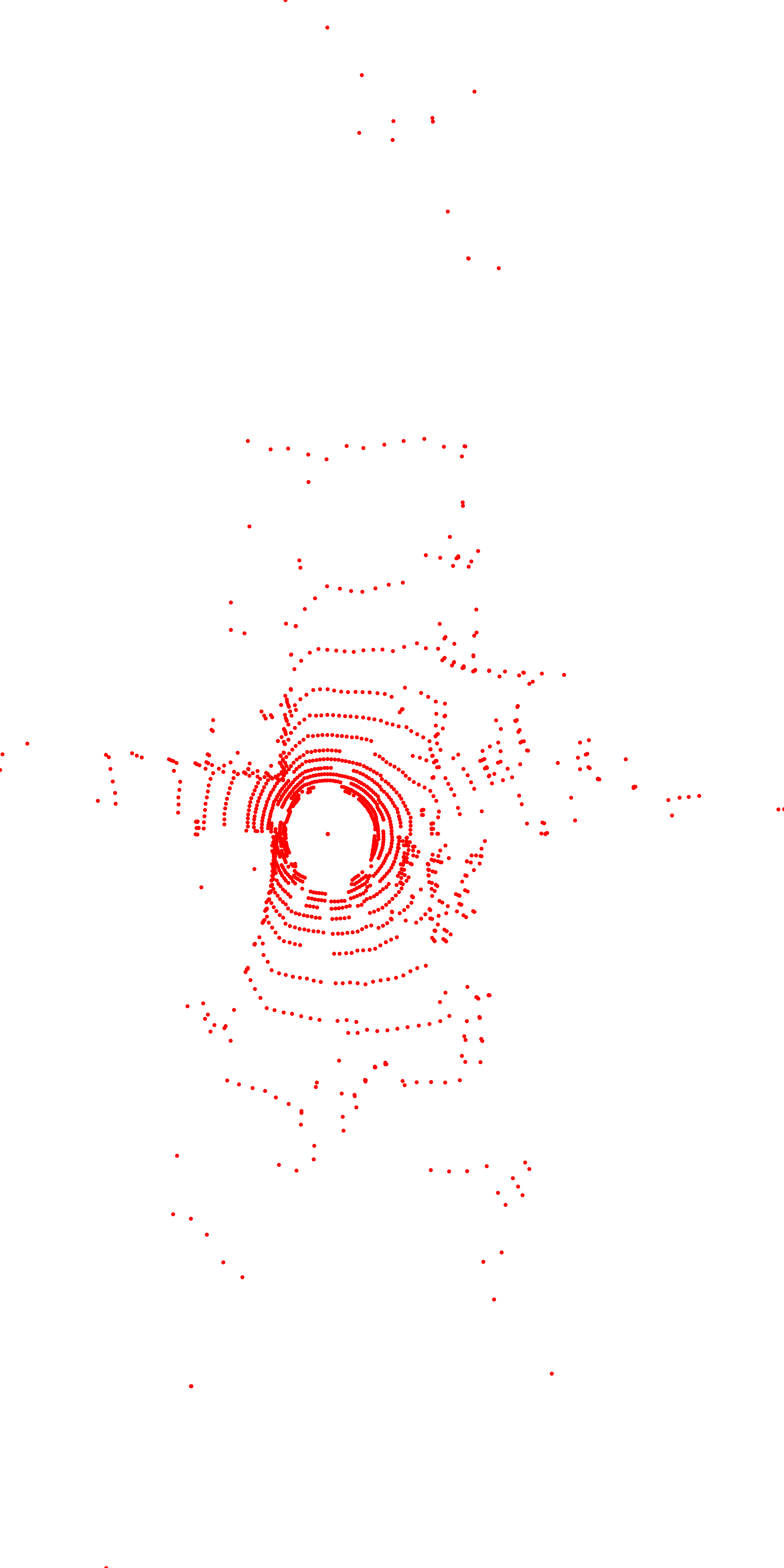}} & \frame{\includegraphics[width=0.32\linewidth, trim=2cm 9cm 5cm  9cm,clip]{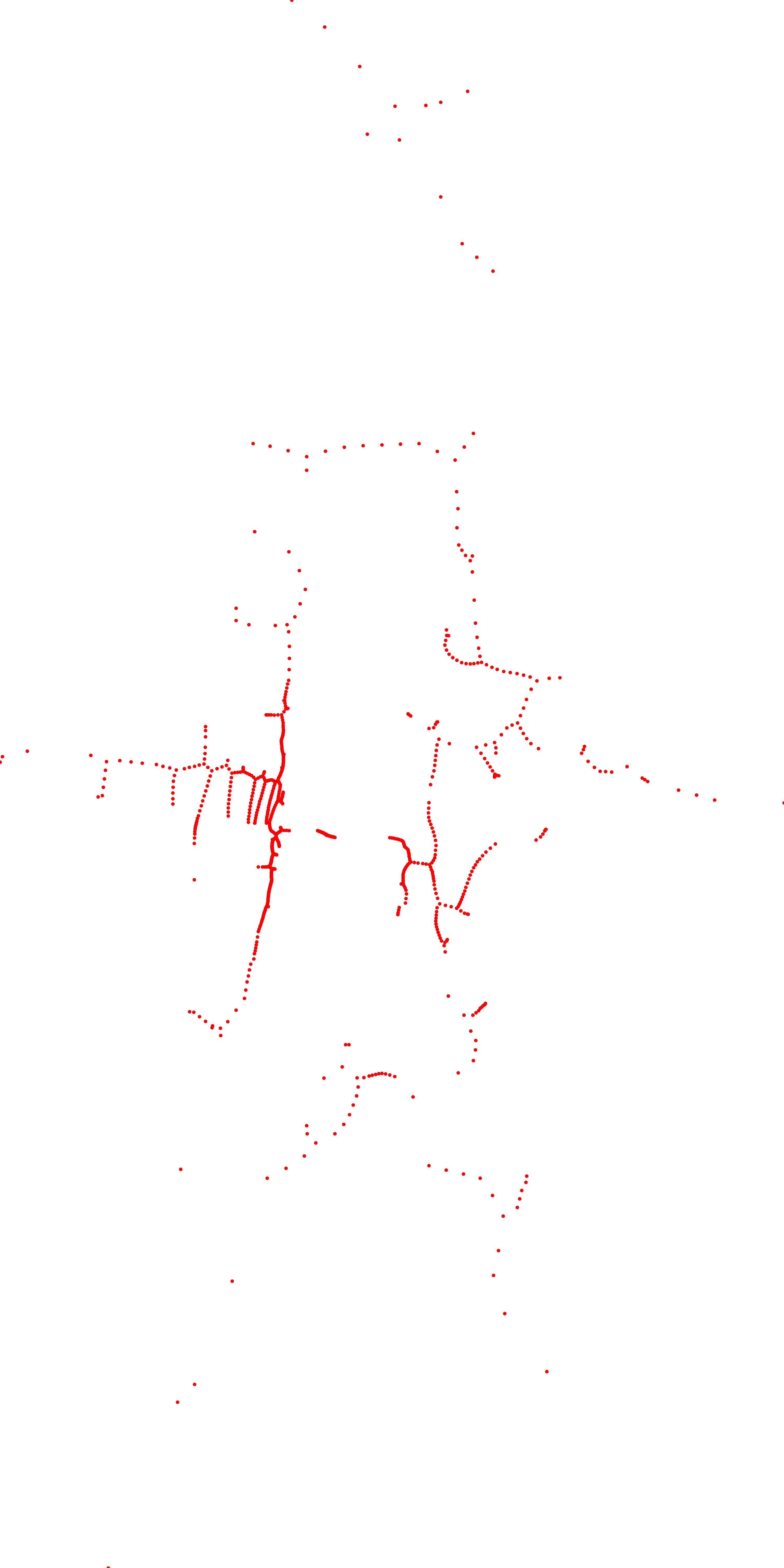}}
\frame{\includegraphics[width=0.32\linewidth, trim=7cm 46.5cm 28cm  47cm,clip]{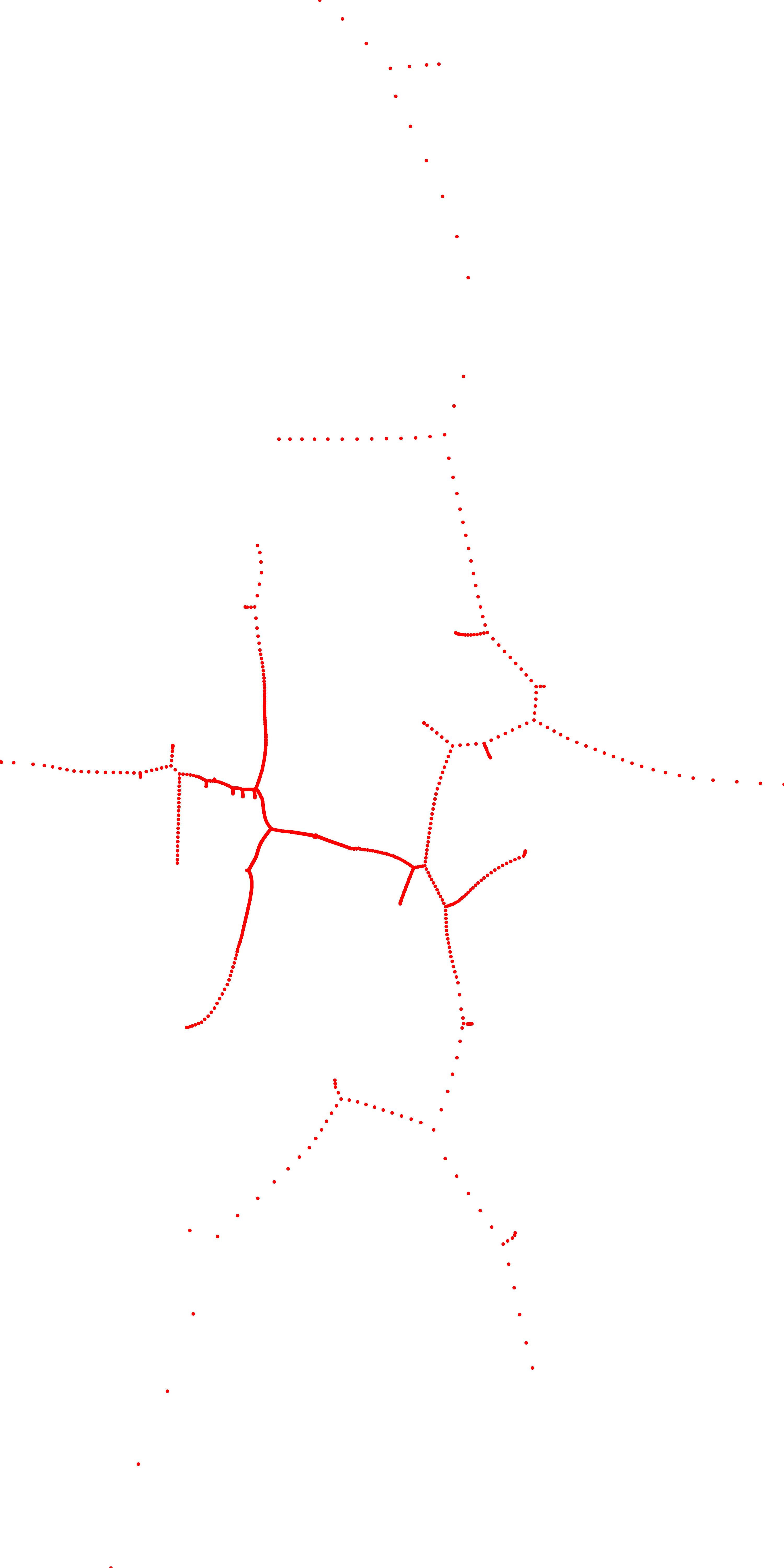}}
\end{tabular}
\caption{\textbf{Left}: Input Dynamic LiDAR \textbf{Middle}: Visualization of the $0$-dim $\mathcal{PH}$ based LiDAR backbone after some epochs. \textbf{Right}: LiDAR backbone at the end as a single connected component.
}

\label{fig:skeleton}
\end{figure}
Intuitively, for all the $0$-dim homology (connected components) possible across the filtration — we would like the one with the earliest birth to persist (have very high values of (\textit{$b_i - d_i$}) and the rest to merge in it (have a minimal value of persistence - $0$) in the ideal case. In the ideal case, $dy_{aug_i}$ is generated in a manner such that filtration over it leads to a \textit{single 0-dim homology at the end of the filtration.} This single connected component serves two purposes: \textbf{(a)} It acts as a prior and ensures that generation of $dy_{aug_i}$ progresses such that filtration over it results in only a single connected component in $PD_0$. \textbf{(b)} The single connected component created along $dy_{aug_i}$ helps to maintain long-range dependencies by connecting global structures via a single connected component (Figure \ref{fig:skeleton}). Due to the above, the $dy_{aug_i}$ adheres to the global shape of the static LiDAR point cloud, as expected. To this end, we add a simple and novel topology-based loss function on the augmented output scan ($dy_{aug_i}$). Let $\{(b_i,d_i)\}_{i=1}^n$ represent  the (\textit{birth, death}) pairs of the $0$-dim homology in the filtration computed for $dy_{aug_i}$.

\begin{equation}
\label{eq:topo_loss}
        \mathcal{L}_{topo}(dy_{aug_i}) = -\sum_{i=1}^n  \mathbbm{1} \{i=1\}(b_i-d_i) -  \mathbbm{1} \{i\neq1\}(b_i-d_i)
\end{equation}

$PD_0$ has two properties: \textbf{(a)} It has atleast one connected component for all values of $\alpha$. \textbf{(b)} $PD_{0}$ starts at $\alpha=0$ and continues for some finite value of $\alpha$, after which increasing it does not add new simplices to the filtration. We conclude that the two loss terms in the \cref{eq:topo_loss} are complementary. Minimizing the persistence of $0$-dim features from $i=2$ to $n$ ensures that the persistence of the first feature is maximized (due to point \textbf{(a)}) and leads to a single connected component. Therefore, the topology loss can be modified as follows -
\begin{equation}
\label{eq:topo_loss_modif}
        \mathcal{L}_{topo}(dy_{aug_i}) = \sum_{i=2}^n  \mathbbm{1} \{i\neq1\}(b_i-d_i)
\end{equation}
Minimizing the above term ensures a single connected component at the end of the filtration of $PD_0$. 

We apply the filtration on the augmented output scan ($dy_{aug_i}$), the $128$-dim and the $512$-dim LiDAR scan generated at the output of the second and fourth LiDAR graph layer of the encoder. The single connected component constraint introduced by $PD_0$ at the output (along with the augmentation error) allows the position of the augmented LiDAR points (in $dy_{aug_i}$) to be perturbed in a manner that follows the global static LiDAR backbone. While the augmentation loss provides an estimate of the region where a generated static LiDAR point must be placed, the backbone constraint ensures that the position of the augmented point is in accordance with the global shape of the LiDAR scan. The $0$-dim $\mathcal{PH}$ based regularization on the second and fourth layer of the encoder ensures that the intermediate graph representation also follows the same homology constraints that outline the global LiDAR topology and are compliant with $0$-dim $\mathcal{PH}$. The single connected component based global LiDAR backbone serves as a regularizer at the intermediate layers as well.

Given ($dy_i, st_i$) refers to the dynamic and static scan pair, $dy_{aug_i}$ refers to the \glidar{}'s augmented output, $L_{128}$ and $L_{512}$ refer to the updated LiDAR node embeddings in $128$-dimensional and $512$-dimensional space at the $2^{nd}$ and $4^{th}$ encoder layer, the loss function $\mathcal{L}(dy_{aug_i})$ of \glidar{} is 
\begin{equation}
\begin{aligned}
={} & \mathcal{L}_{topo}(dy_{aug_i}) + \mathcal{L}_{topo}(L_{128}) + \mathcal{L}_{topo}(L_{512}) \\
      & + AE(st_i, dy_{aug_i})
\end{aligned}
\end{equation}

\subsection{Paired Scan Generation}
\label{sec:paired_data_creation}
\citet{kumar2021dynamic} demonstrate a pipeline to collect correspondence LiDAR pairs from scratch. Existing datasets in the literature (e.g. KITTI) may not have such data available. 
We devise a simple method to obtain such pairs in order to demonstrate \glidar{}'s capabilities on such datasets. 

Assume we have a sequence of scans $K$ = \{$k_i$: 1,2,3...\}. We require pairs $K_D$, $K_S$ — $K_D$ is the set of dynamic, and $K_S$ denotes the static scans. We follow the following steps to transform $K$ into $K_S$, $K_D$, by utilizing semantic segmentation information.  \textbf{(a)} We create two copies of $K$: $K_S$ and $K_D$. We use segmentation labels to remove all dynamic objects from $K_S$ — these become our static scans. \textbf{(b)} For the dynamic part, we first remove existing dynamic objects from $K_D$ using semantic labels. We divide the LiDAR scan ($K_{D}$) into eight equal sectors along the azimuth angle. We identify a source sector — that has the presence of the most number of dynamic objects. We also identify a target sector — devoid of any objects (and) is likely to have dynamic objects (e.g, an empty sector in the opposite driving lane that can be augmented with dynamic objects moving in the opposite direction). \textbf{(c)} We extract dynamic objects from the source using $K$ (via seg. labels) and augment them in the target sector in $K_D$ identified in the previous step. To ensure that the augmentation looks realistic across the whole LiDAR sequence, dynamic objects extracted from contiguous LiDAR scans (from the source sector) are inserted into the target sector contiguously across the same LiDAR scans. For more details, please refer to the Supplementary.



%% file: sec/05Experiments.tex
\section{Experiments}
We demonstrate the quantitative performance of \glidar{} on static points augmentation along the static LiDAR backbone using six baselines on five metrics (Table \ref{tab:recon_comp}) and on SLAM in segmentation-assisted and segmentation-devoid settings (results in Supplementary) on three datasets. We also qualitatively demonstrate the LiDAR binary mask (static and dynamic points) generation in Figure \ref{fig:lidar_recon} and \ref{fig:carla-seg}.
\subsection{Experimental Setup}
The architecture of \glidar{} is explained in Section \ref{sec:lidar_graph_model} and Figure \ref{fig:glidar}.
Our models are trained using an NVIDIA A100 GPU for a maximum of 100 epochs. 
We use Adam optimizer with a weight decay of 0.0001 and a cosine annealing scheduler with a minimum learning rate of 0.001.

\begin{table*}[htbp]
\setlength{\tabcolsep}{10pt}
\footnotesize{
\begin{tabular}{c l cc  cc  cc  cc  cc} 
\toprule
& \multirow{2}{*}{Methods} &\multicolumn{2}{c}{\textbf{CD}} & \multicolumn{2}{c}{\textbf{JSD ($10^{-3}$)}} & \multicolumn{2}{c}{\textbf{MMD}} & \multicolumn{2}{c}{\textbf{RMSE ($10^{-3}$)}} & \multicolumn{2}{c}{\textbf{EMD ($10^{-3}$)}}\\
 \cmidrule(rl){3-4}  \cmidrule(rl){5-6}  \cmidrule(rl){7-8} \cmidrule(rl){9-10} \cmidrule(rl){11-12}
&  & \textbf{Dense} & \textbf{Sparse} & \textbf{Dense} & \textbf{Sparse} & \textbf{Dense} & \textbf{Sparse} & \textbf{Dense} & \textbf{Sparse} & \textbf{Dense} & \textbf{Sparse}   \\ 
\midrule
\parbox[t]{2mm}{\multirow{6}{*}{\rotatebox[origin=c]{90}{\sc{KITTI}}}}&CP3   &  73.78 & 13.98  &   21.20 & 25.60  & 0.58&0.14 & 119.20 & 117.10 &  23.00 & 19.00      \\ 
&Topo-AE& 50.60 & 26.27 & 0.79 & 1.14 & 1.13 & 1.13 & 56.18 & 67.40 & 4.42 & 6.80                                  \\
&C-Net  & 12.63 & 2.82 & 0.04 & 0.23 & 0.13 & 0.13 & 38.50 & 31.70 & 30.16 & 16.14                           \\ 
&DSLR  & 17.10 & 4.94  &  0.36 & 0.51 & 0.016 & 0.05 & 49.20 & 56.80 & 1.25 & 1.97                             \\ 
&MOVES & 7.10 & 3.57   &  0.43 & 0.62 & \textbf{0.02} & \textbf{0.02} & 39.50 & 48.00 & 1.40 & 1.95                 \\ 
&\textbf{GLiDR}   & \textbf{3.57} & \textbf{1.60}  &  \textbf{0.05} & \textbf{0.20}   &  0.06 & 0.03   &  \textbf{13.71} & \textbf{23.50}  &   \textbf{0.05} & \textbf{1.32}             \\ \midrule
\parbox[t]{2mm}{\multirow{6}{*}{\rotatebox[origin=c]{90}{\sc{CARLA-64}}}}&CP3    & 614.19 & 17.13   &   604.90 & 5.50  & 1.52&0.28  &   220.80&173.50 & 138.40 & 37.50                       \\  
&Topo-AE & 57.45 & 74.71 & 0.93 & 2.71 & 1.61 & 1.752 & 67.90 & 110.45 & 16.80 & 45.30              \\ 
&C-Net   & 14.17 & 7.96 & 0.66 & 1.34 & 0.87 & 1.05 & 53.96 & 76.98 & 20.66 & 35.33                     \\ 
&DSLR    & 13.69 & 16.42 & 0.55 & 1.27 & 0.92 & 0.87 & 65.40 & 84.74 & 7.48 & 16.54                     \\ 
&MOVES & 8.26 & 15.67  &  0.43& 1.62 & \textbf{0.75} & 0.77 & 44.20 & 92.46 & 6.73 & 18.85                  \\ 
&\textbf{GLiDR}   & \textbf{2.08} & \textbf{0.91} & \textbf{0.21} & \textbf{0.22} & 0.81 & \textbf{0.72} & \textbf{29.23} & \textbf{29.30}& \textbf{2.60} & \textbf{2.57}    \\                 \midrule         
\parbox[t]{2mm}{\multirow{6}{*}{\rotatebox[origin=c]{90}{\sc{ARD-16}}}}&CP3    & 12.27 & 8.40 & 10.10 & 1.50  & 0.14 & 0.09 & 98.88 & 122.5&         15.70 & 41.60                            \\ 
&Topo-AE & 12.15 & 2.96 & 0.39 & 0.24 & 1.27 & 1.38 & 44.33 & 35.02&        4.70 & 2.30                            \\ 
&C-Net   & 2.78 & 1.19  & 0.23 & 0.19 & 0.34 & 0.65 &  34.88 & 31.34   &     14.17 & 16.83                         \\ 
&DSLR    & 0.40 & 0.27   &  324.37 & 166.99   & \textbf{0.02} & \textbf{0.02}  &  34.07 & 34.39   &   1.58 & 1.69              
            \\ 
&MOVES & 0.34 & 0.33  &  436.30 & 159.31 & \textbf{0.02} & \textbf{0.02} &   \textbf{14.69} & 33.40  &    2.30 & 0.70   \\ 
&\textbf{GLiDR}   & \textbf{0.29} & \textbf{0.22}  &   \textbf{0.061} & \textbf{0.02}  &  \textbf{0.02} & \textbf{0.02}   &      18.80 & \textbf{18.06}  &        \textbf{0.04} & \textbf{0.05}    \\
\bottomrule
\end{tabular}
}
\caption{Static Augmentation Comparison of \glidar{} against five baselines using five metrics for sparse and dense LiDAR. For all metrics, lower is better.}
\label{tab:recon_comp}
\end{table*}

\subsection{Datasets} 
\textbf{(a)} KITTI Odometry dataset \cite{geiger2012we} is a 64-beam LiDAR dataset. It has 11 sequences with ground truth poses. Sequences 00-02,04-07,09-11 are used for training, 03 for validation, and 08 for testing \glidar{}.  We test our paired scan generation (Section \ref{sec:paired_data_creation}) pipeline on KITTI.\\
\textbf{(b)} ARD-16 is a 16-beam industrial dataset collected using an UGV. It is distinct from urban road-based datasets in terms of structure and the type of objects. It is 4$\times$ sparser than other datasets. It has correspondence information available. We follow the train, test, and protocol mentioned at \citet{dslr-git}.\\
\textbf{(c)} CARLA-64 is an extensive simulated 64-beam urban dataset with correspondence information. We follow the train, test protocol from \citet{dslr-git}.

We use LiDAR range image representations. The original dimensions for KITTI and CARLA are $3\times 64 \times 1024$, while it is $3\times 16 \times1024$ for ARD-16. The second and third dimensions represent the height and width of the range image. For all experiments, we \textit{sparsify the outermost dimension} by $8\times$ by selecting every $8^{th}$ column in the range image. The outermost dimension is reduced to $128$ for all experiments. For KITTI and CARLA-64, we use two versions - sparse (16-beam) and dense (64-beam). For the former, we extract LiDAR points generated by every fourth beam. The final shape is $3\times 16 \times 128$ ($32\times$ sparser). The dense version has a shape of $3\times64\times128$.
ARD-16 is already a 16-beam dataset. We use the original version and a sparser 8-beam version for ARD-16 with shapes $ 3\times 16 \times 128$ and $3 \times 8 \times 128$, respectively.


\subsection{Baselines}
\subsubsection{Static Points Augmentation}
\label{sec:baselines}
We compare \glidar{} against baselines that demonstrate strong performance on new static points augmentation and point cloud completion across several modalities. We do not use LiDARGen \cite{zyrianov2022learning}, NeRF-LiDAR \cite{zhang2023nerf},
Ultra-LiDAR \cite{xiong2023learning} for reasons explained in Section \ref{related_generative}. We use the following methods as baselines for our work \textbf{(a)} CP3 \cite{xu2023cp3} \textbf{(b)} Topo-AE \cite{pmlr-v119-moor20a} \textbf{(c)} CoarseNet (C-Net) \cite{wang2022coarse} \textbf{(d)} DSLR \cite{kumar2021dynamic} \textbf{(e)} MOVES \cite{kumar2023movese}. We give details of the baselines in the Supplementary.

\subsection{Evaluation Metrics}
\subsubsection{Static Points Augmentation}
We rigorously test \glidar{} using 5 metrics - \textbf{(1)} Chamfer's Distance \textbf{(2)} Earth Mover's Distance \textbf{(3)} Jenson Shannon Divergence \textbf{(4)} Root Mean Square \textbf{(5)} Maximum Mean Discrepancy. We provide details on these in the Supplementary.
\section{Results}
\subsection{Static Points Augmentation}
We compare the static point augmentation results of \glidar{} against the baselines in Table \ref{tab:recon_comp}. \glidar{} performs better than the baselines across most metrics. It retains the global topology of sparse LiDAR and generates points along the global static backbone accurately (Figure \ref{fig:lidar_recon}). This is due to the explicit attention on the global topology induced by the $0$-dim $\mathcal{PH}$ prior and the graph layers. Stacked graph layers learn global shape properties in the high-dimensional embedding space. Closeness in this space represents semantic similarity over long distances in the original LiDAR. Similar nodes (e.g. belonging to a long-range structure) far in the input space gain proximity in the high-dimensional space. $0$-dim $\mathcal{PH}$ constraints capture the global static LiDAR structure in a single connected component (Figure \ref{fig:skeleton}) and enforce \glidar{} to generate points along this backbone. 

The $0$-dim $\mathcal{PH}$ prior, and the stacked graph layers work well despite reduced LiDAR point density. Both look over a LiDAR on a global scale. It enables \glidar{} to capture global dependencies even if manifested through a sparse set of points over a long sequence. These characteristics allow \glidar{} to identify accurate static structures and backbone and align the augmentation process accurately along them.

\subsubsection{Segmentation of Dynamic Objects}
\label{sec:dy_obj_seg}
\begin{figure}
\setlength{\tabcolsep}{1pt}
\begin{tabular}{c c c} 
\frame{\includegraphics[width=0.34\linewidth, trim=8.9cm 0cm 8.9cm  0cm,clip]{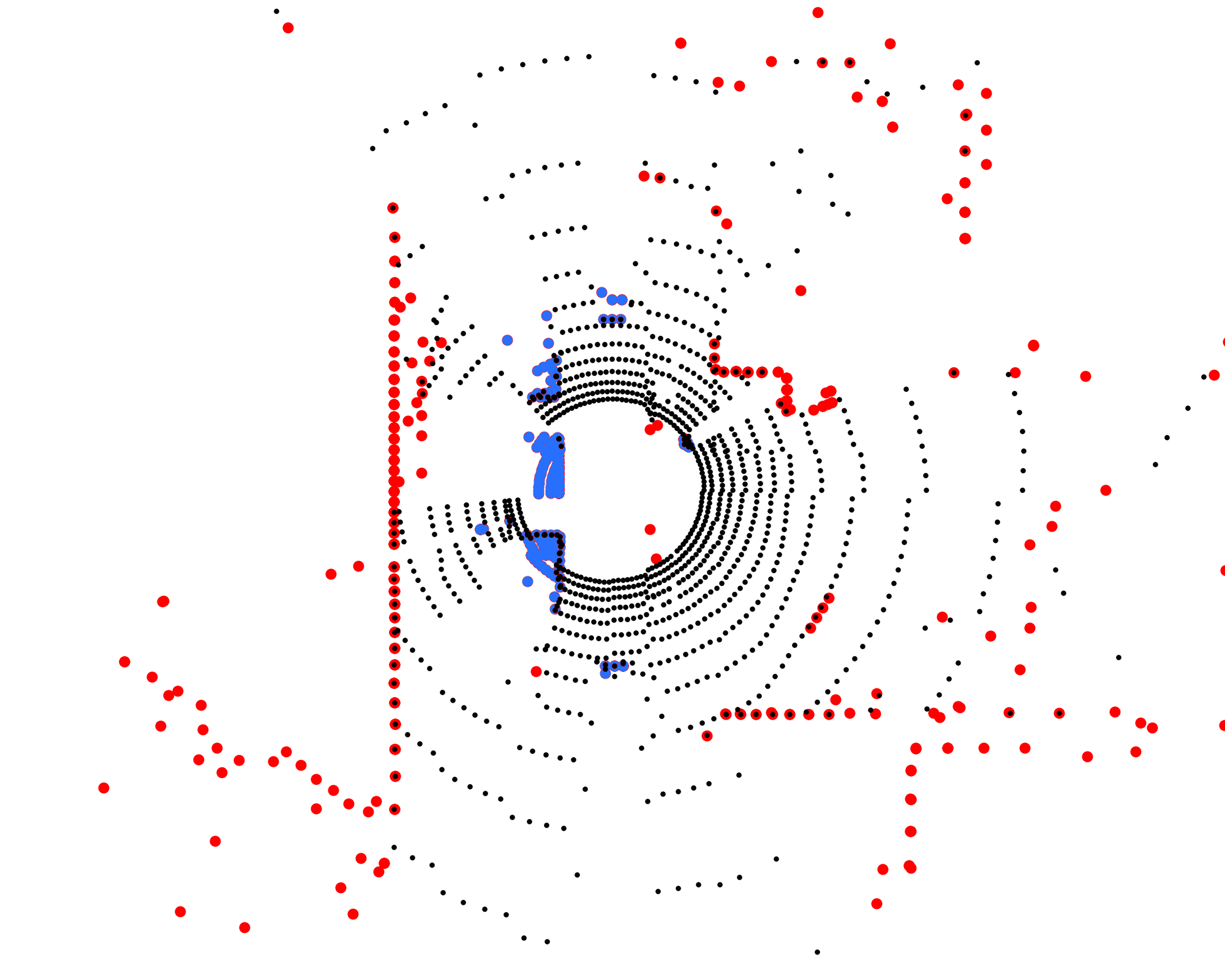}}  & \frame{\includegraphics[width=0.34\linewidth, trim=8.9cm 0cm 8.9cm  0cm,clip]{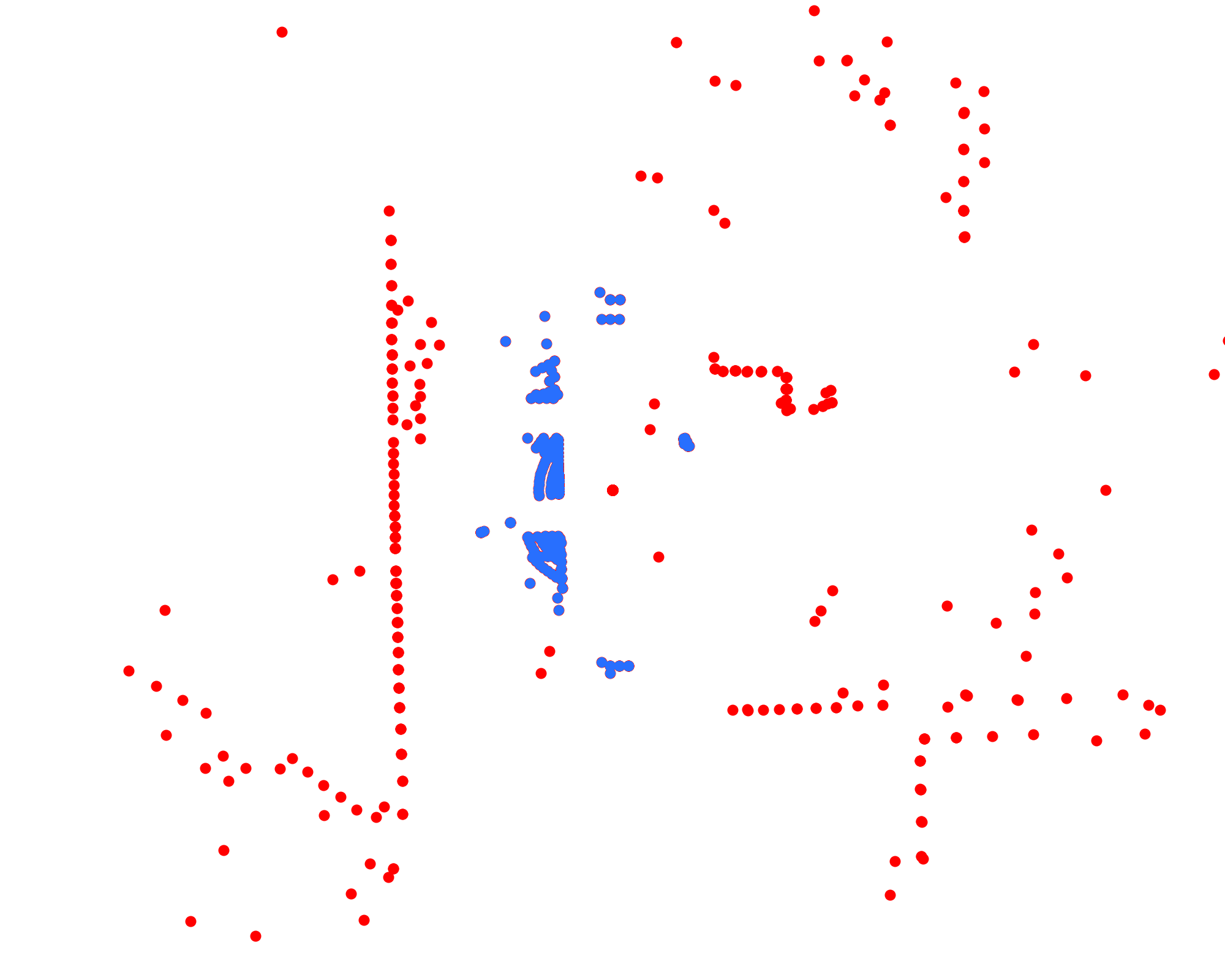}}  & \frame{\includegraphics[width=0.34\linewidth, trim=1.6cm 0cm 1.6cm  0cm,clip]{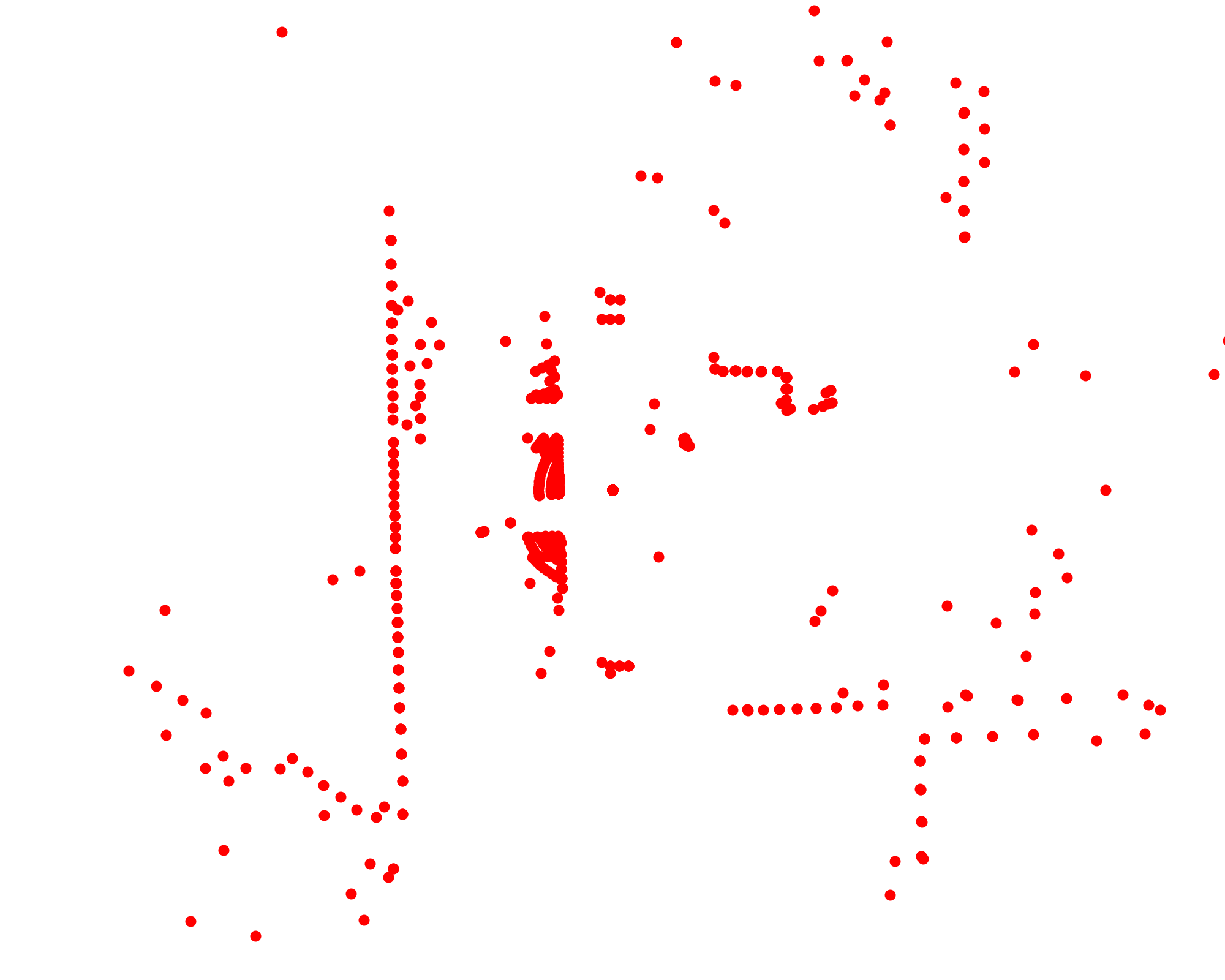}} \\
\footnotesize{Input Sparse scan} & \footnotesize{\glidar{} generated mask} & \footnotesize{Best baseline} 
\end{tabular}
\caption{\textbf{Middle:} Seg. mask generated by \glidar{} \textbf{Right:} Seg. mask generated by the best baseline. Blue denotes dynamic points. Red denotes the static non-ground points used for SLAM. The best baseline fails to segregate the static and dynamic points.}
\label{fig:carla-seg}
\end{figure}

\glidar{} generates accurate segmentation masks of dynamic objects in the absence of segmentation and object information by subtracting the dynamic input from the augmented output of \glidar{}. The rest of the points belong to static objects. The accuracy of these masks against those generated using the best baseline for sparse scans is shown in Figure \ref{fig:lidar_recon} and \ref{fig:carla-seg}. The baseline fails to separate dynamic and static points. It does not capture the global topology (Figure \ref{fig:lidar_recon}), resulting in leakage of all LiDAR points in the dynamic object mask during scan subtraction. We demonstrate the segmentation results for ARD-16 in the Supplementary material.

%% file: sec/06ablation.tex
\subsection{Ablation Studies}
\glidar{} has two parts - the LiDAR Graph network module that forms the main part of the network and the 0-dim $\mathcal{PH}$ based LiDAR backbone regularizer. We perform ablation studies to study the impact of these modules on \glidar{} in Table \ref{tab:ablation}. We demonstrate the ablation results on the CARLA test set. We observe a positive impact of both on \glidar{}'s performance. 

\begin{table}[!htbp]
\setlength{\tabcolsep}{10pt}
\begin{center}

\footnotesize{
\begin{tabular}{c c c c c} 
\toprule
 \multicolumn{1}{c}{} & \multicolumn{2}{c}{\textbf{Graph}}       & \multicolumn{2}{c}{\textbf{Topology + Graph}} \\
 \cmidrule(rl){2-3} \cmidrule(rl){4-5}
  \multicolumn{1}{c}{{\textbf{}}} & \textbf{Sparse} & \textbf{Dense} & \textbf{Sparse} & \textbf{Dense} \\
  \midrule
CARLA &  1.18&2.36     &   \textbf{1.04} & \textbf{2.08}    \\
\bottomrule

\end{tabular}
}
\end{center}

\caption{Ablation studies to study the effect of the LiDAR graph layers and the 0-dim $\mathcal{PH}$ constraint on \glidar{}.}
\label{tab:ablation}
\end{table}

We provide another ablation to where we replace the graph layers with standard convolutional layers - \textit{GNN vs CNN ablation study} using the sparse and dense version with CARLA dataset in Table \ref{tab:comparison}. ConvNets are not good at generating accurate long range structures due to only local receptivity, while graphs along with 0-dim$ \mathcal{PH}$ learn them effectively. 
\begin{table}[!htbp]
\setlength{\tabcolsep}{10pt}
\centering
\footnotesize
\begin{tabular}{lcc}
\toprule
\textbf{} & \textbf{ConvNets} & \textbf{GNN} \\
\midrule
\textbf{Sparse} & 16.91 & \textbf{1.18} \\
\textbf{Dense} & 11.3 & \textbf{2.36} \\
\bottomrule
\end{tabular}
\caption{Ablation Result for ConvNets and \glidar{}}
\label{tab:comparison}
\end{table}
\subsection{Performance}
 The inference rate of GLiDR is 4ms/scan on an NVIDIA A100 GPU, and the time for preprocessing is 30ms/scan on an Intel Xeon Silver 4208 CPU@2.10 GHz processor.

%% file: sec/07conclusion.tex
\section{Limitations} \glidar{} undergoes training on the respective training splits for different datasets. The model trained on an outdoor/ industrial/ indoor setting will find it hard to generalize to other environments. This is an important and fundamental problem of domain adaptation, which is not the goal of this work. Our premise is - \glidar{} works extremely well in sparse environments when trained with scans captured in similar environments without the use of any form of labeled information.

\section{Conclusion}
Sparse LiDAR in restricted and constrained settings can pose numerous challenges to navigation. We address these challenges by augmenting them with points along static structures. We present \glidar{} that utilizes a graph representation for LiDAR. We regularize the model using 0-dimensional $\mathcal{PH}$ constraint that provides a static LiDAR backbone along which points can be accurately generated. The new points allow \glidar{} to achieve superior navigation performance against current standards that may or may not use labeled information. We are hopeful that our approach for \glidar{} will be very useful for future research on standard shape-based point cloud research and applications.
\newpage
